\documentclass[a4paper,fleqn]{cas-dc}

\usepackage[authoryear]{natbib}
\usepackage{subfig}
\usepackage{tabularx}

\usepackage{siunitx}

\def\tsc#1{\csdef{#1}{\textsc{\lowercase{#1}}\xspace}}
\tsc{WGM}
\tsc{QE}
\tsc{EP}
\tsc{PMS}
\tsc{BEC}
\tsc{DE}

\begin{document}
\let\WriteBookmarks\relax
\def\floatpagepagefraction{1}
\def\textpagefraction{.001}
\shorttitle{Multi-session multimodal underwater mapping}
\shortauthors{P. Philip-Ifabiyi et~al.}

\title [mode = title]{Multi-Session Multimodal Underwater Mapping with Acoustic and Optical Imaging}                      



\author[1,2]{Precious Philip-Ifabiyi}[
                        orcid=0009-0009-3680-7723]

\ead{precious.philip-ifabiyi@fer.hr}
\credit{Conceptualisation, Methodology, Software,
Writing – original draft}
\affiliation[1]{organization={Computer Vision and Robotics Research Institute (ViCOROB), University of Girona, Campus Montilivi, Edifici P4},
    postcode = {17003},
                city={Girona},
                country={Spain}}

\author[1]{Valerio Franchi}[
                        orcid=0000-0002-9592-9618]
\ead{valerio.franchi@udg.edu}
\credit{Conceptualisation, Methodology, Supervision, Writing – review \& editing}

\author[2,3]{Fausto Ferreira}[orcid=0000-0003-1954-5388]
\ead{fausto.ferreira@fer.hr}
\credit{Conceptualisation, Methodology, Supervision, Writing – review \& editing}
\affiliation[2]{organization={University of Zagreb Faculty of Electrical Engineering and Computing},
                city={Zagreb},
                country={Croatia}}
\affiliation[3]{organization={MARBLE - Centre of Excellence in Maritime Robotics and Technologies for Sustainable Blue Economy}, city={Zagreb},
                country={Croatia}}

\author[1]{Nuno Gracias}[orcid=0000-0002-4675-9595]
\cormark[1]
\ead{ngracias@silver.udg.edu}
\credit{Conceptualisation, Methodology, Data collection, Supervision, Writing – review \& editing}

\cortext[cor1]{Corresponding author}


\begin{abstract}
Accurate seafloor mapping is essential for marine science, archaeology, and environmental monitoring. However, integrating data from side-scan sonar and optical cameras collected across separate survey sessions remains challenging due to positioning drift and sensor offsets. This paper presents a multi-session, multimodal underwater mapping framework based on factor graph optimization. The method jointly optimizes vehicle trajectories, 3D landmark positions, sensor extrinsics, and per-session global alignment transformations. By combining rigid inter-session corrections with local trajectory deformations, it compensates for both inter-session offsets and intra-session distortions from accumulated navigation errors. A host-relative landmark parameterization extends inverse depth to side-scan sonar, confining each landmark to the valid side of its host. The methodology was validated on real-world datasets from the Catalan coast. Against a rigid-alignment baseline, the mean sonar reprojection error is four times lower, and the median error on tie points withheld from optimization falls to 2.22 m, versus 3.48 m for the baseline and 6.24 m before optimization. The hosted parameterization recovers anchors from 10 m offsets to about 1 m, whereas with world-frame landmarks, even 0.5 m offsets grow rather than shrink. These findings demonstrate the potential of the proposed framework to generate coherent multimodal seafloor maps from heterogeneous underwater surveys.

\end{abstract}




\maketitle

\section{Introduction}
Accurate mapping and classification of seabed topography, structures, and habitats are crucial for numerous applications, including fundamental scientific research, maritime archaeology, and ecological monitoring. Traditionally, bathymetric and seafloor imaging data are acquired using acoustic sensors. Multibeam echosounders (MBES) provide bathymetry of the seafloor, while side-scan sonars (SSS) generate detailed acoustic backscatter imagery, revealing textural information about the seafloor. Sonar-based technologies are particularly essential for extensive surveys and operations in deep or turbid waters where visibility is poor \citep{7761354}.

However, the interpretation of acoustic data, particularly SSS imagery, is a complex and labor-intensive process, typically performed manually and offline by domain experts in marine geology and biology \citep{RAJANI2023115647}. SSS imagery is further affected by inherent limitations, such as low spatial resolution, variability in feature appearance across surveys, geometric distortions, and susceptibility to acoustic noise \citep{woock2010deep}.

To aid the interpretability of acoustic data and improve the robustness of seafloor characterization, optical imaging serves as a valuable complementary modality. Typically deployed from Autonomous Underwater Vehicles (AUVs) or Remotely Operated Vehicles (ROVs), optical imaging systems capture high-resolution visual information that acoustic sensors cannot provide. Their effective range is nonetheless constrained by light attenuation (absorption and scattering) and water turbidity \citep{song2022optical}. 

A significant operational challenge arises when fusing acoustic and optical maps collected across different survey sessions due to vehicle localization inaccuracies. In the absence of continuous, precise external positioning (e.g., from Long Baseline (LBL) systems or GNSS), the dead-reckoning navigation systems inherent to some AUVs accumulate drift over time. These inaccuracies lead to spatial misalignments between acoustic and optical imagery, which are normally acquired in different missions and thus possess their own georeferencing uncertainties. Such misalignments, especially between datasets from different sessions, fundamentally limit effective data fusion, comprehensive interpretation, and long-term monitoring.

This paper introduces a novel methodology for multimodal seafloor mapping that integrates side-scan sonar and optical imagery. To the best of our knowledge, this is the first approach that systematically addresses the spatial alignment and fusion of these two sensing modalities across multiple survey sessions. We achieve this by performing unimodal and multimodal feature detection and matching on optical and side-scan sonar images. These feature matches are then used by a factor graph optimization backend, alongside other sensor/filtered measurements, to estimate globally-aligned session trajectories. The proposed framework is adaptable for aligning datasets composed solely of side-scan sonar or optical imagery from multiple sessions. The proposed framework enhances the interpretation and classification of the ocean floor and provides a foundation for further developments in autonomous underwater navigation and exploration. In this sense, the code developed in this work is made available and open-source for the benefit of the community at: \url{https://github.com/CIRS-Girona/optical-acoustic-slam}.

In summary, our contributions are as follows:
\begin{itemize}
\item  We propose a framework that globally aligns side-scan and optical images collected across different sessions into a consistent, georeferenced, multi-session multi-modal map.  It jointly estimates session alignment, trajectories, landmarks, and camera extrinsics. 
\item A host-relative landmark parameterization that extends the principles of inverse-depth parameterization from monocular cameras to side-scan sonar. It confines each landmark to the valid side of the sensor that first observed it, thus eliminating mirrored solutions. It also allows a session to move rigidly through its anchor alone, without displacing every landmark it observes.
\item The framework is able to deal with datasets where only the filtered poses of the vehicles are available, and datasets where both filtered poses and sensor measurements are available, such as DVL velocities. Furthermore, the use of a constant velocity motion model and a keyframe selection strategy ensure smoothness of the estimated trajectory and computational tractability of the method.

\end{itemize}

The remainder of this paper is organized as follows: Section \ref{sec:related_work} reviews the relevant literature on acoustic sensing, optical imaging, multimodal localization, and mapping techniques.
Section \ref{sec:methodology} presents the proposed methodology for aligning sonar and optical images, including data preprocessing, feature extraction, and the multi-session, multimodal optimization strategy.
Section \ref{sec:results} discusses the experimental setup and data used, presents the results, and analyzes the performance of the proposed approach. Section \ref{sec:limitations} discusses the limitations of the proposed approach.
Section \ref{sec:conclusion} concludes the paper by summarizing key findings and suggesting directions for future research.
\section{Related Work}
\label{sec:related_work}

\subsection{Acoustic Mapping}
The use of SSS images in Simultaneous Localization and Mapping (SLAM) has been an active research area for over two decades \citep{aulinas2010feature, 1282547, zhang2024fully, zhang2023dense}. \citet{aulinas2010feature} and \citet{zhang2024fully} employed feature-based SLAM techniques to extract and track features across multiple SSS images. While \citet{aulinas2010feature} used an Extended Kalman Filter (EKF)-based approach, \citet{zhang2024fully} adopted pose-graph optimization, demonstrating improved trajectory estimation. \citet{zhang2023dense} further introduced a subframe-based dense SLAM framework that integrates SSS data to optimize AUV trajectory and seafloor mapping. Their results indicated that incorporating constraints from SSS images effectively reduced localization drift. 
More recently, \citet{xie2025neurss} proposed NeuRSS, which alternates between SSS-based bathymetry estimation through neural rendering and pose-graph SLAM. The estimated bathymetry provides elevation priors for sonar landmarks, reducing geometric ambiguity during relative pose estimation. The corrected trajectory then improves the bathymetric reconstruction. NeuRSS focuses on sonar-based localization and bathymetric mapping, whereas our work addresses the alignment of SSS and optical observations across separate survey sessions.

Beyond SSS-based SLAM, research effort has also been directed at feature-based SLAM frameworks using forward-looking sonars (FLS) \citep{westman2018feature, li2018pose}.

\subsection{Multimodal Feature Matching}
The fusion of sonar and optical imagery for underwater mapping has garnered significant attention, primarily due to the complementary nature of these sensing modalities. Existing methodologies for multimodal feature matching can be broadly categorized into three main approaches: (1) transformation-based methods that convert images into a common domain before feature matching; (2) direct feature extraction techniques that identify correspondences across modalities without domain transformation; and (3) projection-based methods that project features from one modality into the coordinate system of another.

The first category focuses on transforming one imaging modality to replicate the style of another, thereby simplifying the feature-matching process. \citet{ABU2023109868} proposed a framework that transforms optical images into Synthetic Aperture Sonar (SAS) images, followed by handcrafted feature descriptor extraction for classification. \citet{jang2019cnn} developed a style-transfer algorithm using a VGG-19 network to convert acoustic images into an optical style before applying the SIFT descriptor for matching. While innovative, the generalizability of this method cannot be assessed given that it was validated on a single test case. Building on this, \citet{s21217043} extended the approach by modifying the VGG-19 network to perform bidirectional transformations between acoustic and optical images. Their method utilized HardNet descriptors for feature matching, which reportedly yielded superior results compared to SIFT. However, their evaluation was also conducted on a limited dataset, highlighting the need for broader testing to confirm its robustness. In a subsequent work, \citet{jang2021multi} enhanced their original approach by integrating SuperGlue \citep{sarlin20superglue} for more robust feature matching between the style-transferred sonar image and the optical image, which significantly increased the quantity and quality of correspondences. This body of work underscores the potential of style transfer but also illustrates the need for evaluation across more diverse underwater environments to establish broader applicability.

The second category of research bypasses domain transformation in favor of direct cross-modal feature matching. \citet{liu2020scale} introduced a scale-adaptive matching algorithm for optical and forward-looking sonar (FLS) images, which combined iterative image enhancement with a correlation filter (MOSSE) and Gaussian scale-space. This technique effectively addressed noise, scale, and angular variations, achieving high matching success rates. However, its evaluation in a controlled pool environment revealed performance degradation with poor image quality or in the presence of acoustic shadows, suggesting limitations for real-world marine applications. Recent data-driven methods learn modality-invariant features from cross-modal data. XoFTR \citep{tuzcuouglu2024xoftr} trains a detector-free transformer on visible-thermal pairs, while MINIMA \citep{ren2025minima} generates synthetic multimodal training pairs from RGB images and uses them to fine-tune existing matchers, including LoFTR \citep{sun2021loftr}, LightGlue \citep{lindenberger2023lightglue}, and the DINOv2-based RoMa \citep{edstedt2024roma}. On side-scan sonar registration, the results are mixed. SuperPoint with LightGlue performed best on repeat-track imagery, although results degraded on noisy data \citep{katrusha2025change}. For images from adjacent survey lines, learning-based methods performed inconsistently, with MINIMA generalizing best, and none matched a non-rigid registration method initialized from manually selected correspondences \citep{lei2026coarse}. 

To the best of our knowledge, no learned matcher has yet been shown to establish side-scan sonar-optical correspondences.  Nonetheless, research effort has been devoted on ways to reduce the strong view dependence and geometric distortions that affect SSS imagery, thus providing building blocks for optical–SSS matching. Example of such are the method of \citep{lei2026blind}, which uses knowledge distillation to correct geometric distortions from a single SSS image, without navigation or attitude measurements, and PhysDNet \citep{lei2026physdnet} which separates sonar intensity into seabed reflectivity, terrain elevation, and propagation loss using a physics-guided network.
The backend used in our proposed approach is independent of how correspondences are obtained, and the estimated trajectories could restrict a learned matcher's search to small windows in future work.

The third category uses the known intrinsic and extrinsic sensor parameters to project features between modalities. This approach is highly dependent on accurate sensor calibration and projection models. \citet{10308592} developed a system that uses a multibeam FLS to resolve the scale ambiguity inherent in monocular camera systems. Their method accounts for the sonar's lack of elevation data by representing a sonar feature as a vertical line segment on the camera's image plane. A visual feature from the SLAM-generated point cloud is considered a match if its 2D projection falls on this sonar-derived segment. \citet{qiu2024improved} employed a similar concept but fused sonar data with features from a stereo camera to obtain precise depth information. Instead of projecting a single point, their method defines a 3D spatial region corresponding to a strong sonar echo and associates all visual features within this volume as a group of "candidate" matches. In a different approach, \citet{zhang2024integration} integrated a stereo camera, imaging sonar, and an IMU within a tightly-coupled SLAM framework. Their method projects detected visual landmarks into the 2D sonar image coordinate system and establishes a match by finding the closest sonar landmark whose Euclidean distance falls below a predefined threshold.

\subsection{Multi-Sensor Fusion}
This section reviews key works in multimodal mapping and SLAM that specifically integrate acoustic and optical sensors for underwater applications.

The inherent limitations of single-sensor systems in dynamic underwater environments have driven the development of multi-sensor fusion techniques. While vision-based systems excel in clear water, they falter in turbid or low-light conditions where acoustic sensors remain robust. On the other hand, images produced by acoustic sensors have lower resolution compared to those of vision-based systems \citep{7761354}. 

\citet{hover2012advanced} demonstrated an integrated system for autonomous ship hull inspection that combines data from an imaging sonar and a monocular camera. Their approach is notable in that it does not perform direct feature matching between the two sensor modalities. Instead, the sensors operate concurrently, often imaging different parts of the hull, and their measurements are fused as separate constraints within a unified pose-graph SLAM framework (iSAM). This back-end fusion allows for real-time, drift-free vehicle control relative to the previously unseen hull structure by leveraging the strengths of each sensor independently.

\citet{kunz2012autonomous} developed a framework for AUV navigation and mapping in unstructured environments, such as coral reefs, by fusing data from multibeam sonar and stereo cameras. The method solves for vehicle trajectories by incorporating constraints from both sensor types into a factor graph. Visual constraints are derived from landmark-based reprojection error, and sonar constraints produce relative pose error. A key contribution is the inclusion of sensor offsets as variables to be estimated within the graph, improving the precision of the final map.

\citet{10308592} developed a SLAM framework that utilizes a multibeam FLS to correct the scale ambiguity of a monocular visual SLAM system by estimating a depth ratio. This ratio is stabilized over time using Maximum Likelihood Estimation (MLE) on a set of correspondences. Although the camera-sonar combination significantly improves localization, the final 3D map reconstruction in their work relies solely on the rescaled point cloud from the monocular camera.

Extending their earlier work on feature matching, \citet{jang2021multi} integrated their opti-acoustic matching into a multi-session SLAM framework. While \citet{negahdaripour2009opti} conducted preliminary work on opti-acoustic stereo imaging, \citet{jang2021multi} further developed and implemented an opti-acoustic pairwise factor, utilizing iSAM \citep{kaess2008isam} as the backend for SLAM optimization. Inspired by the work of \citet{kim2010multiple}, anchor nodes were used in the factor graph to estimate the relative transformation between the acoustic session and the camera session. 
The work of Jang et al. (2021) establishes an important foundation for combining acoustic and optical observations acquired in separate sessions. However, this work was demonstrated using a DIDSON FLS, and its applicability to side-scan sonar based mapping has not been explored. Our work builds on these multi-session estimation principles and addresses their application to SSS–optical field surveys. SSS waterfall observations identify ping time, channel, and slant range, and their spatial interpretation depends on the vehicle trajectory, sensor attitude, and landmark geometry. Our formulation combines the corresponding slant-range and along-track constraints with optical projections and the available navigation measurements, allowing joint adjustment of session alignment and local trajectories. The contribution lies in this SSS–optical integration and its evaluation on heterogeneous towfish and AUV surveys. Anchor nodes, in-graph extrinsic estimation, and the individual motion and projection models are established estimation components.

Other notable tightly coupled systems like SVIn2 fuse mechanical scanning profiling sonar, visual, inertial, and depth data to achieve robust performance (including initialization, loop closing, and relocalization) under challenging underwater conditions such as haze, low light, and motion blur \citep{rahman2022svin2}.

Despite the advancements in multimodal underwater mapping, a comprehensive review of the literature reveals several unaddressed areas. No existing work was found that focuses specifically on the fusion of side-scan sonar (SSS) imagery with optical camera images for joint mapping and SLAM. Research in multimodal underwater SLAM has instead focused predominantly on FLS-camera systems, leaving the SSS–camera combination largely unexplored. This omission stems from fundamental operational constraints that arise when the two sensors are used together. First, their effective operating altitudes differ substantially: optical cameras require close proximity to the seabed to achieve sufficient image resolution and contrast, whereas side-scan sonars are deployed at higher altitudes to maximize area coverage and minimize acoustic shadowing. Operating both sensors simultaneously, therefore, necessitates a compromise that degrades the performance of at least one modality. Second, the sensors exhibit non-overlapping fields of view when mounted in a downward-looking configuration. The optical camera's narrow field of view corresponds spatially to the acoustic blind zone directly beneath a side-scan sonar. As a consequence, no common scene features are simultaneously observable in both modalities, which prevents the application of standard feature-based data association and fusion techniques.

 \section{Methodology}
\label{sec:methodology}

\subsection{System Overview and Data Acquisition}
\label{sec:system_overview}

\begin{figure*}[th]
    \centering
    \includegraphics[width=\linewidth]{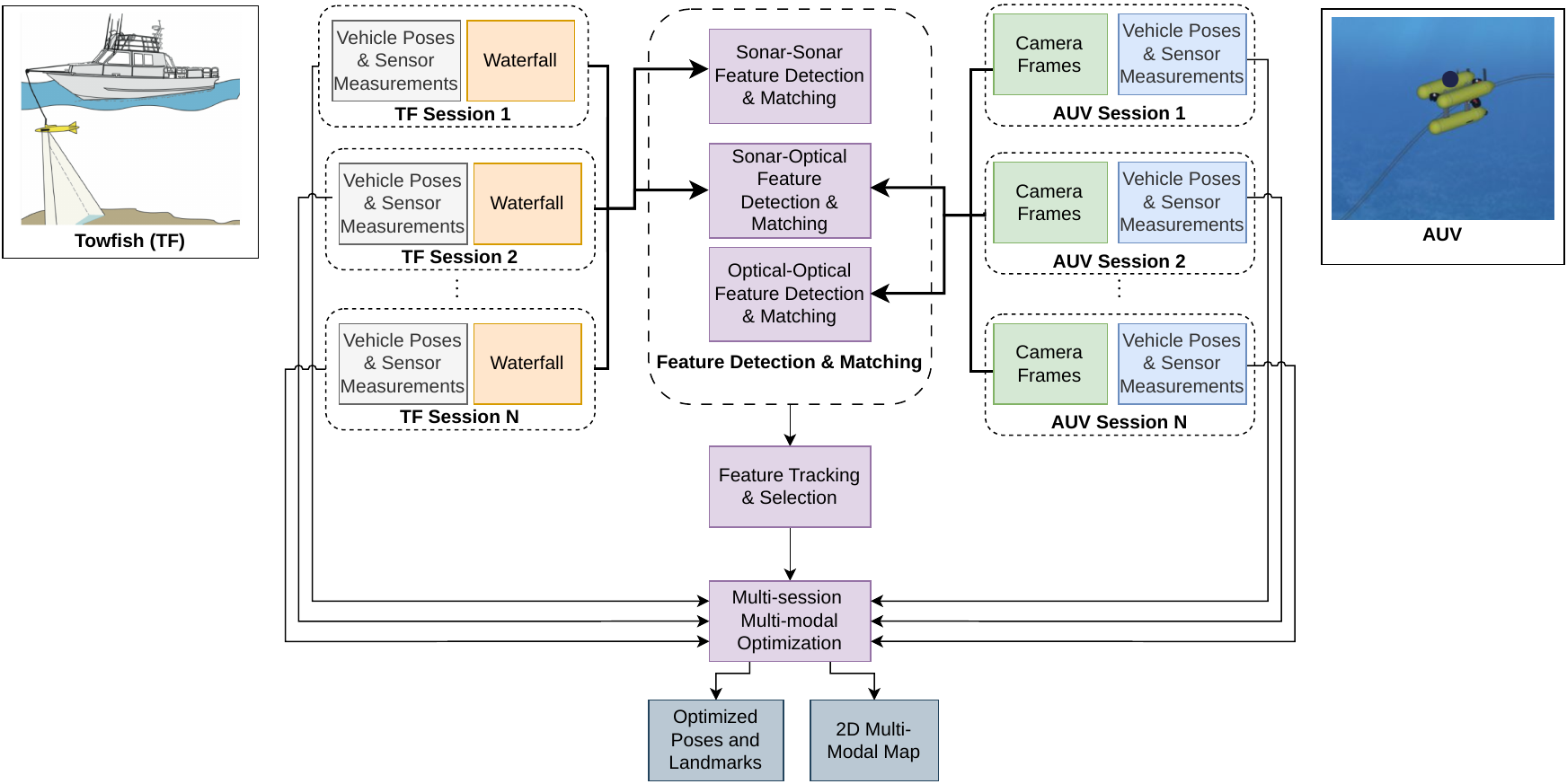} %
    \caption{An overview of the proposed system pipeline, illustrating the data flow from filtered sensor inputs through feature matching and final optimization to produce a multimodal map.}
    \label{fig:system_overview}
\end{figure*}

The proposed system is designed to process datasets collected during separate survey sessions, potentially involving different underwater platforms. An overview of the data flow is presented in Figure \ref{fig:system_overview}. The pipeline is initiated with sensor data and initial pose estimates from each session. This includes waterfall representations of SSS data and optical camera frames. The system uses three types of feature matches: sonar--sonar, optical--optical, and sonar--optical cross-modal matching. These matches are organized into feature tracks that can exist within and across sessions.
In the scope of this paper, a feature track is a collection of observations, acoustic or optical, of the same 3D point on the seafloor. Finally, a multi-session, multimodal optimization backend processes these tracks to yield optimized vehicle poses and landmark positions, culminating in a unified 2D multimodal map.

The datasets utilized in this work were gathered in a shallow-water marine environment characterized by seagrass beds of \textit{Posidonia oceanica} \citep{barcelona2021meadow} intertwined with areas of sand and hard-bottom. A more in-depth description of the datasets is provided in Section~\ref{sec:results}. Data collection was performed using two platforms in multiple sessions:
\begin{itemize}
    \item A towfish, equipped with a side-scan sonar and an integrated attitude and pressure sensor (providing roll, pitch, heading, and depth), was towed by a surface vessel. The towfish's navigation solution was achieved by combining multiple data sources: a vessel-mounted Real-Time Kinematic (RTK) GNSS provided a high-accuracy global position for the ship, while an Ultra-Short Baseline (USBL) acoustic positioning system measured the towfish's submerged position relative to the vessel. The attitude sensor provided direct orientation measurements for the submerged platform.

    \item An Autonomous Underwater Vehicle (AUV), equipped with one or more intrinsically calibrated optical cameras, was used to capture high-resolution imagery of the seafloor. Camera-to-body extrinsic parameters were determined through a manual measurement procedure and are therefore subject to uncertainties of up to approximately 5$^\circ$ in orientation and 10 cm in translation. The AUV's navigation solution was provided by an onboard Integrated Navigation System (INS) built on a Fiber-Optic Gyroscope (FOG) and a Micro-Electro-Mechanical Systems (MEMS) accelerometer, and aided by a Doppler Velocity Log (DVL) and a pressure sensor for depth. At the start of each session/mission, GNSS measurements obtained at the surface were used to initialize the vehicle's navigation state.
\end{itemize}

\subsection{Feature Detection, Matching, and Tracking}
\label{sec:feature_processing}

A fundamental component of this framework is the establishment of robust data associations (features) both within a single data stream (unimodal) and between different sensor types (multimodal).

\subsubsection{Optical-Optical Feature Matching}
For registering consecutive optical images, features are detected and described using the Scale-Invariant Feature Transform (SIFT) algorithm. We chose SIFT because it is a well-established and robust feature detector and descriptor. Given that the data was collected near the seafloor with adequate artificial illumination, no explicit image pre-processing was performed. SIFT features are matched between temporally sequential images to establish local motion constraints. Then, RANSAC is used to remove outliers in the matches.

\subsubsection{Sonar-Sonar and Sonar-Optical Feature Matching}
Due to significant differences in data phenomenology and the presence of high-texture, low-feature regions like seagrass, robust automated feature matching between sonar and optical images, as well as between two sonar images, remains an open research challenge. Consequently, correspondences for both sonar-sonar and sonar-optical pairs were established manually in this work using a custom annotation tool\footnote{https://github.com/CIRS-Girona/xtf-annotation-tool}. Automated sonar and cross-modal matching, although a topic of ongoing research effort \citep{lei2026coarse}, are outside the scope of this study.

Matches were selected at recognizable physical features, including distinctive seabed boundary configurations and salient objects visible in the respective observations. We use the unoptimized sonar and optical trajectories to generate a preliminary multi-session multimodal map. Although this map is not accurately aligned, it provides an approximate co-registration of all sessions, which narrows the search for corresponding regions. In this map, the annotator finds areas where similar geographic features and landmarks appear in more than one session. 
Each SSS observation is recorded by its ping index, port or starboard channel, and range bin, while optical observations are recorded in image coordinates. These associations are subsequently organized into landmark tracks and incorporated into the optimization described in Section \ref{sec:factor_graph_optimization}.

\subsubsection{Feature Tracking and Selection}
Following pairwise matching, features are organized into tracks that represent the same 3D landmark observed over time. A feature track can consist of observations from a single modality (e.g., optical-only or sonar-only) or from multiple modalities (a mix of optical and sonar observations). These tracks form the basis of the constraints used in the global optimization. To enhance the robustness of this process, a selection mechanism is applied to the optical-only feature tracks.

\subsubsection{Optical-Only Feature Track Filtering Algorithm}
\label{sub:feature_track_filtering}
To improve the quality of the data associations and reduce computational overhead, a selection process is used to retain only the most reliable optical-only feature tracks. The core principle is to prioritize tracks that are both long, indicating feature stability, and well-distributed throughout the image sequence.

The selection criterion is as follows: for each image in a session, we identify the $N$ feature tracks with the greatest total length (i.e., the highest number of observations) that appear in that image. The final set of optical-only tracks used in the optimization is the union of these top $N$ tracks from every image.

This strategy effectively discards short, potentially spurious tracks while ensuring that every part of the trajectory is constrained by its most persistent and reliable visual features. The $N$ parameter should be set to at least 4 tracks to improve trajectory reconstruction.

\subsection{Multi-Session Optimization}
\label{sec:factor_graph_optimization}

\subsubsection{Coordinate Frames}
\label{sec:coordinate_frames}
The optimization variables are expressed in four types of frames. The world frame $\mathcal{W}$ is the georeferenced NED frame in which the navigation data of all sessions are provided. Each session $j$ has a local anchor frame $\mathcal{A}^j$, and $\mathcal{B}_k$ denotes the vehicle base frame at keyframe $k$. We write ${}^{\mathcal{W}}\bar{\mathbf{T}}^{j}_{k} \in \mathrm{SE}(3)$ for the georeferenced navigation pose of $\mathcal{B}_k$ in $\mathcal{W}$ at keyframe $k$ of session $j$. The trajectory of each session is estimated in its own anchor frame $\mathcal{A}^j$, which is defined by the first pose of the session, and the placement of $\mathcal{A}^j$ in $\mathcal{W}$ is estimated jointly with all other variables. Each landmark is expressed in the sensor frame of its host observation (Section~\ref{sec:landmark_param}). This is the camera frame of the first camera observation for landmarks seen by a camera, and the sonar frame of the first observation for sonar-only landmarks.

\subsubsection{Problem Formulation}
\begin{figure*}[th]
    \centering
    \includegraphics[width=\linewidth]{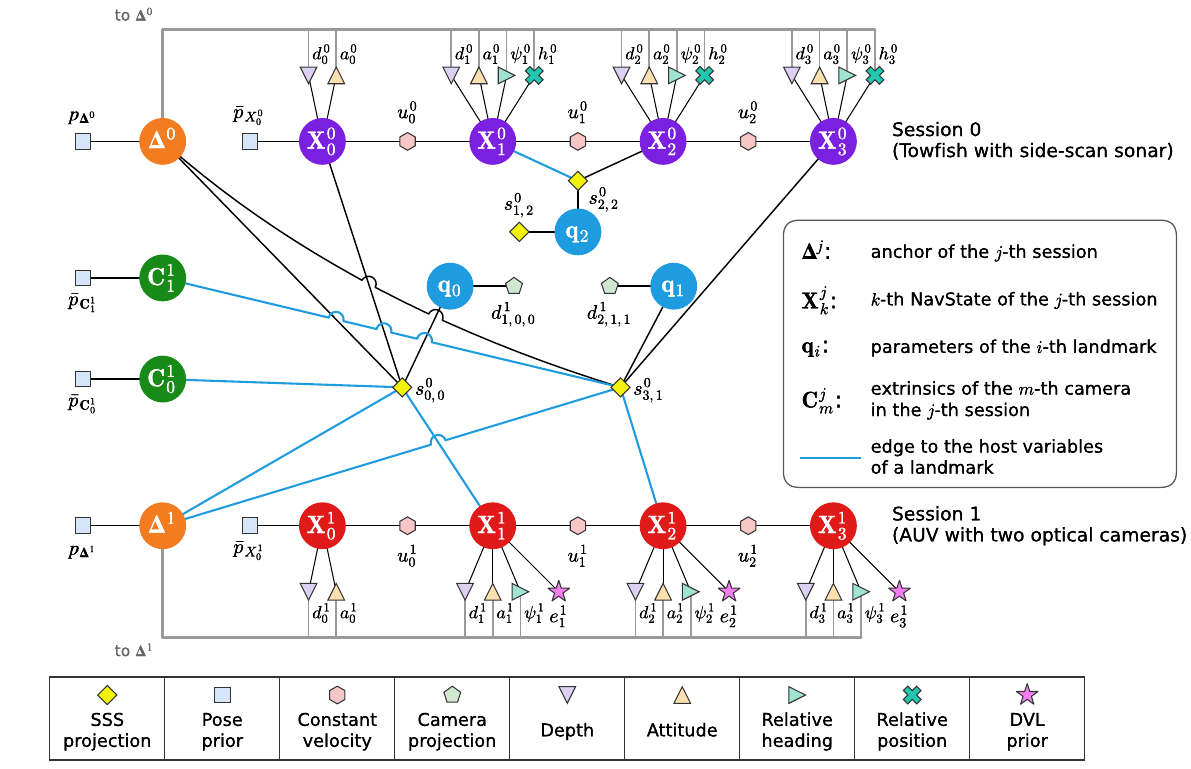} %
    \caption{The factor graph representation of the proposed multi-session, multimodal mapping framework. It connects vehicle states ($\mathbf{X}_k^j$), landmark parameters ($\mathbf{q}_i$), camera extrinsics ($\mathbf{C}_m^j$), and session anchors ($\boldsymbol{\Delta}^j$) through various measurement and motion factors. Landmarks are parameterized relative to a host (Section~\ref{sec:landmark_param}): $\mathbf{q}_0$ and $\mathbf{q}_1$ are hosted on a camera observation, and the sonar-only landmark $\mathbf{q}_2$ is hosted on its first sonar observation.}
    \label{fig:factor_graph}
\end{figure*}

To jointly optimize the vehicle trajectories, landmark positions, and sensor extrinsics, we formulate the problem as a factor graph. This graphical model allows for the flexible fusion of heterogeneous information sources. The graph, shown in Figure \ref{fig:factor_graph}, consists of variable nodes representing the quantities to be estimated and factor nodes representing probabilistic constraints on these variables, derived from sensor measurements and motion models.

The primary variables to be estimated are:
\begin{itemize}
    \item $\boldsymbol{\boldsymbol{\Delta}}^j \in \mathrm{SE}(3)$: The anchor node of session $j$, i.e.\ the pose of the anchor frame $\mathcal{A}^j$ in the world frame $\mathcal{W}$.

    \item $\mathbf{X}_k^j = \begin{bmatrix} \mathbf{T}_k^j & \mathbf{v}_k^j \end{bmatrix} \in \mathrm{SE}(3)\times\mathbb{R}^3$: The 9-DOF navigation state of the vehicle at time $k$ in session $j$, comprising the pose $\mathbf{T}_k^j \in \mathrm{SE}(3)$ of the base frame $\mathcal{B}_k$ in the anchor frame $\mathcal{A}^j$ (with translation $\mathbf{t}_k^j$ and orientation $\mathbf{R}_k^j$) and its linear velocity $\mathbf{v}_k^j \in \mathbb{R}^3$, also expressed in $\mathcal{A}^j$.

    \item $\mathbf{q}_i \in \mathbb{R}^3$: The parameters of landmark $i$. Each landmark is parameterized relative to the sensor pose of one of its observations, its host (Section~\ref{sec:landmark_param}), and its world position $\mathbf{p}_i \in \mathbb{R}^3$ in $\mathcal{W}$ is a function of $\mathbf{q}_i$ and of that pose.

    \item $\mathbf{C}_m^j \in \mathrm{SE}(3)$: The extrinsic calibration of camera $m$ in AUV session $j$, i.e.\ the pose of the camera in the base frame $\mathcal{B}_k$.
\end{itemize}

We adopt the concept of an anchor node for each session to align data from different sessions into a common global reference frame \citep{kim2010multiple, mcdonald2013real}. The anchor, $\boldsymbol{\boldsymbol{\Delta}}^j$, represents the transformation of the $j$-th session's local coordinate frame relative to the world frame, so the pose of the vehicle in $\mathcal{W}$ at keyframe $k$ is

\begin{equation}
    {}^{\mathcal{W}}\mathbf{T}^{j}_{k} = \boldsymbol{\Delta}^{j}\,\mathbf{T}^{j}_{k}.
    \label{eq:world_pose}
\end{equation}

We denote by $\bar{\boldsymbol{\Delta}}^j \triangleq {}^{\mathcal{W}}\bar{\mathbf{T}}^{j}_{0}$ the first georeferenced navigation pose of session $j$. It is the initial value of $\boldsymbol{\Delta}^j$ (Section~\ref{sec:initial_estimates}), and its horizontal position and heading are the mean of the anchor prior (Eq.~\eqref{eq:prior_anchor}).

The measurements, $\mathbf{Z}$,  are detailed alongside their corresponding factor definitions in subsequent sections.

Let the complete set of variables to be optimized be $\boldsymbol{\Theta} = \{\{\boldsymbol{\boldsymbol{\Delta}}^j\}, \{\mathbf{X}_k^j\}, \{\mathbf{q}_i\}, \{\mathbf{C}_m^j\}\}$. The goal is to find the set of state variables, $\boldsymbol{\Theta}$, that maximizes the posterior probability given all measurements and priors, $\mathbf{Z}$, which corresponds to the Maximum a Posteriori (MAP) estimate:

\begin{equation}
    \boldsymbol{\Theta}^* = \underset{\boldsymbol{\Theta}}{\operatorname{argmax}} \ P(\boldsymbol{\Theta} | \mathbf{Z})
\end{equation}

\subsubsection{Data Pre-processing}
A key pre-processing step is to convert the raw side-scan sonar keypoints, identified in the waterfall image space by the triplet (\texttt{ping\_index}, \texttt{side}, \texttt{bin\_index}), into metric slant range measurements. \texttt{ping\_index} defines the position of the ping in the along-track sequence, \texttt{side} specifies whether the feature was observed on the port (left) or starboard (right) channel of the sonar, and \texttt{bin\_index} is the position of the sample within that channel (counted outwards from the sonar). Both indices start at zero.
The slant range, $R_k$, for a keypoint is determined via linear interpolation based on the maximum recorded slant range, $R_{\text{max}}$, for that specific ping and side. The calculation is:
\begin{equation}
    R_k = R_{\text{max}} \cdot \frac{\text{bin\_index}_k + 1}{\text{bin\_index}_{\text{max}} + 1}
    \label{eq:slant_range_interpolation}
\end{equation}
The resulting SSS measurement vector, $\mathbf{s}_{k,i}^j$, used in the optimization is therefore $\mathbf{s}_{k,i}^j = \begin{bmatrix} R_k & 0 \end{bmatrix}^\top$. The second element represents the zero along-track distance, as the measurement is assumed to lie on a plane perpendicular to the vehicle's direction of travel. The SSS projection factor and its associated noise model are detailed in Section \ref{para: sss factor}.

\subsubsection{Factor Definitions and Error Models}
\label{sec:factor_definitions}
This section details the mathematical models for the factors used in the optimization. Henceforth, for a pose $\mathbf{T}$, we write $\mathbf{R}(\mathbf{T})$ for its rotation, $\mathbf{t}(\mathbf{T})$ for its translation, and $\psi(\mathbf{T})$ for its yaw (heading); $\psi(\cdot)$ is also applied to rotations. For a translation, $z(\cdot)$ is its vertical component and $[\cdot]_{xy}$ its horizontal components. $\langle\cdot\rangle$ wraps an angle to $[-\pi, \pi)$, and $\mathbf{e}_3$ is the vertical axis of $\mathcal{W}$.

\paragraph{Initial State and Anchor Priors}
These factors initialize the optimization by providing an absolute reference for each session. A prior is placed on the first state of each session, $\mathbf{X}_0^j$, and on the session's anchor pose in the world frame, $\boldsymbol{\boldsymbol{\Delta}}^j$. 

The error for the initial state prior, $\mathbf{r}_{\text{X}_0}(\cdot)$, is a 9-dimensional vector composed of a 6-DOF pose error and a 3-DOF velocity error. The error is computed by treating each component appropriately: the pose error is calculated on the tangent space of the $\mathrm{SE}(3)$ manifold, while the velocity error is calculated in the $\mathbb{R}^3$ space. Given a prior mean $p_{\mathbf{X}_0^j}$, composed of pose $\bar{\mathbf{T}}_{\mathbf{X}_0^j}$ and velocity $\bar{\mathbf{v}}_{\mathbf{X}_0^j}$, the residual vector is given by:
\begin{equation}
    \mathbf{r}_{\text{X}_0}\left(\mathbf{X}_0^j, p_{\mathbf{X}_0^j}\right) = 
    \begin{bmatrix}
        \log\left(\left(\bar{\mathbf{T}}_{\mathbf{X}_0^j}\right)^{-1} \mathbf{T}_0^j\right)^{\vee} \\
        \mathbf{v}_0^j - \bar{\mathbf{v}}_{\mathbf{X}_0^j}
    \end{bmatrix}
    \label{eq:prior_initial_state_revised}
\end{equation}
\noindent where $\log(\cdot)^\vee$ maps the manifold error to a Euclidean tangent space ($\mathbb{R}^6$) for optimization. The prior pose is the identity, $\bar{\mathbf{T}}_{\mathbf{X}_0^j} = \mathbf{I}$, which defines $\mathcal{A}^j$ as the frame of the first keyframe. With $\mathbf{T}_0^j = \mathbf{I}$, the anchor coincides with the first keyframe's world pose. The depth, roll, and pitch of the anchor, which the anchor prior below leaves free, are therefore those of the first keyframe's world pose, which the depth and attitude factors constrain like those of any other keyframe.

The anchor prior models the session-wide offset of the georeferenced navigation, which affects only the horizontal position and the heading. Its residual is
\begin{equation}
    \mathbf{r}_{{\Delta}}\left(\boldsymbol{\Delta}^j,\bar{\boldsymbol{\Delta}}^j\right) =
    \begin{bmatrix}
        \left[\mathbf{t}(\boldsymbol{\Delta}^j) - \mathbf{t}(\bar{\boldsymbol{\Delta}}^j)\right]_{xy} \\
        \left\langle \psi\left(\boldsymbol{\Delta}^j\right) - \psi\left(\bar{\boldsymbol{\Delta}}^j\right) \right\rangle
    \end{bmatrix}
    \label{eq:prior_anchor}
\end{equation}
\noindent The navigation and landmark measurements determine the sessions only relative to one another, so the horizontal position and heading of the map as a whole must be fixed by a choice of gauge. We use the first sonar session as the reference and give it a tight anchor prior.

\paragraph{Camera Extrinsics Prior}
To provide a good initial guess and constrain the optimization, a prior is placed on the extrinsic calibration, $\mathbf{C}_m^j$, for each camera $m$ in each AUV session $j$. The residual vector $\mathbf{r}_\text{C}(\cdot)$ is defined similarly on the tangent space of the $\mathrm{SE}(3)$ manifold:
\begin{equation}
    \mathbf{r}_\text{C}\left(\mathbf{C}_m^j, \bar{p}_{\mathbf{C}_m^j}\right) = \log\left(\left(\bar{p}_{\mathbf{C}_m^j}\right)^{-1} \mathbf{C}_m^j\right)^{\vee}
    \label{eq:prior_extrinsics}
\end{equation}
\noindent where $\bar{p}_{\mathbf{C}_m^j}$ is the initial estimate for the camera's pose.

\paragraph{Navigation Factors}
\label{sec:nav_factors}
The components of the navigation data described in Section~\ref{sec:system_overview} differ in how their errors behave within a session. On both platforms, depth is measured by a pressure sensor, and roll and pitch are referenced to gravity, so their errors do not accumulate, and we treat them as free of a session-wide offset. These components are constrained in the world frame.

In contrast, the horizontal position and heading can carry an error common to a whole session. The AUV's horizontal position is initialized from GNSS at the start of the session and then propagated by the INS from the DVL velocity and the heading, so its error grows with the distance traveled. The towfish position does not drift, since every fix is a USBL measurement relative to the GNSS-positioned vessel. However, its errors are not necessarily independent between fixes: systematic effects, such as an error in the assumed sound speed or a misalignment of the USBL transducer, change slowly and can displace a whole track. We allow for such an error with a session-wide offset. The heading of either platform can likewise be offset by a constant amount, for example by a residual mounting misalignment of the heading sensor. We therefore model these components as a session offset plus white noise. Session-wide offsets are what the anchor $\boldsymbol{\Delta}^j$ models, so these components are constrained relative to the position and heading of the anchor.

Let the navigation measurement be ${}^{\mathcal{W}}\bar{\mathbf{T}}^{j}_{k}$, and let ${}^{\mathcal{W}}\mathbf{T}^{j}_{k}$ be the estimated world pose of Eq.~\eqref{eq:world_pose}. A navigation measurement yields up to four residuals.

\begin{itemize}
    \item \textbf{Depth factor:} The depth of the world pose is compared with the measured depth:
    \begin{equation}
        r_\text{d}\left(\boldsymbol{\Delta}^j, \mathbf{X}_k^j, {}^{\mathcal{W}}\bar{\mathbf{T}}^{j}_{k}\right) = z\left({}^{\mathcal{W}}\mathbf{t}^{j}_{k}\right) - z\left(\bar{\mathbf{t}}^{j}_{k}\right)
        \label{eq:prior_depth}
    \end{equation}

    \item \textbf{Attitude factor:} Roll and pitch are constrained through the direction of gravity expressed in the body frame:
    \begin{multline}
        \mathbf{r}_\text{a}\left(\boldsymbol{\Delta}^j, \mathbf{X}_k^j, {}^{\mathcal{W}}\bar{\mathbf{T}}^{j}_{k}\right) = \\
        \boldsymbol{\epsilon}_{\mathbb{S}^2}\left(\mathbf{R}\left({}^{\mathcal{W}}\mathbf{T}^{j}_{k}\right)^{\top}\mathbf{e}_3,\; \mathbf{R}\left({}^{\mathcal{W}}\bar{\mathbf{T}}^{j}_{k}\right)^{\top}\mathbf{e}_3\right)
        \label{eq:prior_orientation}
    \end{multline}
    \noindent where $\boldsymbol{\epsilon}_{\mathbb{S}^2}(\cdot,\cdot) \in \mathbb{R}^2$ is the error between two unit vectors in the tangent space of the unit sphere at the measured direction. A rotation about the vertical axis does not change the gravity direction in the body frame, so this residual is independent of yaw.

    \item \textbf{Relative heading factor:} The heading of the world pose relative to the heading of the anchor is compared with the measured relative heading:
    \begin{multline}
        r_\psi\left(\boldsymbol{\Delta}^j, \mathbf{X}_k^j, {}^{\mathcal{W}}\bar{\mathbf{T}}^{j}_{k}, \bar{\boldsymbol{\Delta}}^j\right) = \Bigl\langle \psi\left({}^{\mathcal{W}}\mathbf{T}^{j}_{k}\right) - \psi\left(\boldsymbol{\Delta}^j\right) \\
        - \left(\psi\left({}^{\mathcal{W}}\bar{\mathbf{T}}^{j}_{k}\right) - \psi\left(\bar{\boldsymbol{\Delta}}^j\right)\right) \Bigr\rangle
        \label{eq:prior_heading}
    \end{multline}

    \item \textbf{Relative horizontal position factor:} The horizontal offset of the world pose from the anchor position is compared with the measured offset, both in the axes of $\mathcal{W}$:
    \begin{multline}
        \mathbf{r}_{xy}\left(\boldsymbol{\Delta}^j, \mathbf{X}_k^j, {}^{\mathcal{W}}\bar{\mathbf{T}}^{j}_{k}, \bar{\boldsymbol{\Delta}}^j\right) = \left[\mathbf{t}({}^{\mathcal{W}}\mathbf{T}^{j}_{k}) - \mathbf{t}(\boldsymbol{\Delta}^j)\right]_{xy} \\
        - \left[\mathbf{t}({}^{\mathcal{W}}\bar{\mathbf{T}}^{j}_{k}) - \mathbf{t}(\bar{\boldsymbol{\Delta}}^j)\right]_{xy}
        \label{eq:prior_horizontal}
    \end{multline}
\end{itemize}

\noindent All four factors are binary factors between $\mathbf{X}_k^j$ and $\boldsymbol{\Delta}^j$. Towfish keyframes receive all four. AUV keyframes receive the depth, attitude, and relative heading factors but no horizontal position factor, because while the vehicle is submerged, the filtered position is an integration of the DVL velocity and the heading rather than an independent measurement. The first keyframe of a session receives only the depth and attitude factors, since its relative position and heading are zero by definition and are fixed by the prior on $\mathbf{X}_0^j$.

The relative factors and the anchor prior together model the navigation as offset by an unknown session-wide horizontal shift $\mathbf{b}_{xy}^j = [\mathbf{t}(\boldsymbol{\Delta}^j) - \mathbf{t}(\bar{\boldsymbol{\Delta}}^j)]_{xy}$ and heading offset $b_\psi^j = \langle\psi(\boldsymbol{\Delta}^j) - \psi(\bar{\boldsymbol{\Delta}}^j)\rangle$. Substituting these into Eqs.~\eqref{eq:prior_heading} and \eqref{eq:prior_horizontal} gives

\begin{align}
    r_\psi &= \bigl\langle \psi({}^{\mathcal{W}}\mathbf{T}^{j}_{k}) - \psi({}^{\mathcal{W}}\bar{\mathbf{T}}^{j}_{k}) - b_\psi^j \bigr\rangle, \nonumber \\
    \mathbf{r}_{xy} &= \bigl[\mathbf{t}({}^{\mathcal{W}}\mathbf{T}^{j}_{k}) - \mathbf{t}({}^{\mathcal{W}}\bar{\mathbf{T}}^{j}_{k})\bigr]_{xy} - \mathbf{b}_{xy}^j,
    \label{eq:nav_offset}
\end{align}

\noindent and Eq.~\eqref{eq:prior_anchor} becomes a zero-mean prior on $(\mathbf{b}_{xy}^j, b_\psi^j)$.

\paragraph{DVL Prior Factor}
For some of the optical sessions used in this work, the DVL and gyroscope measurements were collected during the mission. For sessions with these measurements, we incorporate a DVL factor as a prior on the velocity component of the robot's state.

The measurement model of the DVL is given by:

\begin{multline}
    \mathbf{h}_\text{e} \Big(\mathbf{X}^j_{k} \Big) = \Big({}^{\mathcal{D}}\mathbf{R}_{\mathcal{B}} \cdot \left(\mathbf{R}_k^j\right)^{\top} \cdot \mathbf{v}^j_k \Big) \\ 
    + \Big({}^{\mathcal{D}}\mathbf{t}_{\mathcal{B}} \times \Big( {}^{\mathcal{D}}\mathbf{R}_{\mathcal{B}} \cdot {}^{\mathcal{B}}\mathbf{R}_{\mathcal{I}} \cdot {}^{\mathcal{I}}\boldsymbol{\omega}_k^j\Big)\Big)
\end{multline}

\noindent where ${}^{\mathcal{I}}\boldsymbol{\omega}_k^j$ is the angular velocity of the vehicle measured by the gyroscope at keyframe $k$ of session $j$, expressed in the gyroscope frame $\mathcal{I}$, ${}^{\mathcal{B}}\mathbf{R}_{\mathcal{I}}$ is the fixed rotation of $\mathcal{I}$ in the base frame $\mathcal{B}$, and ${}^{\mathcal{D}}\mathbf{R}_{\mathcal{B}}$ and ${}^{\mathcal{D}}\mathbf{t}_{\mathcal{B}}$ are the fixed rotation and translation of $\mathcal{B}$ in the DVL frame $\mathcal{D}$.

The residual vector, expressed in $\mathbb{R}^3$, is defined as:
\begin{equation}
    \mathbf{r}_\text{e} = \mathbf{h}_\text{e}\left(\mathbf{X}^j_{k}\right) - \mathbf{e}_k^j
    \label{eq:dvl_prior}
\end{equation}

\noindent where $\mathbf{e}_k^j$ is the DVL velocity measurement at keyframe $k$ of session $j$, expressed in $\mathcal{D}$.

\paragraph{Constant Velocity Factor}
\label{sec:constvel}
To ensure trajectory smoothness and propagate state estimates between keyframes, we incorporate a constant-velocity motion model. We make this assumption because the vehicles used for data collection traveled mostly at constant velocities. This factor places a constraint between two consecutive vehicle states, $\mathbf{X}^j_{k-1}$ and $\mathbf{X}^j_{k}$. The residual vector, $\mathbf{r}_\text{u}(\cdot)$, represents the deviation from the constant velocity assumption, expressed in the local body frame of the first state, $\mathbf{X}^j_{k-1}$.

\begin{equation}
    \begin{aligned}
    &\mathbf{r}_\text{u}\left(\mathbf{X}^j_{k-1}, \mathbf{X}^j_{k}\right) = \\
    &\quad\begin{bmatrix}
        \left(\mathbf{R}_{k-1}^j\right)^\top \left( \mathbf{t}^j_k - \left(\mathbf{t}^j_{k-1} + \mathbf{v}^j_{k-1} \Delta t \right)  \right) \\
        \left(\mathbf{R}_{k-1}^j\right)^\top \left(\mathbf{v}^j_{k} - \mathbf{v}^j_{k-1}\right)
    \end{bmatrix}
    \end{aligned}
    \label{eq:factor_const_vel_bodyframe}
\end{equation}

\noindent where the top block is the position error and the bottom block is the velocity error, both rotated into the body frame of state $\mathbf{X}_{k-1}^j$.

This residual is modeled as a zero-mean Gaussian, $\mathbf{r}_\text{u}(\cdot) \sim \mathcal{N}(\mathbf{0}, \boldsymbol{\Sigma}_\text{u})$. The $6 \times 6$ covariance matrix $\boldsymbol{\Sigma}_\text{u}$ is the exact discretization of a continuous white-noise-acceleration model \citep{barfoot2014batch} with isotropic power spectral density, $\mathbf{Q}_c = {q}\mathbf{I}_3$:

\begin{equation}
    \boldsymbol{\Sigma}_\text{u} =
    \begin{bmatrix}
        \tfrac{\Delta t^3}{3}\,\mathbf{Q}_c & \tfrac{\Delta t^2}{2}\,\mathbf{Q}_c \\[2pt]
        \tfrac{\Delta t^2}{2}\,\mathbf{Q}_c & \Delta t\,\mathbf{Q}_c
    \end{bmatrix},
    \label{eq:cv_cov}
\end{equation}

\noindent where $\Delta t$ is the time interval between states $\mathbf{X}^j_{k-1}$ and $\mathbf{X}^j_k$ and $q$ (in $\mathrm{m}^2/\mathrm{s}^3$) sets the expected rate of change of velocity.

\paragraph{Camera Projection Factor}
The camera factor relates a 3D landmark $\mathbf{p}_{i}$ to its 2D pixel observation $\mathbf{d}_{k,i,m}^j$. Here and in the SSS factor below, $\mathbf{p}_i$ is the world position of the landmark, obtained from its parameters $\mathbf{q}_i$ as described in Section~\ref{sec:landmark_param}. Its residual vector, $\mathbf{r}_{\text{CAM}}(\cdot)$, is defined as the difference between the observed and predicted pixel coordinates:

\begin{equation}
\begin{split}
    \mathbf{r}_{\text{CAM}}\Big( &\boldsymbol{\Delta}^j, \mathbf{X}_{k}^j, \mathbf{C}_m^j, \mathbf{p}_{i}, \mathbf{d}_{k,i,m}^j\Big) =  \\ 
    & h_{\text{CAM}}\left(\boldsymbol{\Delta}^j, \mathbf{X}_{k}^j, \mathbf{C}_m^j, \mathbf{p}_{i}\right) - \mathbf{d}_{k,i,m}^j
    \end{split}
    \label{eq:factor_cam_error}
\end{equation}
The projection function, $h_{\text{CAM}}(\cdot)$, first transforms the landmark from global coordinates into the local 3D Cartesian frame of the camera, ${}^{\mathcal{C}_m}\mathbf{p}_{i}$:
\begin{equation}
    {}^{\mathcal{C}_m}\mathbf{p}_{i} =  \left(\mathbf{C}_m^j\right)^{-1} \cdot \left(\mathbf{T}_{k}^j\right)^{-1} \cdot \left(\boldsymbol{\Delta}^j\right)^{-1} \cdot \mathbf{p}_{i}
    \label{eq:world_to_camera_transform_full}
\end{equation}
\noindent where ${}^{\mathcal{C}_m}\mathbf{p}_i = \begin{bmatrix} X & Y & Z\end{bmatrix}^\top$ is the landmark in the frame of camera $m$. This 3D point is then projected to pixel coordinates using the standard pinhole camera projection model:
\begin{equation}
    h_{\text{CAM}}(\cdot) = \begin{bmatrix} f_x \frac{X}{Z} + c_x \\[3pt] f_y \frac{Y}{Z} + c_y \end{bmatrix}
    \label{eq:pinhole_projection}
\end{equation}
\noindent where $f_x, f_y$ are the camera's focal lengths and $c_x, c_y$ are the principal point coordinates. The residual is modeled as a zero-mean Gaussian, $\mathbf{r}_{\text{CAM}}(\cdot) \sim \mathcal{N}(\mathbf{0}, \boldsymbol{\Sigma}_{\text{CAM}})$,
\noindent where $\boldsymbol{\Sigma}_{\text{CAM}}$ is the $2\times2$ measurement noise covariance, typically assumed to be diagonal:
\[
    \boldsymbol{\Sigma}_{\text{CAM}} = \begin{bmatrix} \sigma_u^2 & 0 \\ 0 & \sigma_v^2 \end{bmatrix}
\]
The terms $\sigma_u$ and $\sigma_v$ represent the standard deviation of the measurement uncertainty in the horizontal and vertical pixel directions, respectively.

\paragraph{Side-Scan Sonar Projection Factor}
\label{para: sss factor}
This factor relates a 3D landmark $\mathbf{p}_{i}$ to its 2D measurement $\mathbf{s}_{k,i}^j$. The residual vector is given by:
\begin{equation}
    \mathbf{r}_{\text{SSS}}\left(\boldsymbol{\Delta}^j, \mathbf{X}_{k}^j, \mathbf{p}_{i}, \mathbf{s}_{k,i}^j\right) = h_{\text{SSS}}\left(\boldsymbol{\Delta}^j, \mathbf{X}_{k}^j, \mathbf{p}_{i}\right) - \mathbf{s}_{k,i}^j
    \label{eq:factor_sss_error}
\end{equation}
The projection function, $h_\text{SSS}(\cdot)$, first transforms the landmark into the sonar's local reference frame:
\begin{equation}
    {}^{\mathcal{S}} \mathbf{p}_{i} =  \left({}^{\mathcal{B}}\mathbf{T}_{\mathcal{S}}\right)^{-1} \cdot \left(\mathbf{T}_{k}^j\right)^{-1} \cdot \left(\boldsymbol{\Delta}^j\right)^{-1} \cdot \mathbf{p}_{i} 
    \label{eq:world_to_sonar_full}
\end{equation}
\noindent where ${}^{\mathcal{S}}\mathbf{p}_i$ is the landmark in the sonar frame and ${}^{\mathcal{B}}\mathbf{T}_{\mathcal{S}} \in \mathrm{SE}(3)$ is the fixed extrinsic pose of the sonar frame $\mathcal{S}$ in the base frame $\mathcal{B}$.
From this 3D point, the function predicts the slant range and along-track distance:
\begin{equation}
    h_{\text{SSS}}(\cdot) = \begin{bmatrix} \| {}^S \mathbf{p}_{i} \| \\[1pt] ({}^S \mathbf{p}_{i})_x \end{bmatrix}
    \label{eq:model_sss}
\end{equation}
\noindent where $\| \cdot \|$ is the Euclidean norm, and $(\cdot)_x$ extracts the x-component of the vector. The residual vector is modeled as a zero-mean Gaussian, $\mathbf{r}_{\text{SSS}}(\cdot) \sim \mathcal{N}(\mathbf{0}, \boldsymbol{\Sigma}_{\text{SSS}})$, where $\boldsymbol{\Sigma}_{\text{SSS}}$ is the $2\times2$ measurement noise covariance, typically assumed to be diagonal:
\begin{equation}
    \boldsymbol{\Sigma}_{\text{SSS}} = \begin{bmatrix} \sigma_r^2 & 0 \\ 0 & \sigma_{at}^2 \end{bmatrix}
\end{equation}

The variance components are not fixed but are calculated adaptively for each measurement. The stochastic model is explicitly designed to translate the inherent feature location uncertainty from the 2D sonar waterfall image into a metric covariance. This image-space uncertainty is resolved into two orthogonal components:

\begin{itemize}
    \item Range Variance ($\sigma_r^2$): The variance for the range component quantifies the feature location uncertainty in the cross-track direction of the waterfall image. It is calculated by converting this pixel uncertainty into a metric variance using the sonar's known range resolution:
    \begin{equation}
        \sigma_r^2 = \left(\eta_r \cdot \rho_r\right)^2
    \end{equation}
    \noindent where $\eta_r$ is the noise parameter representing the standard deviation in range pixels, and $\rho_r$ is the sonar's range resolution (e.g., meters per pixel).

    \item Along-Track Distance Variance ($\sigma_{at}^2$): The variance for the zero along-track distance constraint is modeled as the sum of two independent error sources: the direct along-track uncertainty from feature localization and a geometric uncertainty from the sonar's physical beam width.
    \begin{equation}
        \sigma_{at}^2 = \left(\eta_a \cdot d_a\right)^2 + \left(R_k \cdot \theta_{bw}\right)^2
    \end{equation}
    The first term quantifies the direct along-track uncertainty, where $\eta_a$ is the standard deviation in pixels for the along-track direction and $d_a$ is the local along-track resolution (i.e., the average forward displacement of the vehicle per pixel row in the waterfall). The second term models the geometric uncertainty in the along-track direction caused by the sonar's horizontal beam width, $\theta_{bw}$. This uncertainty grows with the measured slant range, $R_k$, as the beam spreads.
\end{itemize}

\subsubsection{Landmark Parameterization}
\label{sec:landmark_param}
The camera and SSS projection models (Eqs.~\eqref{eq:pinhole_projection} and \eqref{eq:model_sss}) are only physically meaningful on one side of the sensor. A camera observes points in front of it, ${}^{\mathcal{C}_m}\mathbf{p}_i$ with $Z > 0$. A side-scan sonar observes points below it, on the side (port or starboard) of the channel in which the feature was detected. When the landmarks are parameterized directly by their world position $\mathbf{p}_i \in \mathbb{R}^3$, nothing prevents an iterate from placing a landmark on the wrong side of a sensor that observes it. Rejecting such configurations inside the projection factors makes the cost discontinuous, and the Levenberg-Marquardt step is then repeatedly rejected. Projecting regardless lets the optimizer converge to minima in which landmarks are mirrored through a sensor. A world-frame parameterization also couples the rigid displacement of a session to all of its landmarks. Moving the anchor $\boldsymbol{\Delta}^j$ while keeping the observations consistent requires every landmark observed in session $j$ to move with it, which the damping of the Levenberg-Marquardt step penalizes in proportion to the number of landmarks.

We therefore parameterize each landmark relative to a \emph{host}, the sensor pose of the first observation in its feature track, in coordinates that can only describe points on the valid side of that sensor. This follows the inverse-depth parameterization of monocular SLAM \citep{civera2008inverse, engel2017direct}, in which a point is stored as a bearing and an inverse depth relative to the camera in which it was first observed, and extends it to side-scan sonar. The world position of a landmark is a function of its parameters $\mathbf{q}_i$ and of the pose of its host:
\begin{equation}
    \mathbf{p}_i = \pi_i\left(\mathbf{H}_i, \mathbf{q}_i\right) = \mathbf{H}_i \cdot g_i\left(\mathbf{q}_i\right),
    \label{eq:hosted_world_point}
\end{equation}
\noindent where $\mathbf{H}_i \in \mathrm{SE}(3)$ is the world pose of the host sensor and $g_i(\cdot)$ maps the parameters to a point in the host sensor frame. The host pose is composed of existing variables. It is therefore not an additional variable, and it moves with its session's anchor. The landmarks are divided into two sets according to the observations in their feature tracks:
\begin{itemize}
    \item $\mathcal{L}_C$, \textbf{camera-hosted landmarks}: tracks with at least one camera observation, including multimodal tracks and tracks observed by more than one AUV session. The host is the first camera observation, made in session $j_h$ at keyframe $k_h$ by camera $m_h$.
    \item $\mathcal{L}_S$, \textbf{sonar-hosted landmarks}: tracks with only SSS observations. The host is the first sonar observation, made in session $j_h$ at keyframe $k_h$.
\end{itemize}

\paragraph{Camera-Hosted Landmarks}
A landmark $i \in \mathcal{L}_C$ is parameterized by its normalized image coordinates $(\alpha_i, \beta_i)$ in the host camera and the logarithm of its depth, $s_i$:
\begin{equation}
    \mathbf{q}_i = \begin{bmatrix} \alpha_i & \beta_i & s_i \end{bmatrix}^\top, \qquad
    g_i\left(\mathbf{q}_i\right) = e^{s_i} \begin{bmatrix} \alpha_i \\ \beta_i \\ 1 \end{bmatrix},
    \label{eq:cam_hosted_param}
\end{equation}
\noindent with host pose
\begin{equation}
    \mathbf{H}_i = \boldsymbol{\Delta}^{j_h}\,\mathbf{T}^{j_h}_{k_h}\,\mathbf{C}^{j_h}_{m_h}.
    \label{eq:cam_host_pose}
\end{equation}
\noindent The depth $e^{s_i}$ is positive for all real $s_i$, so the landmark stays in front of its host without any constraint on the optimization variable. The inverse depth $e^{-s_i}$ gives an equivalent parameterization. A step $\Delta s_i$ multiplies the depth by $e^{\Delta s_i}$, i.e.\ it is a relative change of depth, so steps and curvature in $s_i$ have the same scale for near and far landmarks.

\paragraph{Sonar-Hosted Landmarks}
The sonar frame has its $x$-axis pointing forward, its $y$-axis to starboard, and its $z$-axis pointing downwards. A landmark $i \in \mathcal{L}_S$ is parameterized by its along-track coordinate $x_i$, the logarithm $l_i$ of its distance from the sonar's $x$-axis, and an unconstrained elevation variable $\phi_i$:
\begin{equation}
    \mathbf{q}_i = \begin{bmatrix} x_i & l_i & \phi_i \end{bmatrix}^\top, \qquad
    g_i\left(\mathbf{q}_i\right) =
    \begin{bmatrix}
        x_i \\
        \varsigma_i\, e^{l_i} \cos\theta_i \\
        e^{l_i} \sin\theta_i
    \end{bmatrix},
    \label{eq:sss_hosted_param}
\end{equation}
\noindent with $\theta_i = \frac{\pi}{2}\,\mathrm{sig}(\phi_i)$, where $\mathrm{sig}(\phi) = 1/(1 + e^{-\phi})$ is the logistic function, $\theta_i \in (0, \pi/2)$ is the depression angle of the landmark below the horizontal plane of the sonar, and $\varsigma_i \in \{-1, +1\}$ is the side of the host observation. The side is fixed by the measurement and is not estimated. The host pose is
\begin{equation}
    \mathbf{H}_i = \boldsymbol{\Delta}^{j_h}\,\mathbf{T}^{j_h}_{k_h}\,{}^{\mathcal{B}}\mathbf{T}_S.
    \label{eq:sss_host_pose}
\end{equation}
\noindent For every value of $\mathbf{q}_i$, the landmark lies strictly below the host sonar and on the side on which it was observed, since $(g_i)_z = e^{l_i}\sin\theta_i > 0$ and $\varsigma_i (g_i)_y = e^{l_i}\cos\theta_i > 0$. The slant range is $\|g_i(\mathbf{q}_i)\| = \sqrt{x_i^2 + e^{2 l_i}}$.

\paragraph{Landmark Factors}
The camera and SSS projection factors (Eqs.~\eqref{eq:factor_cam_error} and \eqref{eq:factor_sss_error}) and their noise models are unchanged. Only their landmark argument changes: wherever $\mathbf{p}_i$ appears in them, and in the cost function below, it stands for $\pi_i(\mathbf{H}_i, \mathbf{q}_i)$ of Eq.~\eqref{eq:hosted_world_point}, and the factor depends on the variables composing $\mathbf{H}_i$ (Eq.~\eqref{eq:cam_host_pose} or \eqref{eq:sss_host_pose}) and on $\mathbf{q}_i$. For a given configuration, every factor therefore has the same residual as in the world-frame formulation. Two cases arise.

For the observation that defines the host, the transformation into the sensor frame (Eq.~\eqref{eq:world_to_camera_transform_full} or \eqref{eq:world_to_sonar_full}) cancels the host pose, and the factor is unary in $\mathbf{q}_i$. For a camera-hosted landmark, the camera model reduces to
\begin{equation}
    \mathbf{r}_{\text{CAM}}^{h}\left(\mathbf{q}_i, \mathbf{d}_{k_h,i,m_h}^{j_h}\right) =
    \begin{bmatrix} f_x\,\alpha_i + c_x \\ f_y\,\beta_i + c_y \end{bmatrix} - \mathbf{d}_{k_h,i,m_h}^{j_h},
    \label{eq:cam_host_factor}
\end{equation}
\noindent which constrains the bearing $(\alpha_i, \beta_i)$ and carries no information about the depth $s_i$. For a sonar-hosted landmark, the SSS model reduces to
\begin{equation}
    \mathbf{r}_{\text{SSS}}^{h}\left(\mathbf{q}_i, \mathbf{s}_{k_h,i}^{j_h}\right) =
    \begin{bmatrix} \sqrt{x_i^2 + e^{2 l_i}} \\ x_i \end{bmatrix} - \mathbf{s}_{k_h,i}^{j_h},
    \label{eq:sss_host_factor}
\end{equation}
\noindent which constrains $x_i$ and $l_i$ and carries no information about the elevation $\phi_i$.

Every other observation of the landmark constrains the parameter left free by its host observation: the depth of a camera-hosted landmark, or the elevation of a sonar-hosted landmark. Its factor connects the observing state, the host state, and the landmark parameters. SSS observations of a camera-hosted (multimodal) landmark are of this kind. When the observing and host states belong to the same session, the anchor $\boldsymbol{\Delta}^j$ appears in both the observing pose and the host pose and cancels, so the residual depends only on the relative pose of the two states. The anchors enter only through observations whose state is in a different session from the host, such as SSS observations of camera-hosted landmarks or camera observations from another AUV session.

The parameterization guarantees the valid side only for the host. A landmark may still lie behind a non-host camera or on the wrong side of a non-host ping. No explicit check or penalty is applied for such configurations. The projection models are evaluated as they stand, and the measurements of the non-host observations are left to correct them.

Since every residual equals its world-frame counterpart at $\mathbf{p}_i = \pi_i(\mathbf{H}_i, \mathbf{q}_i)$, the cost function is unchanged on the set of configurations in which every landmark lies on the valid side of its host. The parameterization changes the variables it minimizes and excludes the configurations in which a landmark is on the wrong side of its host. Two properties follow for the optimization. First, a rigid displacement of a session is a change of its anchor alone, because the landmarks hosted in that session move with $\mathbf{H}_i$. The damping of the Levenberg-Marquardt step therefore no longer penalizes it in proportion to the number of landmarks. Second, the direction that a single observation cannot determine, the depth for a camera or the elevation for a sonar, is a single scalar ($s_i$ or $\phi_i$), while the remaining two parameters are fixed directly by the host observation.

\subsubsection{Keyframe Selection}
\label{sub:keyframe_select}
To ensure computational tractability, a keyframe-based approach is employed rather than adding every vehicle state to the factor graph. A dense trajectory, particularly from a high-frequency sonar, can contain tens of thousands of poses, making the corresponding optimization problem prohibitively large. Therefore, the graph is constructed using only a sparse set of "key states" that are most informative.

A state is selected as a key state if it meets one of two criteria: (1) it has one or more landmark observations (either optical or sonar) associated with it, or (2) it is added at a fixed time interval. The first criterion ensures that all measurement constraints are anchored to a state in the graph, while the second criterion ensures a sufficient number of poses are included to accurately represent the trajectory's shape and enforce smoothness via the motion model.

\subsubsection{Non-Linear Least-Squares Cost Function}
Under the assumption that all noise sources are Gaussian, the MAP estimation problem is equivalent to solving a non-linear least-squares problem. This is achieved by minimizing the negative log-likelihood of the posterior probability. For a Gaussian distribution, this is equivalent to minimizing a sum of squared Mahalanobis distances.

The squared Mahalanobis distance for a residual vector $\mathbf{r}$ with a given covariance matrix $\mathbf{\Sigma}$ is defined as:
\begin{equation}
    \|\mathbf{r}\|^2_{\boldsymbol{\Sigma}} \triangleq \mathbf{r}^\top \boldsymbol{\Sigma}^{-1} \mathbf{r}
    \label{eq:mahalanobis_distance}
\end{equation}

We define the following index sets:
\begin{itemize}
    \item $\mathcal{J}_T$ and $\mathcal{J}_A$: the sets of towfish and AUV sessions, with $|\mathcal{J}_T| = N_T$, $|\mathcal{J}_A| = N_A$, and $\mathcal{J} = \mathcal{J}_T \cup \mathcal{J}_A$.
    \item $\mathcal{O}_{\text{SSS}} = \{(j, k, i)\}$: the set of sonar landmark observations, where $j \in \mathcal{J}_T$ is the session, $k$ is the keyframe index, and $i$ is the landmark index.

    \item $\mathcal{O}_{\text{CAM}} = \{(j, m, k, i)\}$: the set of camera landmark observations, where $j \in \mathcal{J}_A$, $m$ is the camera index, $k$ is the keyframe index, and $i$ is the landmark index.
\end{itemize}

\noindent Using the residual vectors defined in Section \ref{sec:factor_definitions}, the final objective function to be minimized is therefore expressed as the sum of the squared Mahalanobis distances of all factor errors. The MAP estimate, $\boldsymbol{\Theta}^* =$

{\small
\begin{align}
  &\underset{\boldsymbol{\Theta}}{\operatorname{argmin}} \Bigg(
    \sum_{j \in \mathcal{J}} \left(
      \left\|\mathbf{r}^j_{\Delta}\right\|^2_{\boldsymbol{\Sigma}_{\Delta}}
      + \left\|\mathbf{r}^j_{\text{X}_0}\right\|^2_{\boldsymbol{\Sigma}_{\text{X}_0}}
    \right)
    + \underbrace{
      \sum_{j \in \mathcal{J}_A} \sum_m
        \left\|\mathbf{r}^j_{\text{C},m}\right\|^2_{\boldsymbol{\Sigma}_\text{C}}
    }_{\text{Cam.\ Extrinsics}}
    \nonumber \\
  &+ \underbrace{
      \sum_{j \in \mathcal{J}} \sum_k
        \left\|r^j_{\text{d},k}\right\|^2_{\Sigma_\text{d}}
    }_{\text{Depth}} + \underbrace{
      \sum_{j \in \mathcal{J}} \sum_k
        \left\|\mathbf{r}^j_{\text{a},k}\right\|^2_{\boldsymbol{\Sigma}_\text{a}}
    }_{\text{Attitude}} + \underbrace{
      \sum_{j \in \mathcal{J}} \sum_{k \geq 1}
        \left\|r^j_{\psi,k}\right\|^2_{\Sigma_\psi}
    }_{\text{Relative Heading}}
    \nonumber \\
  &+ \underbrace{
      \sum_{j \in \mathcal{J}_T} \sum_{k \geq 1}
        \left\|\mathbf{r}^j_{xy,k}\right\|^2_{\boldsymbol{\Sigma}_{xy}}
    }_{\text{Towfish Rel.\ Horiz.\ Position}} + \underbrace{
      \sum_{j \in \mathcal{J}_A} \sum_k
        \left\|\mathbf{r}^j_{\text{e},k}\right\|^2_{\boldsymbol{\Sigma}_\text{e}}
    }_{\text{AUV DVL}}
    \nonumber \\
  &+ \underbrace{
      \sum_{j \in \mathcal{J}} \sum_k
        \left\|\mathbf{r}^j_{\text{u},k}\right\|^2_{\boldsymbol{\Sigma}_\text{u}}
    }_{\text{Constant Velocity Motion Model}} + \underbrace{
      \sum_{(j,\,k,\,i)\,\in\,\mathcal{O}_{\text{SSS}}}
        \left\|\mathbf{r}^j_{\text{SSS},k,i}\right\|^2_{\boldsymbol{\Sigma}_{\text{SSS}}}
    }_{\text{SSS Landmark Measurements}}
    \nonumber \\
  &+ \underbrace{
      \sum_{(j,\,m,\,k,\,i)\,\in\,\mathcal{O}_{\text{CAM}}}
        \left\|\mathbf{r}^j_{\text{CAM},k,i,m}\right\|^2_{\boldsymbol{\Sigma}_{\text{CAM}}}
    }_{\text{Camera Landmark Measurements}}
  \Bigg)
  \label{eq:least_squares_final}
\end{align}
}
\noindent where each residual is written by its indices, with its arguments as given in its factor definition; for example, $r^j_{\text{d},k} \triangleq r_\text{d}(\boldsymbol{\Delta}^j, \mathbf{X}_k^j, {}^{\mathcal{W}}\bar{\mathbf{T}}^{j}_{k})$, $\mathbf{r}^j_{\text{u},k} \triangleq \mathbf{r}_\text{u}(\mathbf{X}_{k-1}^j, \mathbf{X}_k^j)$ and $\mathbf{r}^j_{\text{CAM},k,i,m} \triangleq \mathbf{r}_\text{CAM}(\boldsymbol{\Delta}^j, \mathbf{X}_k^j, \mathbf{C}_m^j, \mathbf{p}_i, \mathbf{d}^j_{k,i,m})$, with $\mathbf{p}_i = \pi_i(\mathbf{H}_i, \mathbf{q}_i)$ (Eq.~\eqref{eq:hosted_world_point}). For the host observation of a hosted landmark, the camera and SSS terms take the unary forms of Eqs.~\eqref{eq:cam_host_factor} and \eqref{eq:sss_host_factor}.

Feature matching between optical images is performed automatically and is therefore susceptible to outliers arising from incorrect data associations. To ensure these outliers do not corrupt the optimization result, a robust cost function is applied to the camera projection factors. The standard squared-residual term for camera measurements in Equation \ref{eq:least_squares_final} is replaced by the Huber loss, $\rho(\cdot)$:
\begin{equation}
    \sum_{(j,\, m,\, k,\, i)\, \in\, \mathcal{O}_{\text{CAM}}} \rho\left( \left\|\mathbf{r}_{\text{CAM}}\left(\boldsymbol{\Delta}^j, \mathbf{X}_{k}^j, \mathbf{C}_m^j, \mathbf{p}_{i}, \mathbf{d}_{k,i,m}^j\right)\right\|^2_{\boldsymbol{\Sigma}_{\text{CAM}}}\right)
\end{equation}

\noindent We solve the non-linear least-squares problem defined by Equation \ref{eq:least_squares_final} using the Georgia Tech Smoothing and Mapping (GTSAM) library \citep{gtsam}, which implements efficient factor graph-based optimization using the Levenberg-Marquardt algorithm.

\subsubsection{Initial Estimates}
\label{sec:initial_estimates}
Non-linear least-squares solvers like the Levenberg-Marquardt algorithm require initial estimates for all variables in the factor graph. A good initial guess helps the algorithm converge faster and more reliably to the global or a meaningful local minimum. The initialization strategies for each variable type are detailed below. 
\begin{itemize} 
\item {Vehicle State ($\mathbf{X}_k^j$):} The trajectory for each session is initialized in its own local reference frame. The first state, $\mathbf{X}_0^j$, is set to the identity pose. Subsequent states, $\mathbf{X}_k^j$, are then computed relative to this origin using the raw navigation data. The initial velocity ($\mathbf{v}_k^j$) for each state is estimated via finite differencing of the relative position components between consecutive poses. 
\item Anchors ($\boldsymbol{\Delta}^j$): The initial estimate for each session anchor, $\boldsymbol{\Delta}^j$, is set to the first georeferenced pose provided by the navigation data for that session.
\item Camera Extrinsics ($\mathbf{C}_m^j$): Although the precise extrinsic calibration is unknown, an initial estimate for each camera pose, $\mathbf{C}_m^j$, is available. This estimate is used to initialize the variable and to define a relatively tight prior. This constrains the optimization and prevents convergence to physically implausible values. 
\item Landmarks ($\mathbf{q}_i$): Each landmark is first initialized in the world frame. The initialization strategy for a 3D landmark depends on the type of observations in its feature track: 
\begin{itemize} 
\item Sonar-Only Tracks: For tracks containing only SSS observations, each measurement is projected onto a plane representing the seafloor, using the initial vehicle pose and altitude. The ground range $\sqrt{r^2 - a^2}$, where $r$ is the slant range and $a$ is the altitude, is placed perpendicular to the vehicle heading on the side (port or starboard) where the feature was observed.
The initial landmark position is then set to the centroid (average) of these projected points. 
\item Camera-Only Tracks: Triangulation is used to obtain the initial 3D position from multiple camera observations. 
\item Multimodal Tracks: Each camera observation is back-projected as a 3D ray from the initial camera pose and intersected with a horizontal plane located at the vehicle's altitude when the image was captured. The initial landmark position is the average of these intersection points. The sonar observations are not used for this initialization. Most multimodal tracks contain a single camera view, and before the sessions are aligned, the sonar-based estimate can lie several meters away from the camera ray. This results in large initial camera residuals and adversely affects the convergence of the optimization.
\item Special Cases: If no camera ray of a multimodal track intersects the plane and the track has at least two camera views, the camera-only triangulation is combined with the sonar-only position estimate in a weighted average. The weights are proportional to the number of observations from each modality within the track. If triangulation is not possible, the sonar-only estimate is used. For camera-only tracks, if triangulation fails (for example, because of a chirality violation, where the reconstructed point lies behind one or more of the cameras observing it), the landmark is initialized with the same ray–plane intersection used for multimodal tracks.

\end{itemize}

The resulting world point $\mathbf{p}_i^0$ is transformed into the frame of its host using the initial estimates of the host variables, $\mathbf{u}_i = (\mathbf{H}_i^0)^{-1}\,\mathbf{p}_i^0 = [u_x\;u_y\;u_z]^\top$, and converted to the host parameters:
\begin{itemize}
    \item \textbf{Camera-hosted:} $\alpha_i = u_x / u_z$, $\beta_i = u_y / u_z$, $s_i = \ln u_z$. If the initial point lies behind or within 0.1~m of the host camera ($u_z \leq 0.1$~m), it is replaced by the point on the ray of the host pixel $\mathbf{d}^{j_h}_{k_h,i,m_h} = [u\;v]^\top$ at depth $\max(a, 1\,\mathrm{m})$, where $a$ is the vehicle altitude at the host image: $\alpha_i = (u - c_x)/f_x$, $\beta_i = (v - c_y)/f_y$, $s_i = \ln\max(a, 1\,\mathrm{m})$.
    \item \textbf{Sonar-hosted:}
    \begin{equation}
    \begin{aligned}
        x_i &= u_x, \\
        l_i &= \ln\max\left(\sqrt{u_y^2 + u_z^2},\, 0.1\,\mathrm{m}\right), \\
        \phi_i &= \ln\frac{\theta_i^0}{\pi/2 - \theta_i^0}, \\
        \theta_i^0 &= \operatorname{clip}\left(\operatorname{atan2}(u_z, \varsigma_i u_y),\, \theta_{\min},\, \tfrac{\pi}{2} - \theta_{\min}\right)
    \end{aligned}
    \label{eq:sss_hosted_init}
    \end{equation}
    \noindent with $\theta_{\min} = 0.02$~rad. The clipping moves an initial point above the sonar or on the unobserved side to the nearest valid elevation. It also keeps $\phi_i$ away from the saturated tails of the logistic function, where $\partial\theta_i/\partial\phi_ i \to 0$, and the elevation would barely respond to the optimization.
\end{itemize}

\end{itemize} 

\subsection{Post-Processing and Map Generation}
\label{sec:post_processing}
Once the optimization has converged, a series of post-processing steps is performed to generate the final dense trajectories and map products.

\paragraph{Dense Trajectory Interpolation}

The optimization yields a sparse set of poses corresponding to key states. Intermediate poses are then estimated via interpolation to generate a continuous, dense trajectory. This process uses the underlying geometry by treating poses as elements of the Special Euclidean group, $\mathrm{SE}(3)$.

Interpolation is performed along the geodesic path on the manifold, which represents the shortest constant-velocity motion between two poses. For any two consecutive poses, $\mathbf{T}_1$ and $\mathbf{T}_2$, the relative transformation between them is first mapped to the Lie algebra $\mathfrak{se}(3)$ via the logarithmic map. This yields a single vector representing the entire rigid motion. This vector is then scaled according to the desired interpolation timestamp and mapped back from the Lie algebra to the $\mathrm{SE}(3)$ manifold via the exponential map. The resulting transformation is applied to the first pose to obtain the final interpolated pose, ensuring a smooth and physically plausible trajectory.

\paragraph{Map Generation}
\label{sec: map gen}
The final, dense, globally-aligned trajectories are used to generate georeferenced maps for each sensor modality.
\begin{itemize}
    \item \textbf{Optical Mosaic:} The optical images and their corresponding optimized global poses are processed using MosaicViewer \citep{escartin2008globally} to generate a high-resolution, georeferenced optical map of the surveyed area.
    \item \textbf{Sonar Mosaic:} The side-scan sonar data is processed using MB-System \citep{Caress2024MB-System} to produce a georeferenced GeoTIFF image of the acoustic backscatter.
\end{itemize}
\section{Experimental Results \& Discussion}
\label{sec:results}

\subsection{Experimental Setup}
\subsubsection{Dataset Description}
\begin{figure}
\centering

\subfloat[Girona 1000 AUV]{%
    \includegraphics[width=0.48\linewidth]{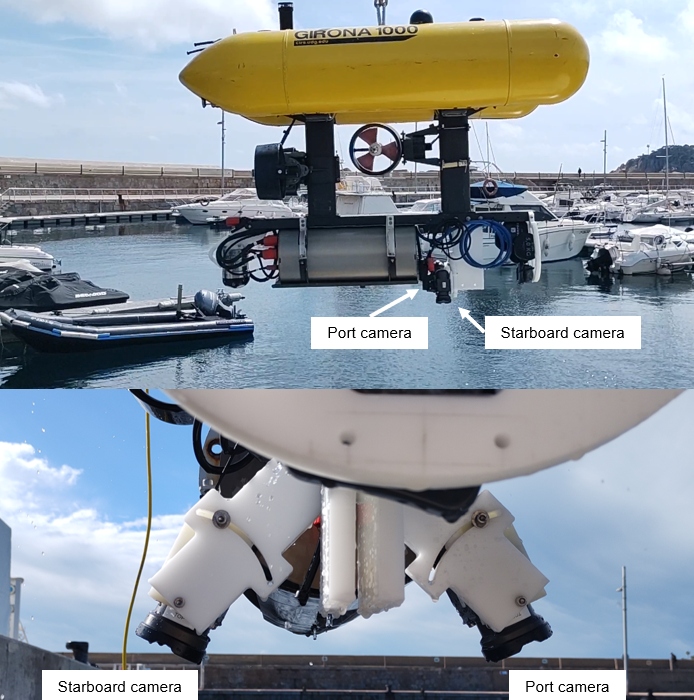}
    \label{fig:girona1000}
}
\hfill
\subfloat[Surface vessel and Klein System 3000 SSS]{%
    \includegraphics[width=0.48\linewidth]{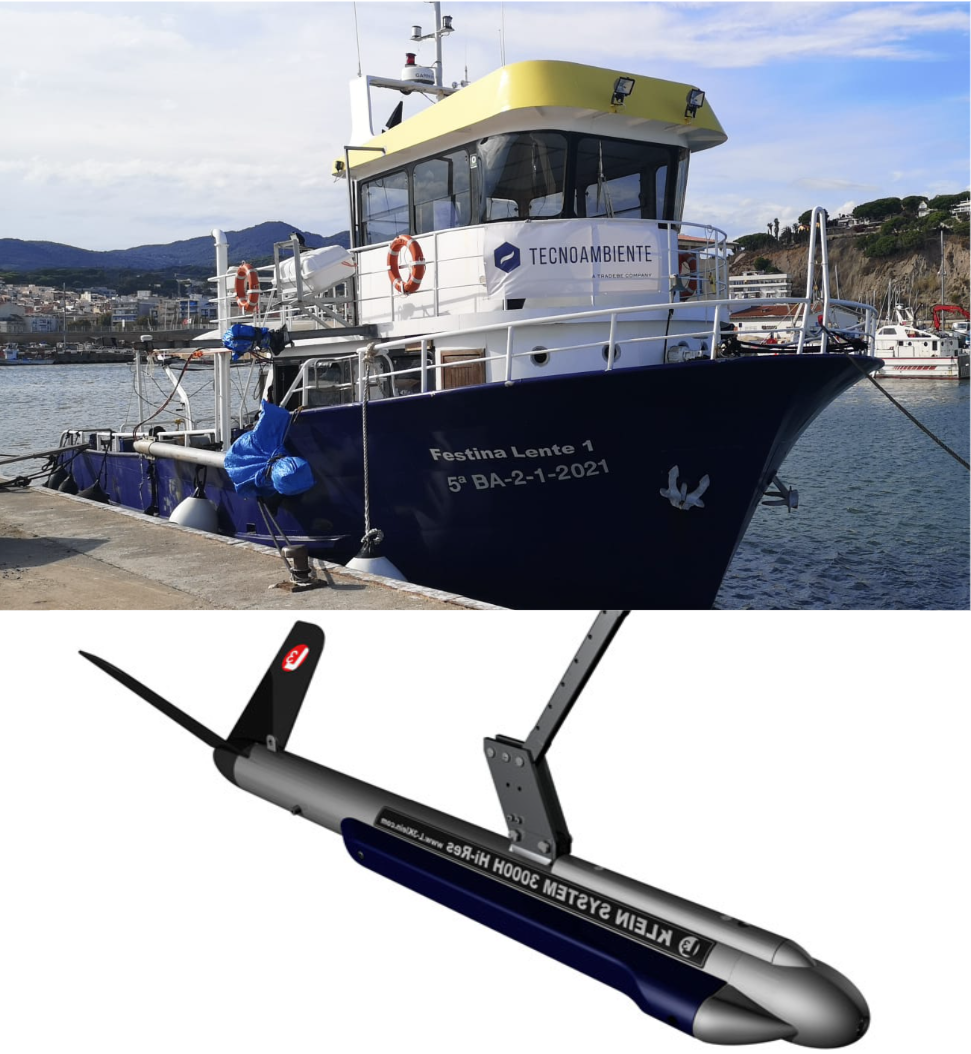}
    \label{fig:klein3000}
}

\caption{Vehicles used for data collection}
\label{fig:veh_data}

\end{figure}

We tested the proposed methodology on optical and side-scan sonar data of the seabed collected during surveys along the Catalan coast. 
The optical data were obtained using the Girona 1000 AUV, a variant of the Girona500 AUV \citep{ribas2011girona}, equipped with down-looking cameras (Figure \ref{fig:girona1000}). In one of the optical sessions, two down-looking cameras were used for data collection. The cameras were mounted in a divergent stereo configuration with an angle of approximately \ang{60} between the cameras' axes. The field of view of the cameras allows for an overlap of approximately \SI{20}{\percent} between the two cameras. In all the other optical sessions, we used a single camera for data collection. The sonar data were collected by Tecnoambiente S.L., an environmental consultancy company, using a Klein System 3000 Side Scan Sonar \citep{klein3000sss, rajani2026benthicat} deployed from a survey vessel (Figure \ref{fig:klein3000}). The localization system of the vehicles is discussed in Section \ref{sec:system_overview}.

Eight sessions, comprising four sonar sessions and four optical sessions, from surveys conducted in the same geographical area, were used to evaluate the proposed framework. Henceforth, the four sonar sessions will be referred to as \textit{SON1}, \textit{SON2}, \textit{SON3}, and \textit{SON4}, respectively, and the optical sessions will be referred to as \textit{CAM1}, \textit{CAM2}, \textit{CAM3}, and \textit{CAM4}. The sonar sessions and \textit{CAM1} are from the survey described in the BenthiCat dataset \citep{rajani2026benthicat}.

\begin{table}[t]
\centering
\caption{Sonar dataset characteristics}
\label{tab:sonar_data}
\resizebox{\columnwidth}{!}{
\begin{tabular}{lcccc}
\toprule
 & SON1 & SON2 & SON3 & SON4 \\
\midrule
Total pings & 13400 & 8695 & 8633 & 8815 \\
No. of bins per ping & 4096 & 4096 & 4096 & 4096\\
Slant range (m) & 100 & 75 & 75 & 75 \\
Month-Year & Oct 2021 & Oct 2021 & Oct 2021 & Oct 2021   \\
Output frequency (Hz) & 7.46 & 9.92 & 9.92 & 9.92 \\
Horizontal beam width & $0.2^\circ$ & $0.2^\circ$ & $0.2^\circ$ & $0.2^\circ$ \\
\bottomrule
\end{tabular}
}
\end{table}

\begin{table}[t]
\centering
\caption{Optical dataset characteristics}
\label{tab:optical_data}
\resizebox{\columnwidth}{!}{
\begin{tabular}{lcccc}
\toprule
 & CAM1 & CAM2 & CAM3 & CAM4 \\
\midrule
No. of frames & 4580 & 9340 & 1156 & 7043 \\
No. of cameras used & 2 & 1 & 1 & 1 \\
Camera frame rate (Hz) & 3 & 4 & 4 & 4 \\
DVL and IMU available & No & Yes & Yes & Yes \\
Month-Year & Apr. 2024 & Jul 2025 & Jul 2025 & Jul 2025 \\
\bottomrule
\end{tabular}
}
\end{table}

Tables \ref{tab:sonar_data}  and \ref{tab:optical_data} summarize the characteristics of the sonar and optical data, respectively. The time difference between the optical and sonar data collection efforts is relatively large and was due to vehicle unavailability and other logistics. It is expected that some changes may have occurred during this period.
However, the seafloor structures that were manually matched across the two modalities are quite distinctive and have similar geometric appearances (taking into account the distinct imaging characteristics of the two sensors).

\subsubsection{Feature Matching}
\begin{figure}
\centering
    \includegraphics[width=\linewidth]{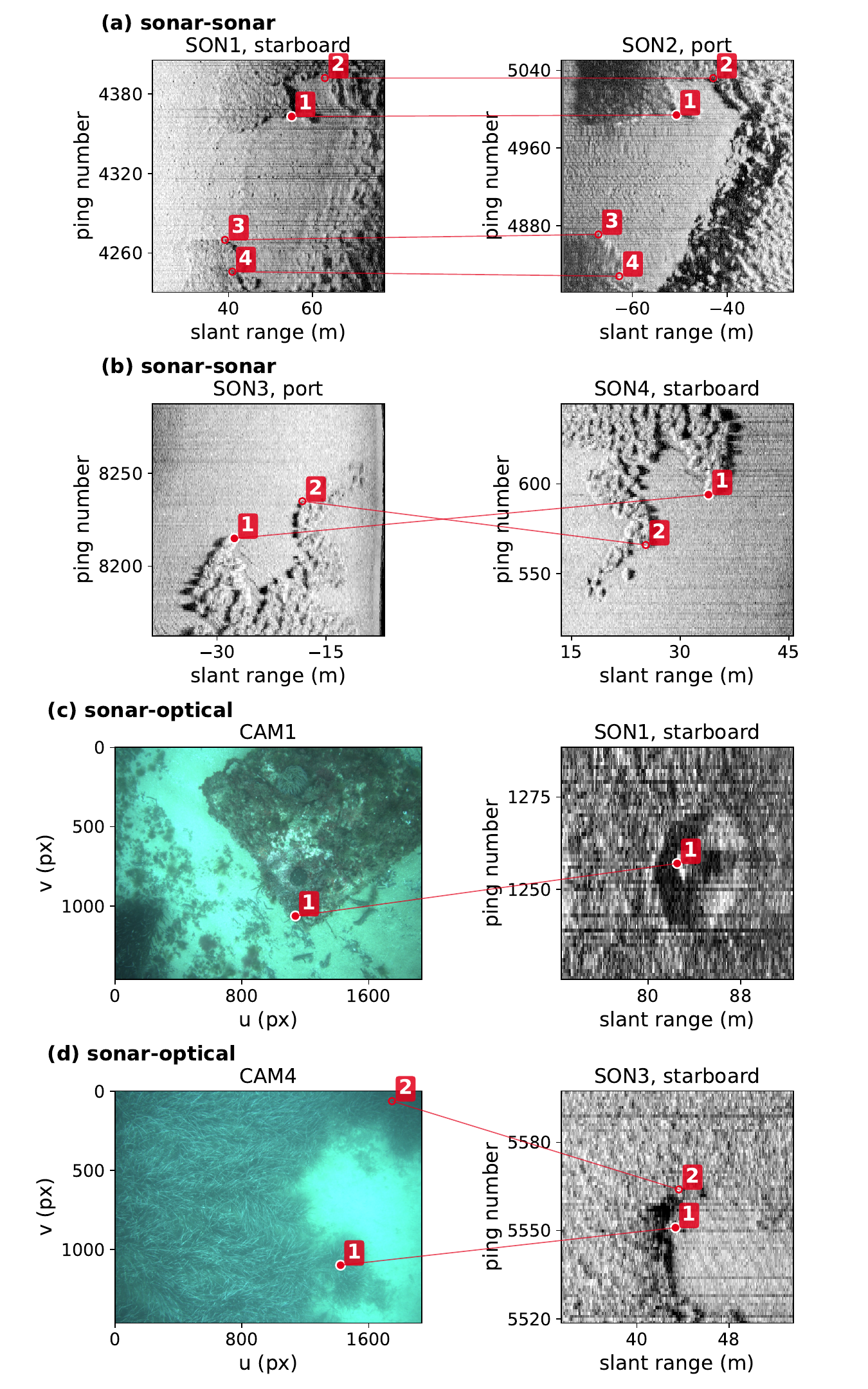} %
    \caption{Examples of manually annotated feature correspondences.}
    \label{fig:feat_corr}

\end{figure}

The distribution of unimodal and multimodal matches used for the optimization is shown in Table \ref{tab:num_matches}. As shown, the number of optical-optical matches is significantly larger than the other types. This is due to the use of SIFT for automatic feature detection and matching, which produced a large number of correspondences despite the presence of repetitive seabed structures such as seagrass leaves. In contrast, the number of matches between sonar and optical images is comparatively small, which can be attributed to the difficulty in observing unique and distinguishable matches between the two modalities by the human annotator. Matches were mainly made along seabed contours and visible landmarks (such as on distinctive transitions from seagrass to sand and on salient objects such as cement blocks used for boat anchoring). Figure \ref{fig:feat_corr} shows examples of feature matching between two sonar sessions and between sonar and optical sessions.

The distribution of unimodal (sonar-only and optical-only) and multimodal feature tracks generated from these matches is reported in Table \ref{tab:num_track_types}.

\begin{table}[t]
\centering
\caption{Number of matches used for optimization}
\label{tab:num_matches}

\begin{tabular}{lc}
\toprule
Match Type & Number of Matches \\
\midrule
Sonar--Sonar & 68 \\
Sonar--Optical & 66 \\
Optical--Optical & 30705 \\
\bottomrule
\end{tabular}

\end{table}

\begin{table}[t]
\centering
\caption{Distribution of track types}
\label{tab:num_track_types}

\begin{tabular}{lc}
\toprule
Track Type & Number of Tracks \\
\midrule
Sonar-only & 64 \\
Optical-only & 14271 \\
Multimodal & 50 \\
\bottomrule
\end{tabular}

\end{table}

\subsubsection{Implementation Details}
The $N$ parameter, defined in Section \ref{sub:feature_track_filtering}, is set to 4. The fixed time interval chosen for keyframe selection, as described in Section \ref{sub:keyframe_select}, is 1 second. The noise parameters of all factors are listed in Table~\ref{tab:noise_params}. Since the dataset provides no uncertainty estimates for the towfish navigation, and the positions were filtered, their errors are strongly time-correlated, and their distribution is unknown. We therefore set $\boldsymbol{\Sigma}_{xy}$ conservatively, which prevents the per-keyframe priors from overweighting correlated navigation errors.

\begin{table}[t]
\centering
\caption{Noise parameters (standard deviations) used in all experiments.}
\label{tab:noise_params}
\footnotesize
\setlength{\tabcolsep}{3pt}
\begin{tabularx}{\columnwidth}{@{}l l >{\raggedright\arraybackslash}X@{}}
\toprule
Factor & Symbol & Value \\
\midrule
\multicolumn{3}{@{}l}{\textit{Priors}} \\
Anchor, SON1 (reference) & $\boldsymbol{\Sigma}_{\Delta}$ & $(x, y)$: $10^{-4}$\,m; yaw: $10^{-4}$\,rad \\
Anchor, other sessions & $\boldsymbol{\Sigma}_{\Delta}$ & $(x, y)$: $20$\,m; yaw: $0.01$\,rad \\
First state, pose & $\boldsymbol{\Sigma}_{\text{X}_0}$ & $10^{-5}$ rad, m (all axes) \\
First state, velocity & $\boldsymbol{\Sigma}_{\text{X}_0}$ & towfish: $0.2$\,m\,s$^{-1}$; AUV with DVL: $10^{-3}$\,m\,s$^{-1}$; AUV without DVL: $0.1$\,m\,s$^{-1}$ \\
Camera extrinsics & $\boldsymbol{\Sigma}_{\text{C}}$ & $0.05$\,rad, $0.05$\,m per axis \\
\midrule
\multicolumn{3}{@{}l}{\textit{Navigation, towfish}} \\
Depth & $\Sigma_{\text{d}}$ & $0.1$\,m \\
Attitude & $\boldsymbol{\Sigma}_{\text{a}}$ & $0.0035$\,rad ($0.2^\circ$) \\
Relative heading & $\Sigma_{\psi}$ & $0.0087$\,rad ($0.5^\circ$) \\
Relative horizontal position & $\boldsymbol{\Sigma}_{xy}$ & $7.5$\,m per axis \\
\midrule
\multicolumn{3}{@{}l}{\textit{Navigation, AUV}} \\
Depth & $\Sigma_{\text{d}}$ & $0.5$\,m \\
Attitude & $\boldsymbol{\Sigma}_{\text{a}}$ & $0.01$\,rad ($0.57^\circ$) \\
Relative heading & $\Sigma_{\psi}$ & $0.01$\,rad ($0.57^\circ$) \\
DVL & $\boldsymbol{\Sigma}_{\text{e}}$ & $0.1$\,m\,s$^{-1}$ per axis \\
\midrule
\multicolumn{3}{@{}l}{\textit{Motion model} ($\mathbf{Q}_c = q\,\mathbf{I}_3$)} \\
Towfish and AUV & $\sqrt{q}$ & $0.05$\,m\,s$^{-3/2}$ \\
\midrule
\multicolumn{3}{@{}l}{\textit{Sonar measurements} ($\boldsymbol{\Sigma}_{\text{SSS}}$)} \\
Sonar-only tracks & $\eta_r, \eta_a$ & $2, 2$\,px \\
Cross-modal tracks & $\eta_r, \eta_a$ & $5, 5$\,px \\
\midrule
\multicolumn{3}{@{}l}{\textit{Camera measurements} ($\boldsymbol{\Sigma}_{\text{CAM}}$)} \\
Optical-only tracks & $\sigma_u, \sigma_v$ & $2, 2$\,px; Huber $k = 5$ \\
Cross-modal tracks & $\sigma_u, \sigma_v$ & $2, 2$\,px; no robust kernel \\
\bottomrule
\end{tabularx}
\end{table}

\subsection{Evaluation}
In the absence of ground-truth trajectory data and multimodal maps, we evaluate the proposed approach using a combination of internal consistency metrics, the alignment accuracy of semantically segmented maps generated from the estimated session trajectories, and qualitative inspection. To further validate the proposed approach, we compare its performance against a baseline that enforces rigid inter-session transformations.

\subsubsection{Session Trajectories}
\begin{figure*}[tbh]
    \centering
    \includegraphics[width=\linewidth]{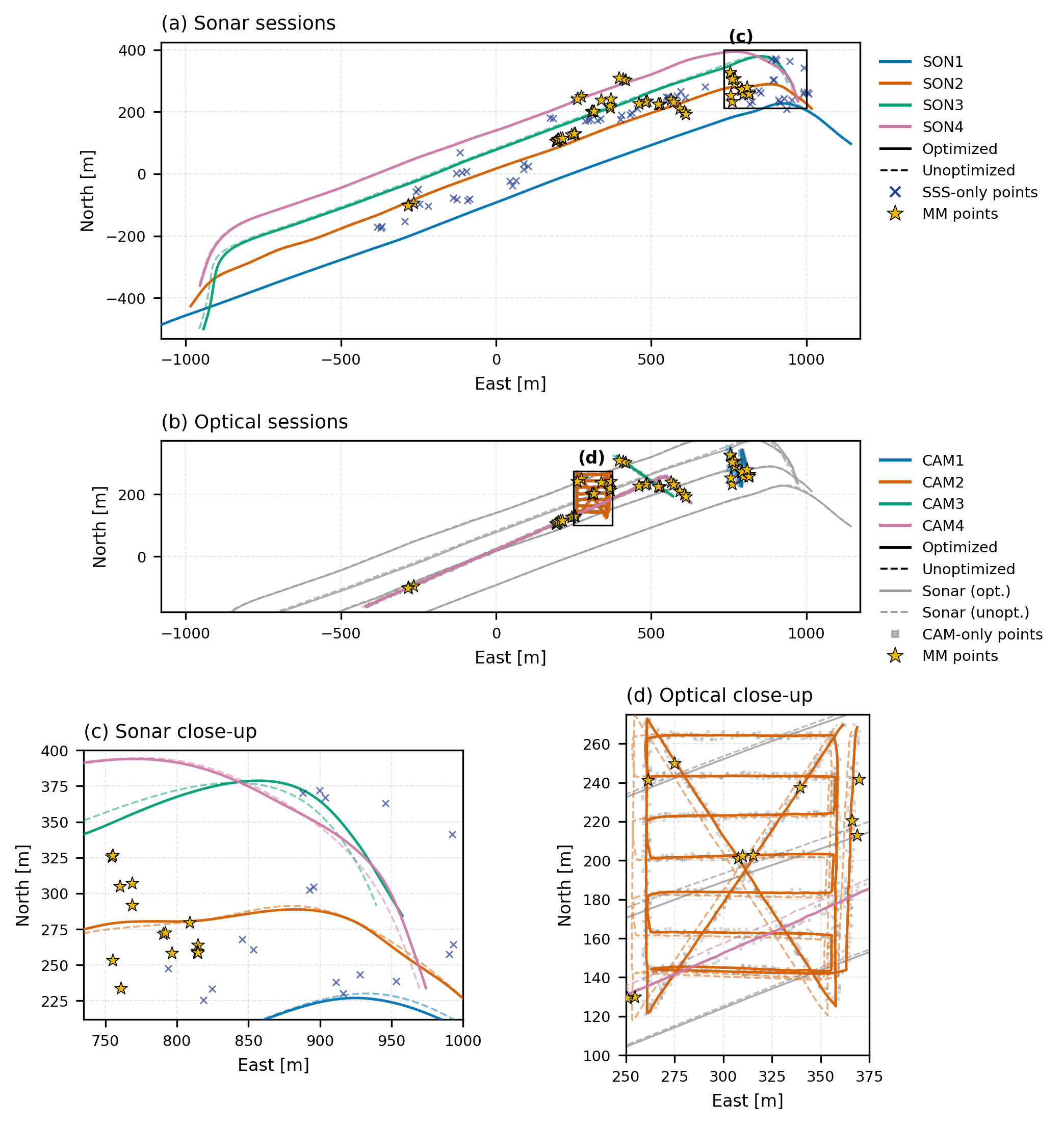} 
    \caption{Session trajectories before (dashed lines) and after (solid lines) optimization. (a) Sonar sessions with sonar-only (SSS-only) and multimodal (MM) points. (b) Optical sessions with optical-only (CAM-only) and MM points; the sonar trajectories are shown in gray for reference. (c) Close-up of the sonar trajectories in the region surveyed by \textit{CAM1}. (d) Close-up of the \textit{CAM2} survey. The close-up regions are marked by black boxes in (a) and (b).}
    \label{fig:session_traj}
\end{figure*}

The trajectories taken by the vehicles before and after optimization are shown in Figure \ref{fig:session_traj}. At survey scale, the optimized trajectories remain close to the unoptimized estimates, indicating that the onboard navigation was reasonably accurate. The corrections become visible at a finer scale. Local deformations are visible in the sonar trajectories near the multimodal points (Figure \ref{fig:session_traj}c), notably along \textit{SON2} where it intersects \textit{CAM1}, and at the turns of \textit{SON3} and \textit{SON4}. Most of these local deformations are smooth along the trajectory rather than abrupt jumps between consecutive poses, which is consistent with the role of the constant velocity factor in keeping the estimated motion physically plausible. The optical trajectories are also corrected (Figure \ref{fig:session_traj}d).

Figure \ref{fig:session_traj} also shows the optimized points observed only in sonar sessions, only in optical sessions, and those observed in both modalities. Points observed only in optical sessions are densely distributed along the optical trajectories due to the large number of automatically detected feature matches. In contrast, points observed only in sonar sessions are more sparsely distributed but still provide sufficient spatial coverage to support the global alignment. Because the sensing range of optical cameras is significantly smaller than that of the side-scan sonar, multimodal points tend to lie close to the camera poses from which they were observed.

\subsubsection{Reprojection Error}
\label{sec:reproj}
\begin{figure}
    \centering
    \includegraphics[width=\linewidth]{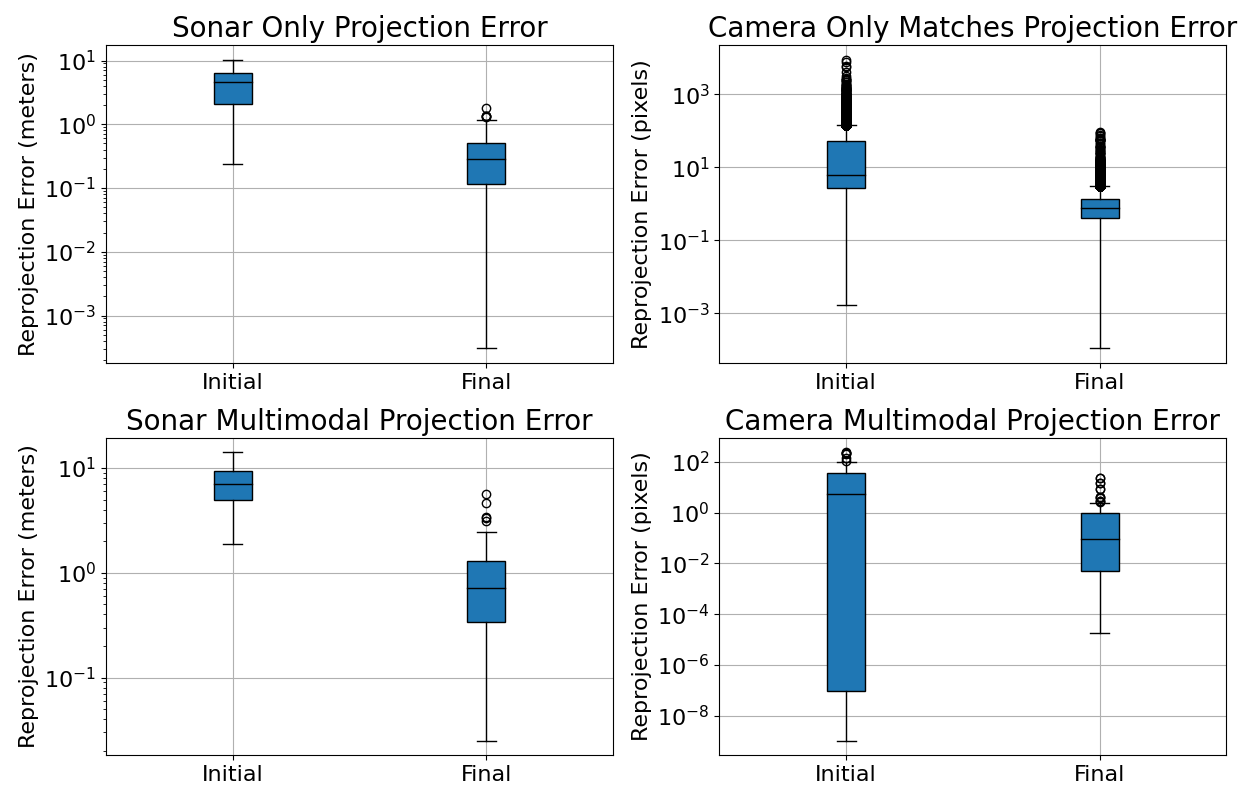}
    \caption{Boxplots comparing the initial and final scalar projection errors, defined as the Euclidean distance between observed and projected measurements. The vertical axis in the plots is represented on a logarithmic scale. }
    \label{fig:boxplot_err}
\end{figure}

The boxplots in Figure \ref{fig:boxplot_err} show the reprojection error magnitudes for the three distinct categories of feature tracks. The reprojection errors for multimodal feature tracks are further divided into sonar and optical components for subsequent analysis. A clear reduction in both the median error and the interquartile range is evident across all categories. The initial errors of the camera observations in multimodal tracks are close to zero for about half of the observations, because each multimodal landmark is initialized on the viewing ray of one of its camera observations. The wide initial box of this category reflects this initialization rather than the initial alignment.

The boxplot of reprojection error for the camera-only tracks shows several outliers, despite the low median error after optimization. Further investigation indicates that most of these outliers are due to inaccurate feature matches, and only a few are caused by specific 3D landmarks that did not converge. The Huber loss helps to ensure that the optimization is not strongly affected by these inaccuracies.

\begin{figure}
    \centering

    \subfloat[Sonar reprojection error]{%
    \label{fig:sonar_reproj_scatter}
    \includegraphics[width=0.45\linewidth]{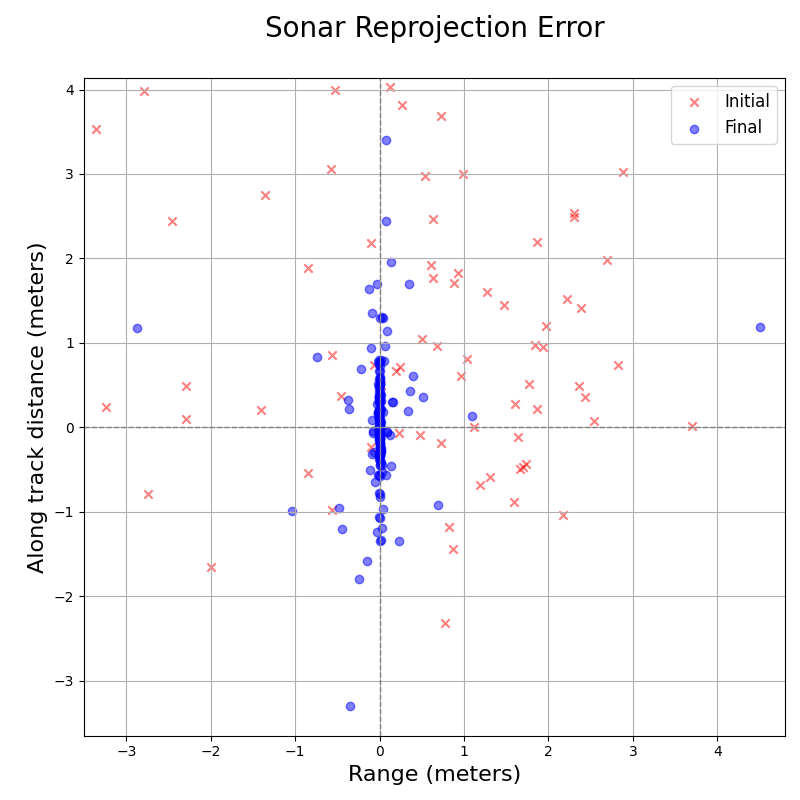}
}
\hfill
\subfloat[Camera reprojection error]{%
    \label{fig:cam_reproj_scatter}
    \includegraphics[width=0.45\linewidth]{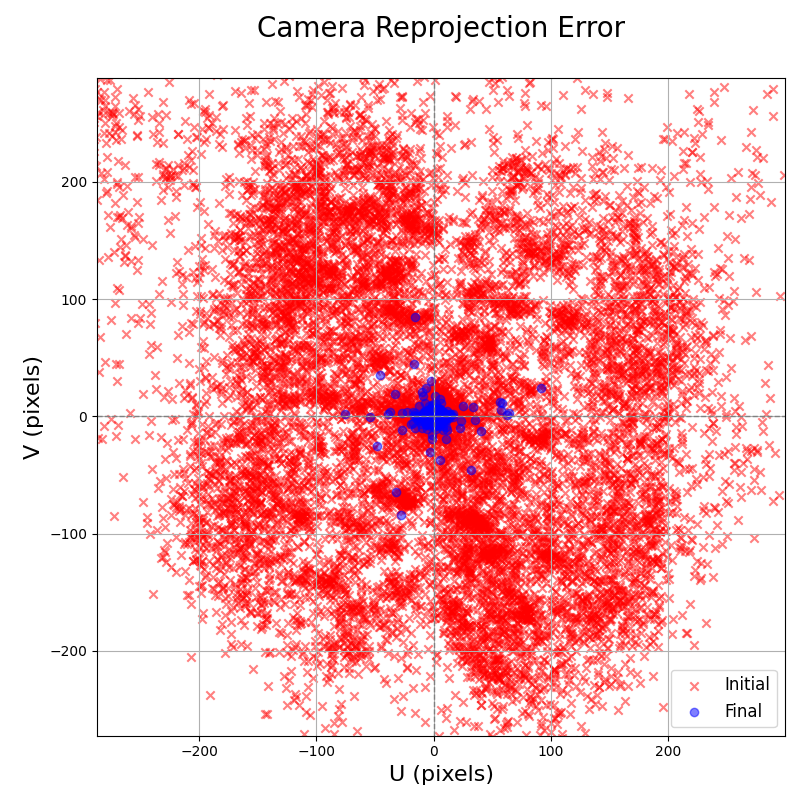}
}

    \caption{Scatter plots of the initial and final projection errors for sonar and camera observations. For visual clarity, the plot axes have been focused on the main error distribution, and extreme outliers are not shown.}
    \label{fig:reproj_scatter}
\end{figure}

The 2D error distributions, visualized in the scatter plots of Figure \ref{fig:reproj_scatter}, give additional detail. For the sonar projection error (Figure \ref{fig:sonar_reproj_scatter}), the final range errors are concentrated near zero, whereas the along-track errors are larger. The remaining sonar error is therefore mainly along-track, the direction in which the towfish position is least constrained. The corresponding scatter plot for the camera reprojection error (Figure \ref{fig:cam_reproj_scatter}) confirms the presence of the previously discussed outliers. Despite these outliers, the plot clearly shows that the vast majority of errors lie within a very small radius around the origin, and their mean is close to zero in both image axes. These reprojection errors measure how consistent the solution is with the measurements it was fitted to.

\subsubsection{Multimodal Map}
We present a qualitative analysis of the final multimodal map, focusing on detailed views of specific regions to highlight the improvements in alignment post-optimization. To clearly visualize the corrections, overlays of corresponding sonar and optical data are shown for several key areas.

\begin{figure*}
\centering

\subfloat[Before optimization]{%
    \includegraphics[width=0.48\linewidth]{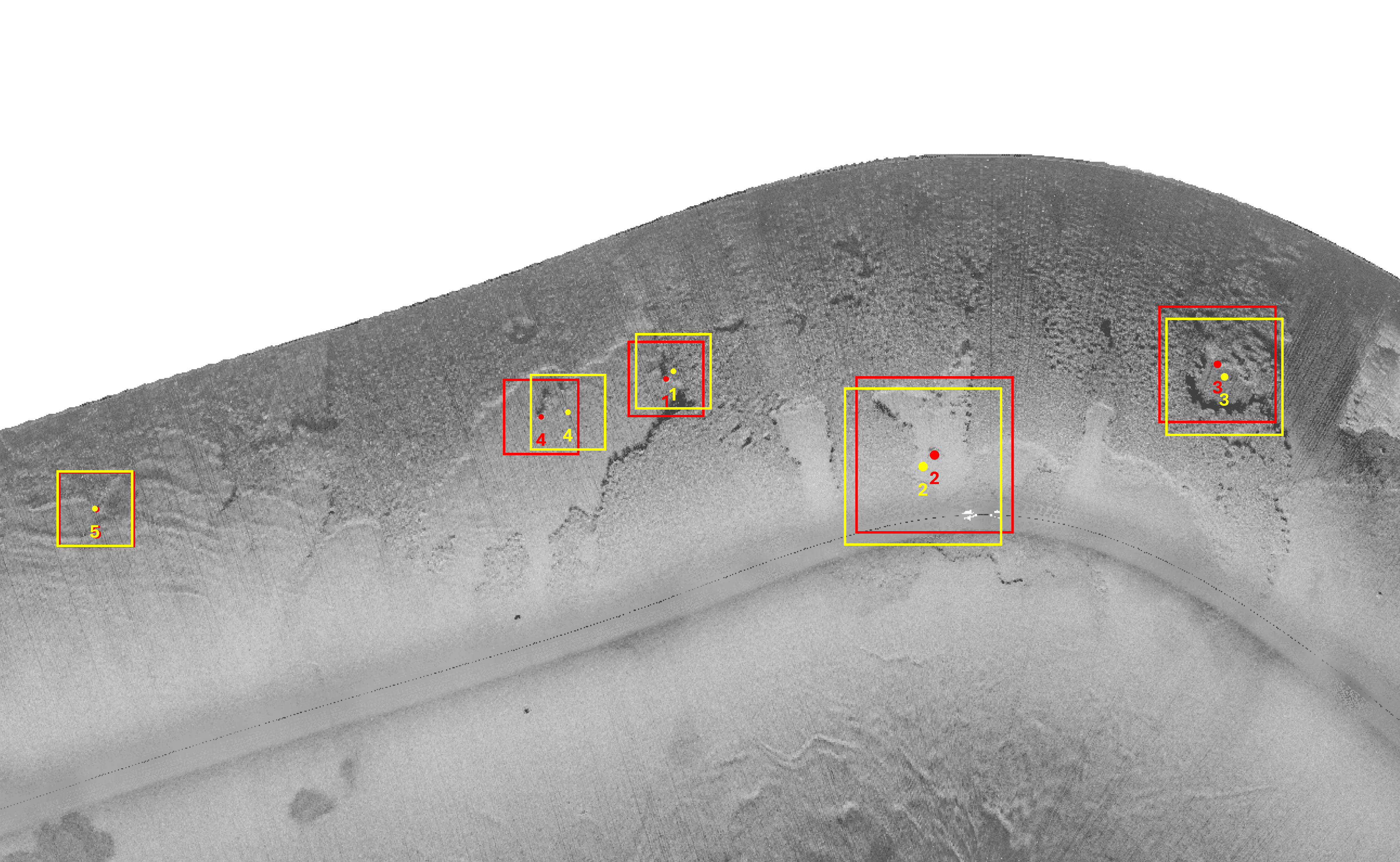}
}
\hfill
\subfloat[After optimization]{%
    \includegraphics[width=0.48\linewidth]{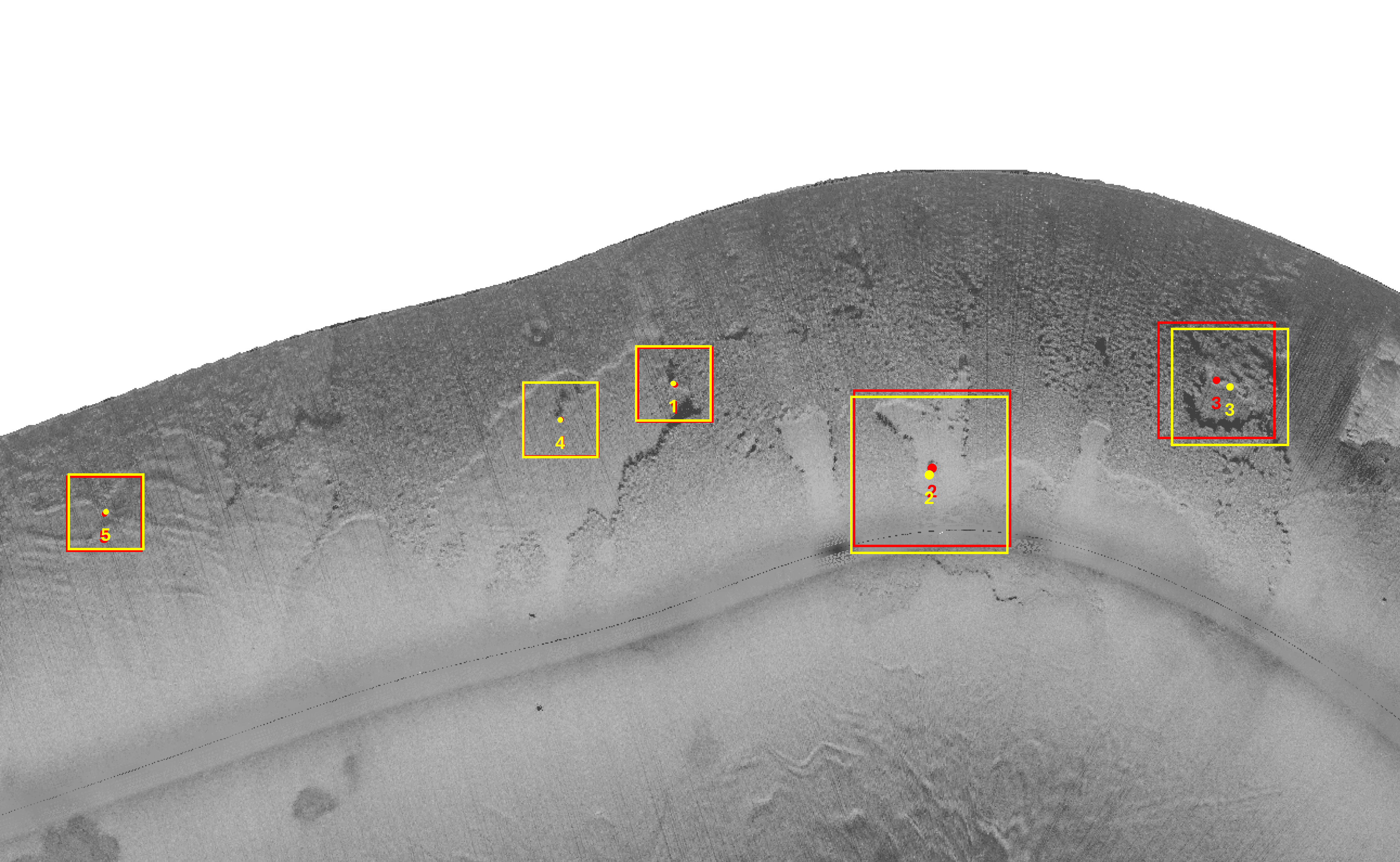}
}

\vspace{-0.3em}

\subfloat[Before optimization (\textit{SON2} overlaid)]{%
    \includegraphics[width=0.48\linewidth]{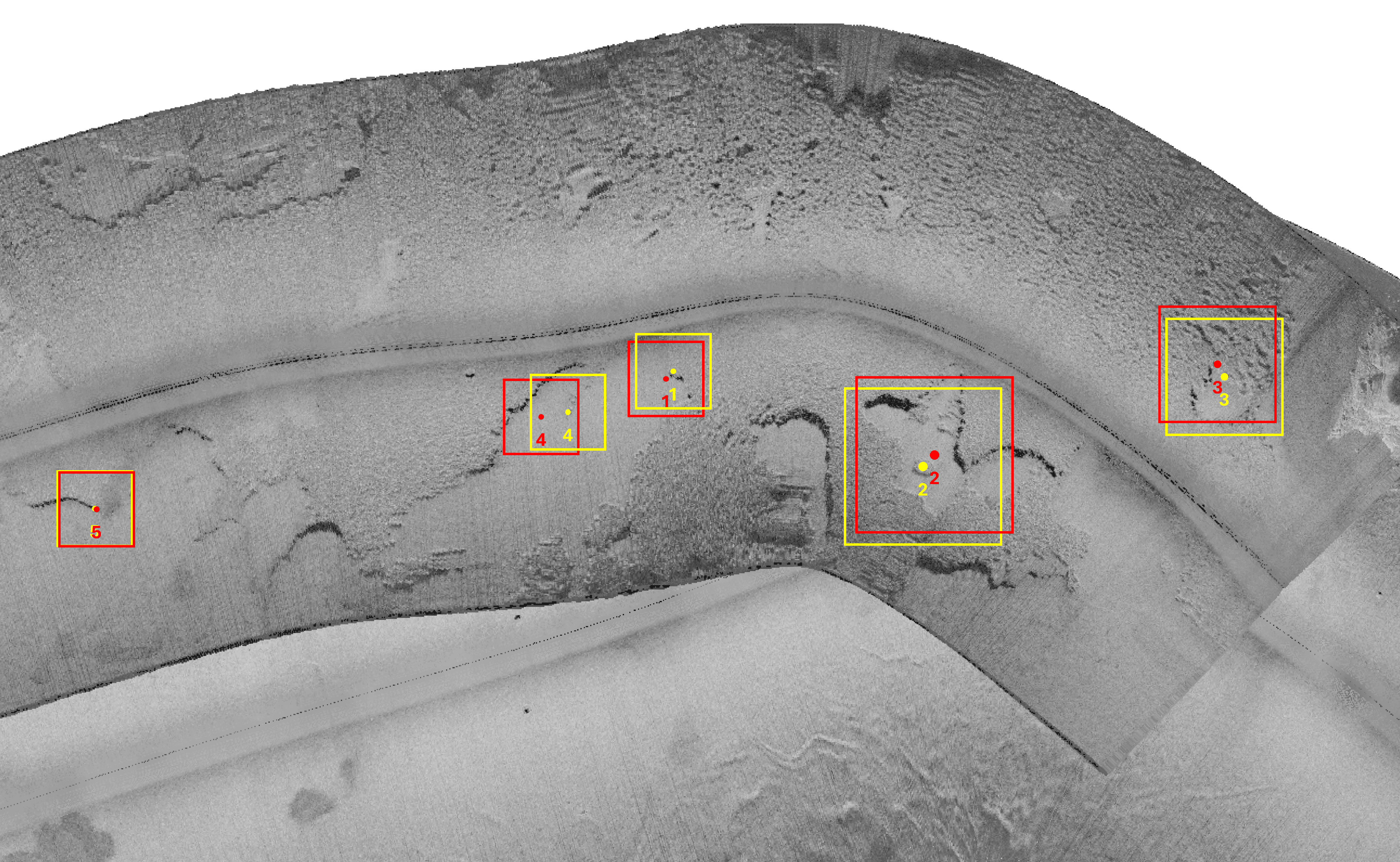}
}
\hfill
\subfloat[After optimization (\textit{SON2} overlaid)]{%
    \includegraphics[width=0.48\linewidth]{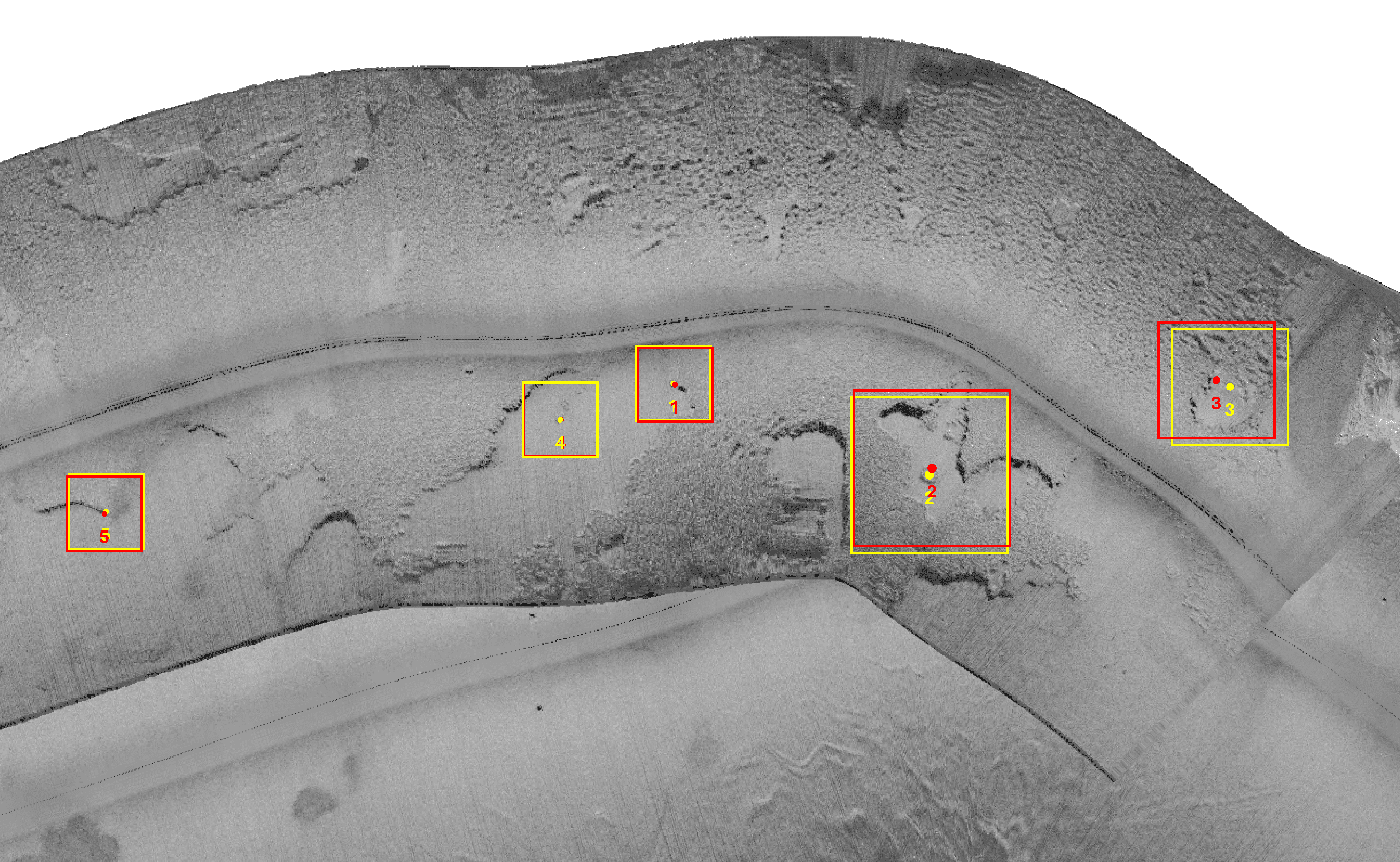}
}

\caption{Mosaics of \textit{SON1} and \textit{SON2} before (a,c) and after (b,d) optimization. Red points and boxes indicate keypoint features and regions of interest in \textit{SON1}; yellow points and boxes indicate the corresponding features in \textit{SON2}. In (c,d), \textit{SON2} is overlaid on \textit{SON1}.}
\label{fig:Qualson23}

\end{figure*}

\begin{figure*}
\centering

\subfloat[Before optimization]{%
    \includegraphics[width=0.48\linewidth]{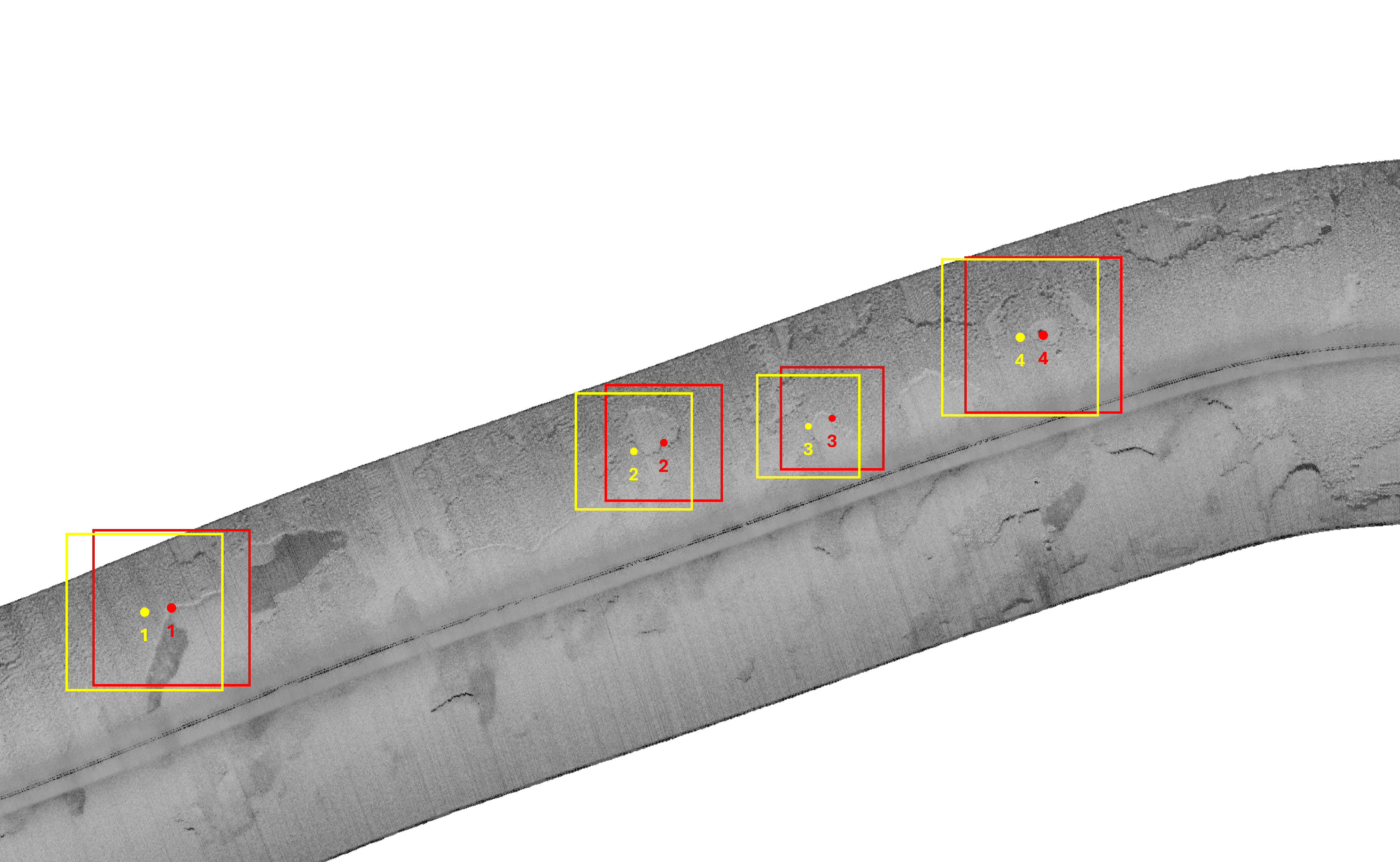}
}
\hfill
\subfloat[After optimization]{%
    \includegraphics[width=0.48\linewidth]{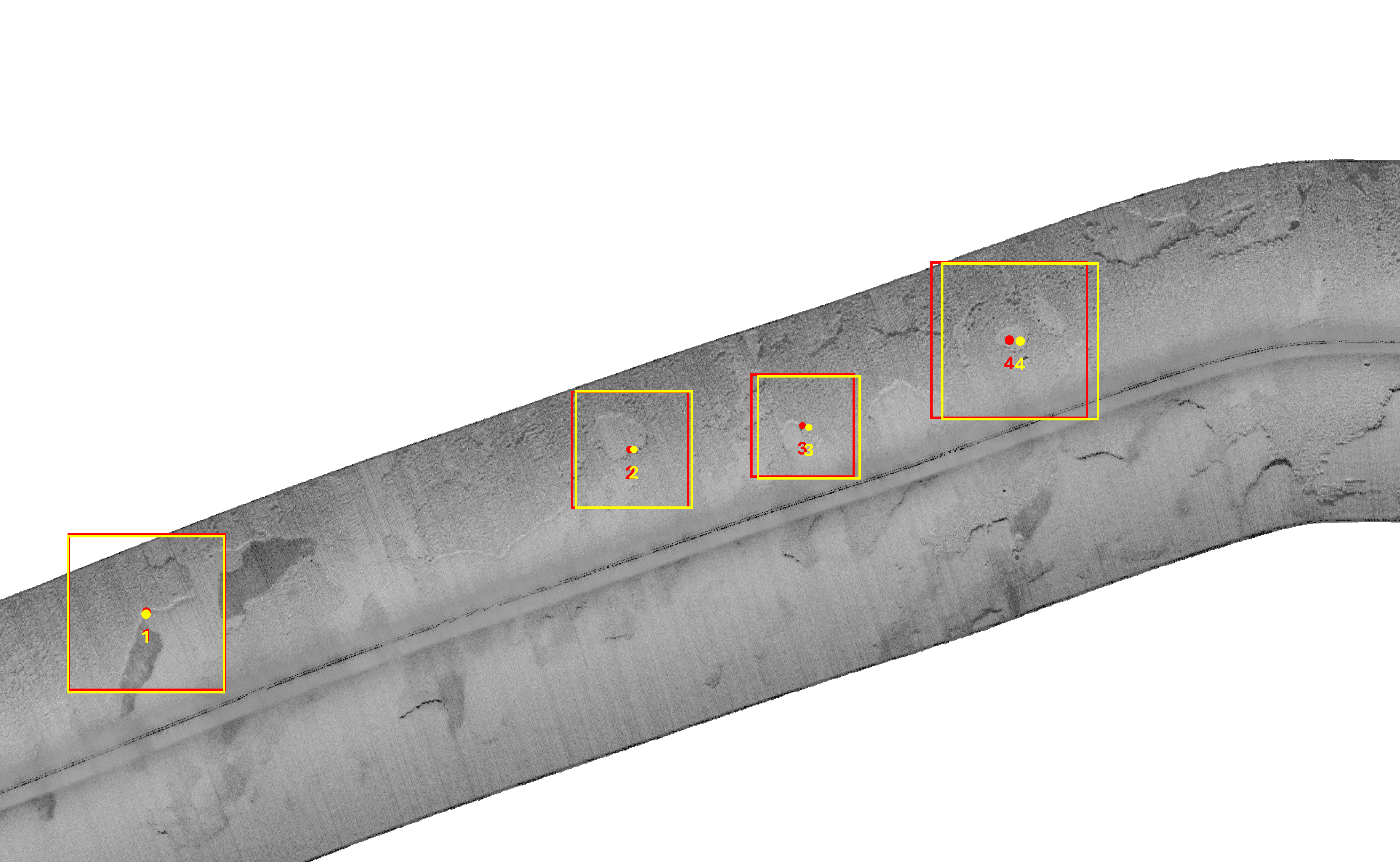}
}

\vspace{-0.3em}

\subfloat[Before optimization (\textit{SON3} overlaid)]{%
    \includegraphics[width=0.48\linewidth]{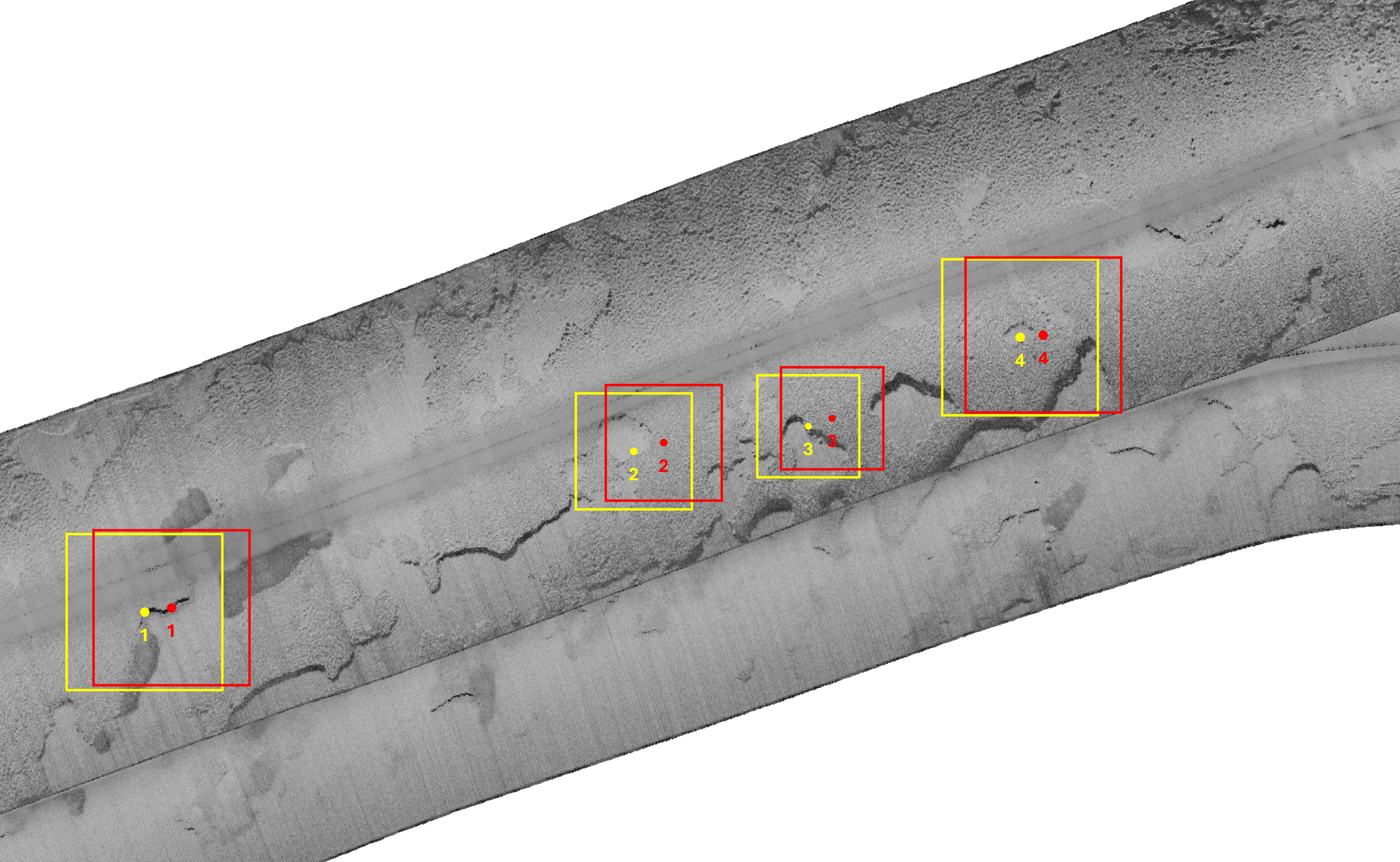}
}
\hfill
\subfloat[After optimization (\textit{SON3} overlaid)]{%
    \includegraphics[width=0.48\linewidth]{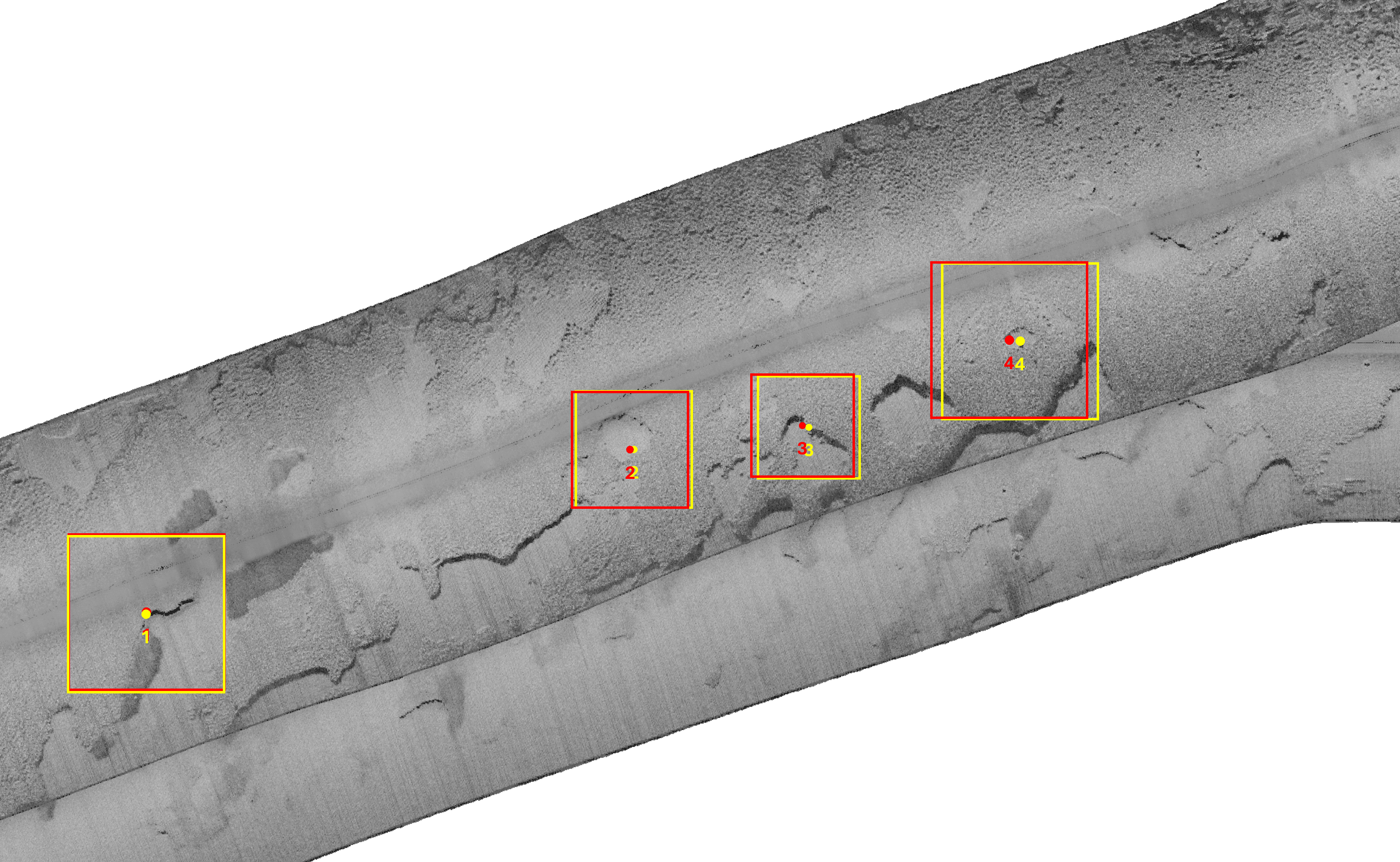}
}

\caption{Mosaics of \textit{SON2} and \textit{SON3} before (a,c) and after (b,d) optimization. Red points and boxes indicate keypoint features and regions of interest in \textit{SON2}; yellow points and boxes indicate the corresponding features in \textit{SON3}. In (c,d), \textit{SON3} is overlaid on \textit{SON2}.}
\label{fig:Qualson34}

\end{figure*}

\begin{figure*}
\centering

\subfloat[Before optimization]{%
    \includegraphics[width=0.48\linewidth]{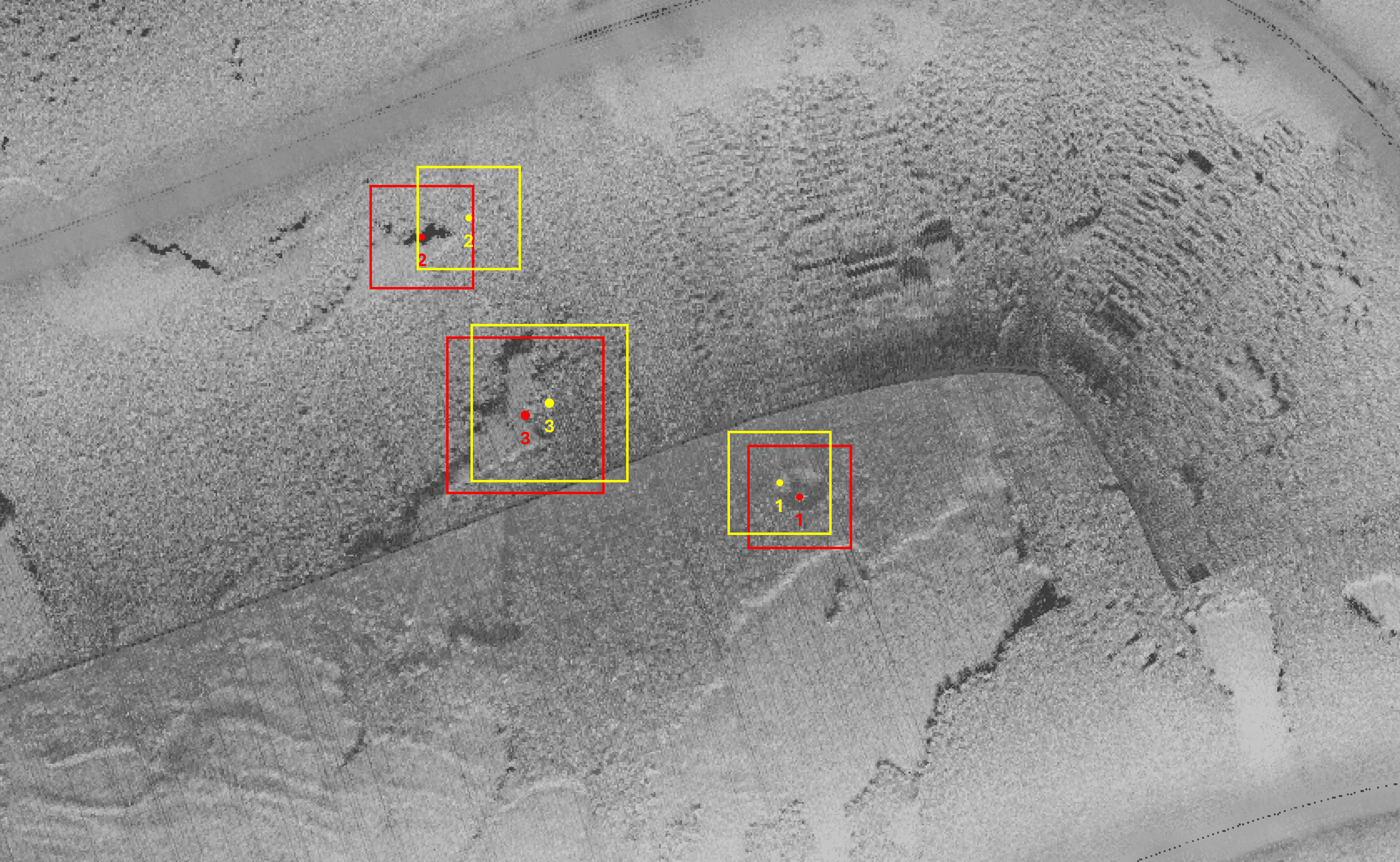}
}
\hfill
\subfloat[After optimization]{%
    \includegraphics[width=0.48\linewidth]{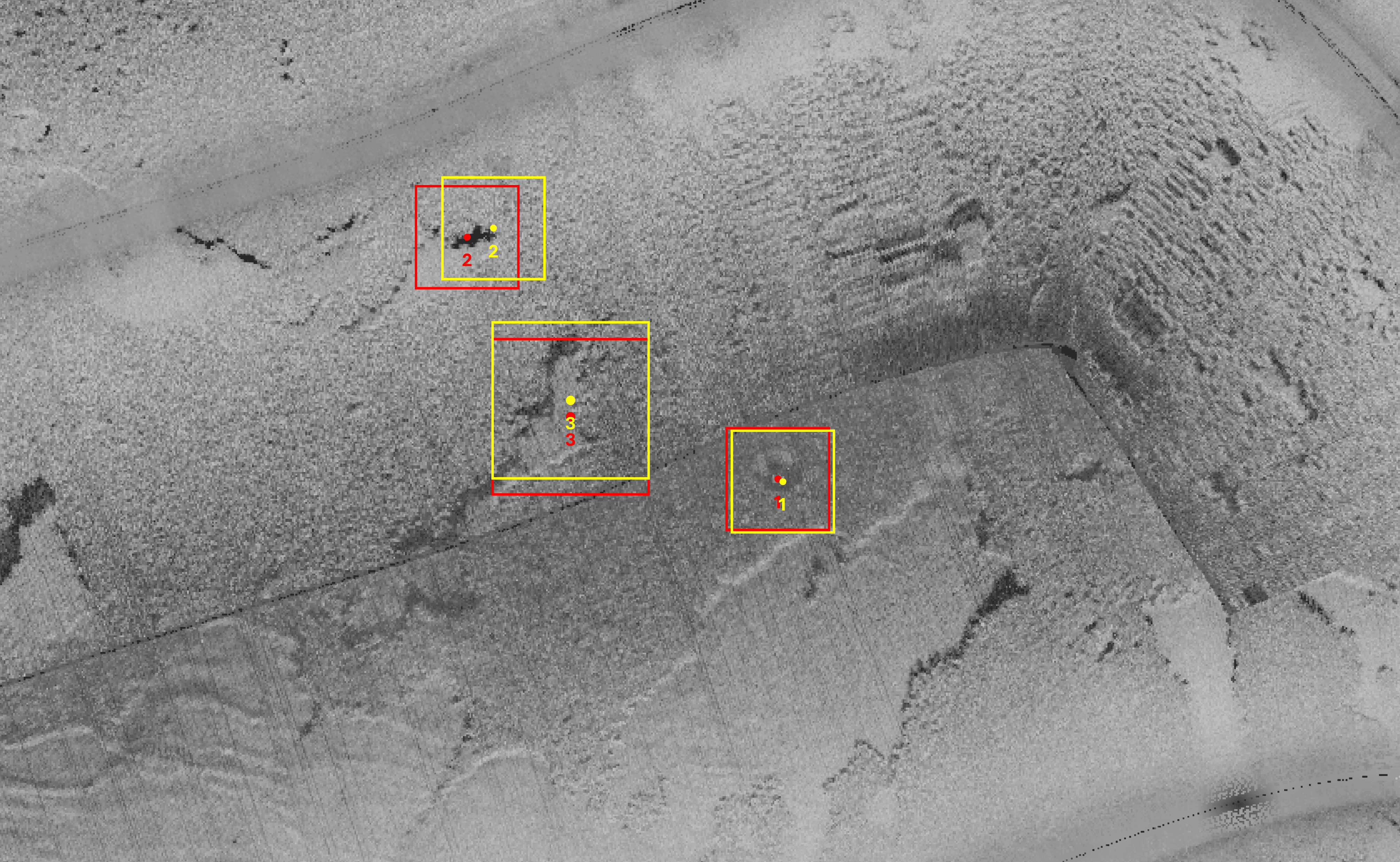}
}

\vspace{-0.3em}

\subfloat[Before optimization (\textit{CAM1} overlaid)]{%
    \includegraphics[width=0.48\linewidth]{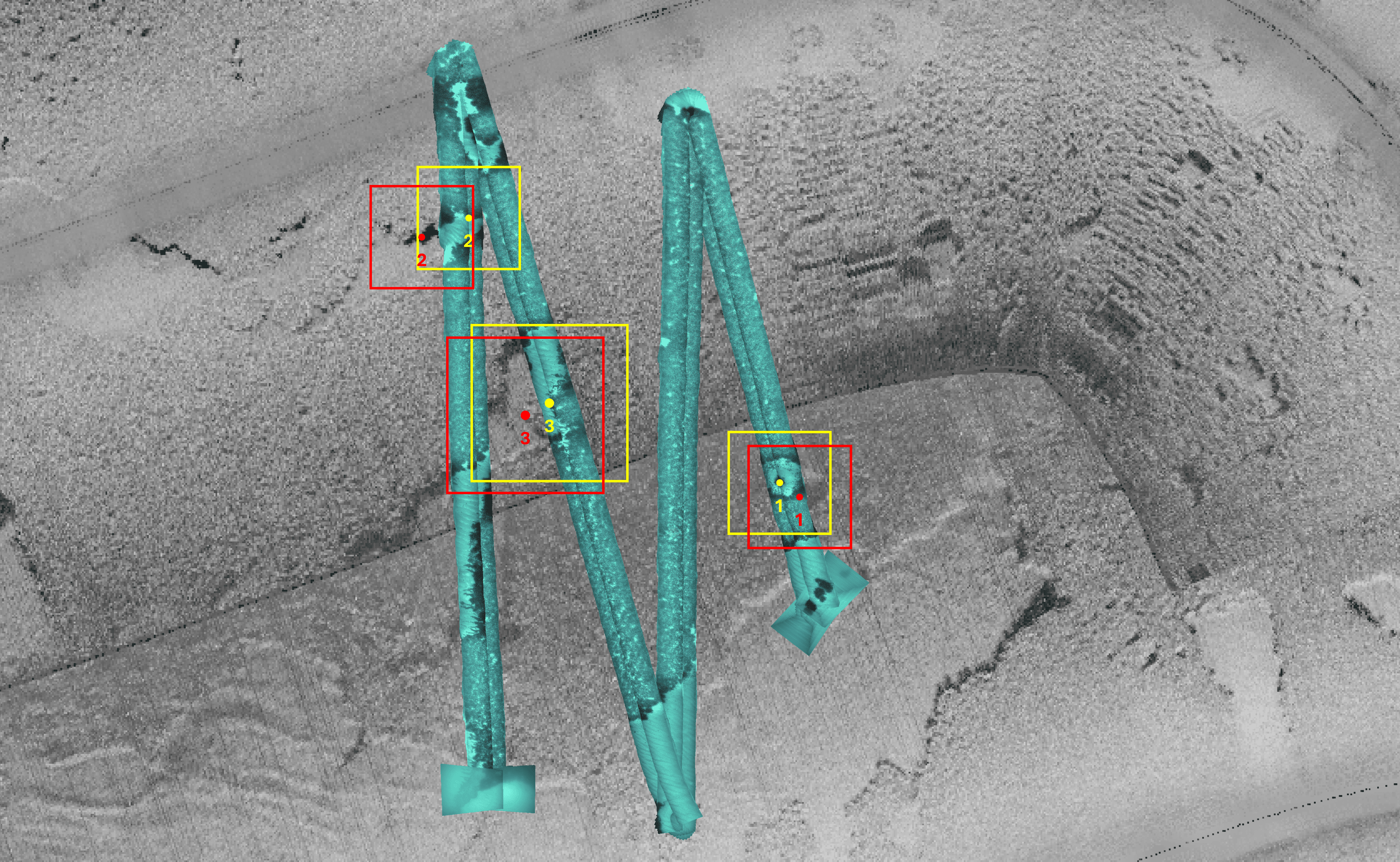}
}
\hfill
\subfloat[After optimization (\textit{CAM1} overlaid)]{%
    \includegraphics[width=0.48\linewidth]{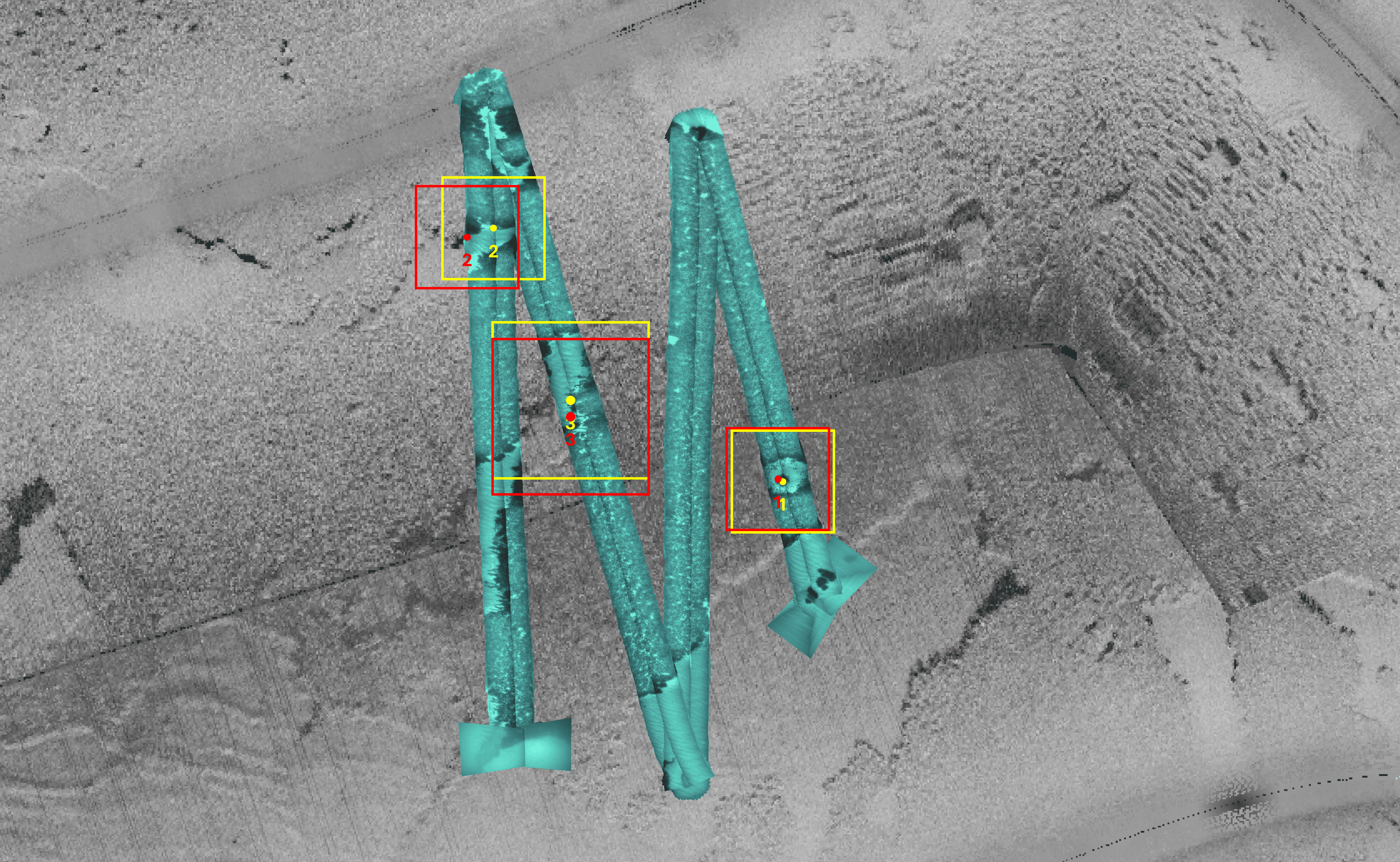}
}

\caption{Mosaics of \textit{SON1}, \textit{SON3}, and \textit{CAM1} before (a,c) and after (b,d) optimization. Red points and boxes indicate keypoint features and regions of interest in \textit{SON1} \& \textit{SON3}; yellow points and boxes indicate the corresponding features in \textit{CAM1}. In (c,d), \textit{CAM1} is overlaid on \textit{SON1} \& \textit{SON3}.}
\label{fig:Qualson23cam1}

\end{figure*}

\begin{figure*}
\centering

\subfloat[Before optimization]{%
    \includegraphics[width=0.48\linewidth]{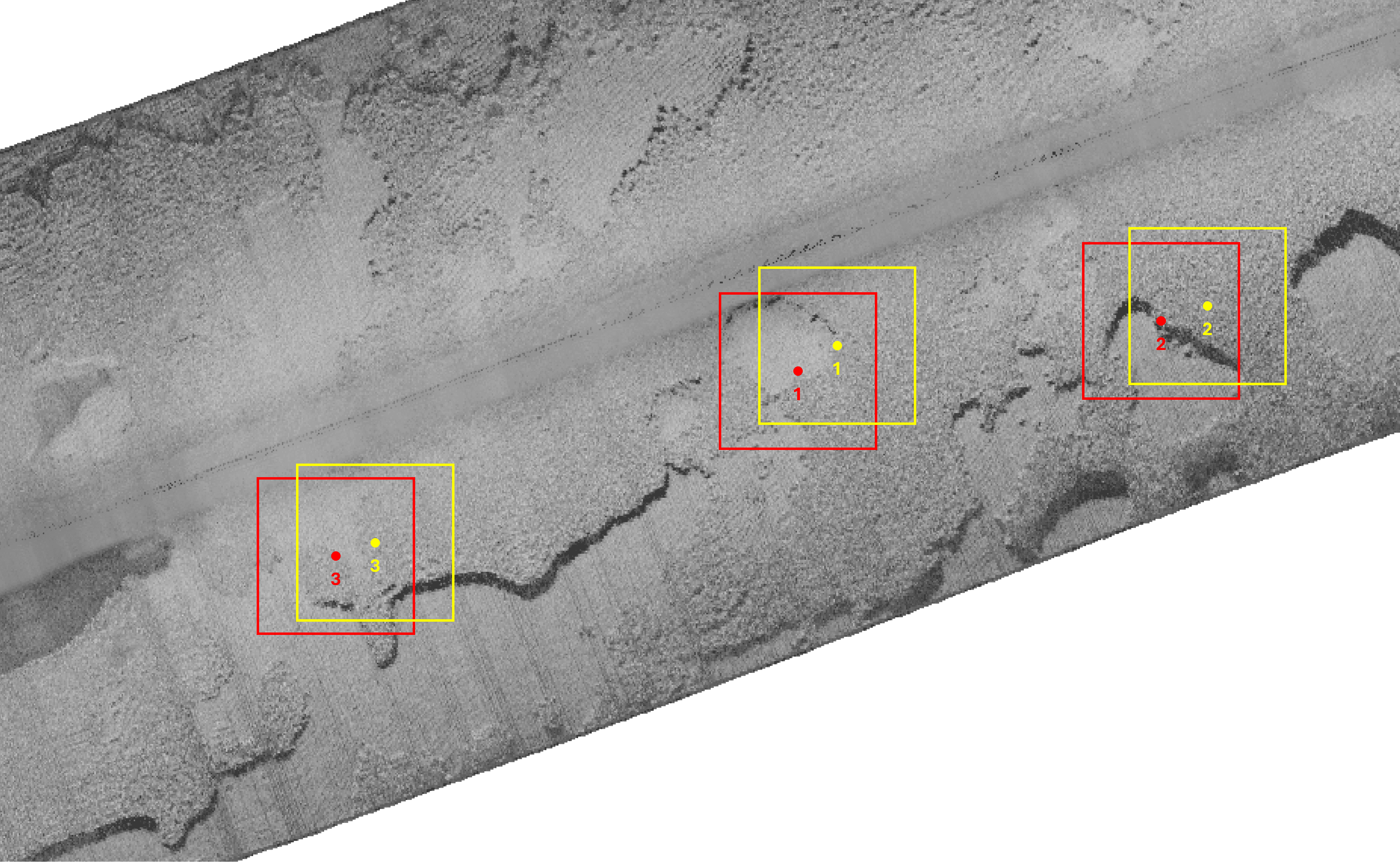}
}
\hfill
\subfloat[After optimization]{%
    \includegraphics[width=0.48\linewidth]{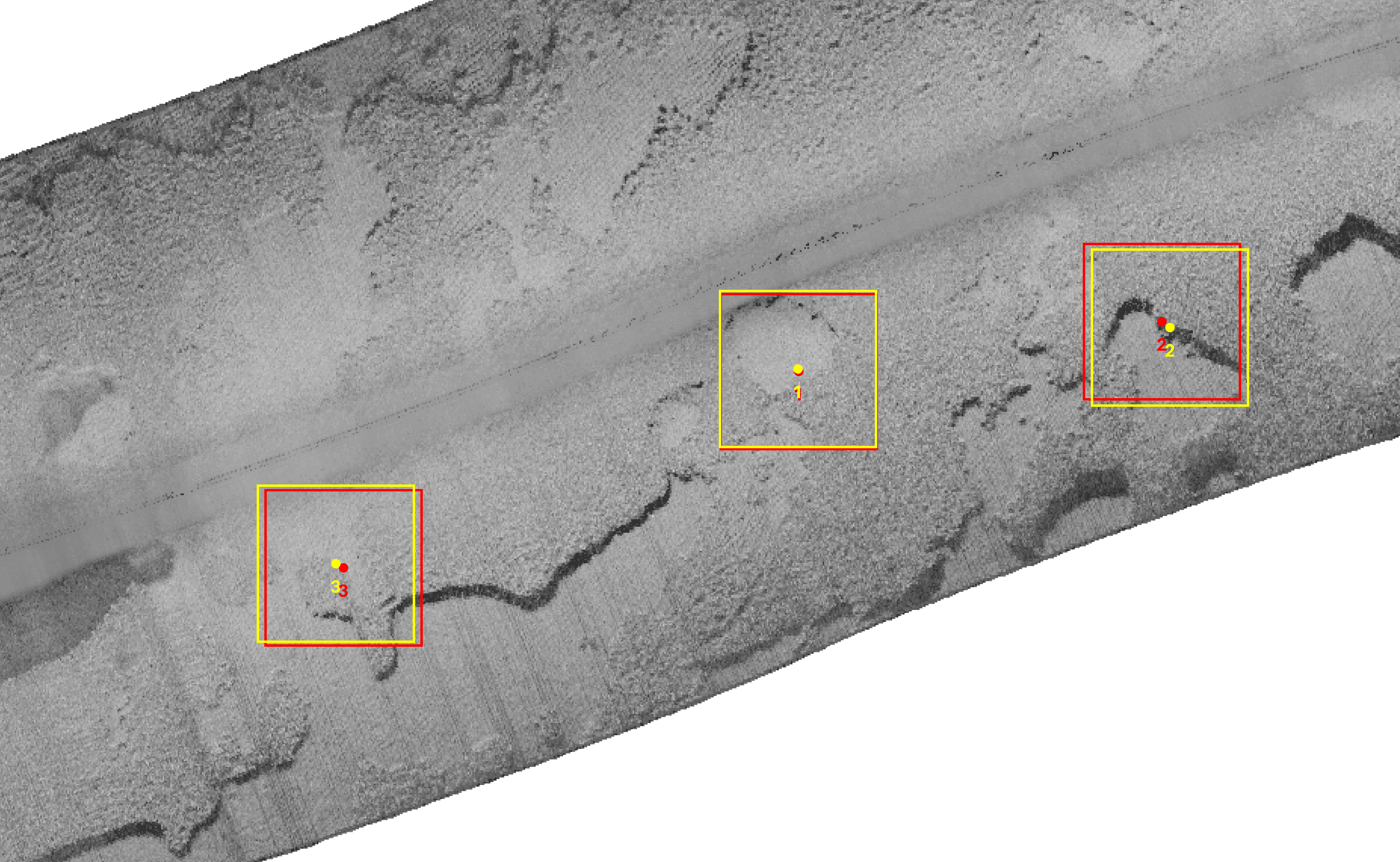}
}

\vspace{1em}

\subfloat[Before optimization (\textit{CAM4} overlaid)]{%
    \includegraphics[width=0.48\linewidth]{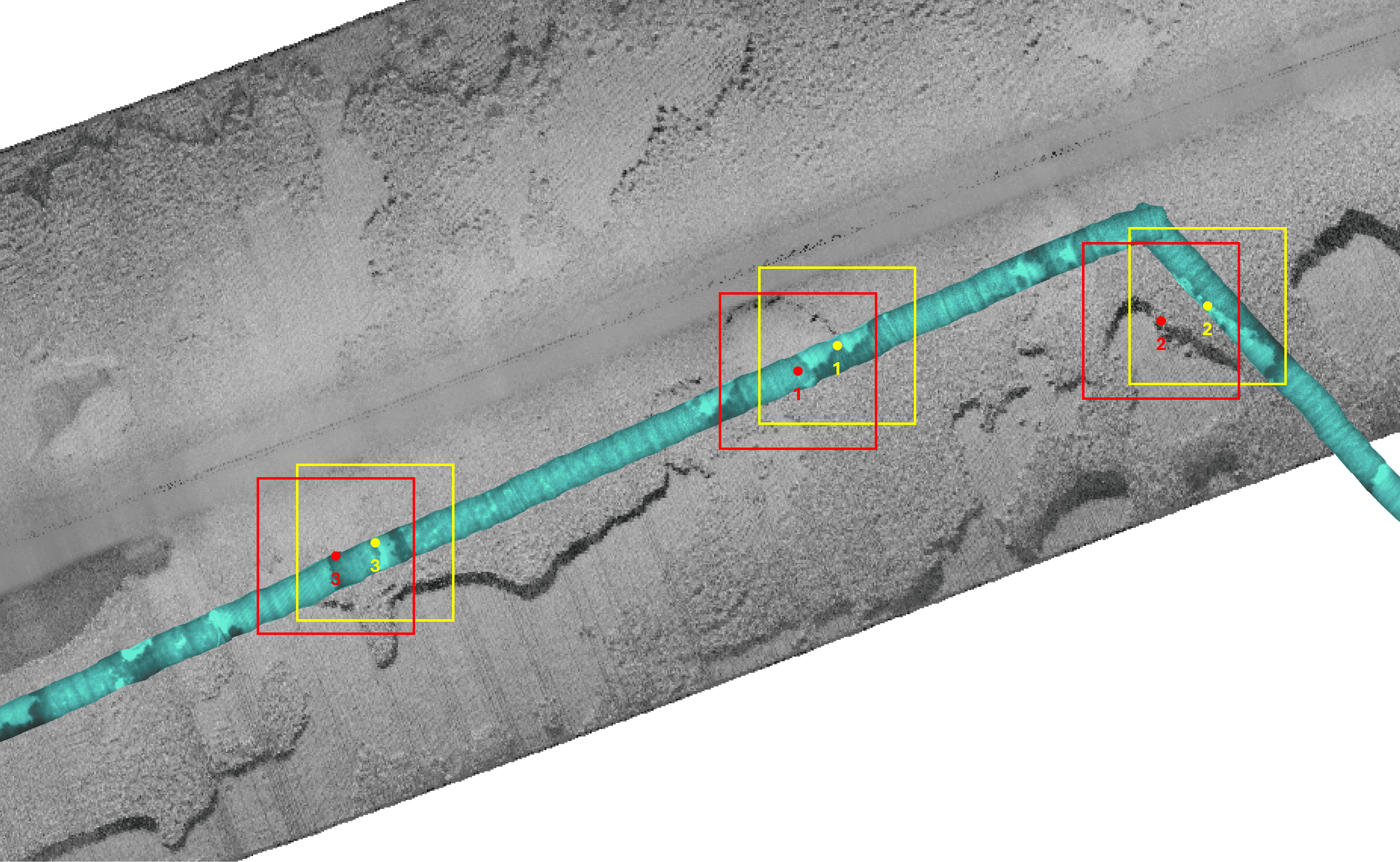}
}
\hfill
\subfloat[After optimization (\textit{CAM4} overlaid)]{%
    \includegraphics[width=0.48\linewidth]{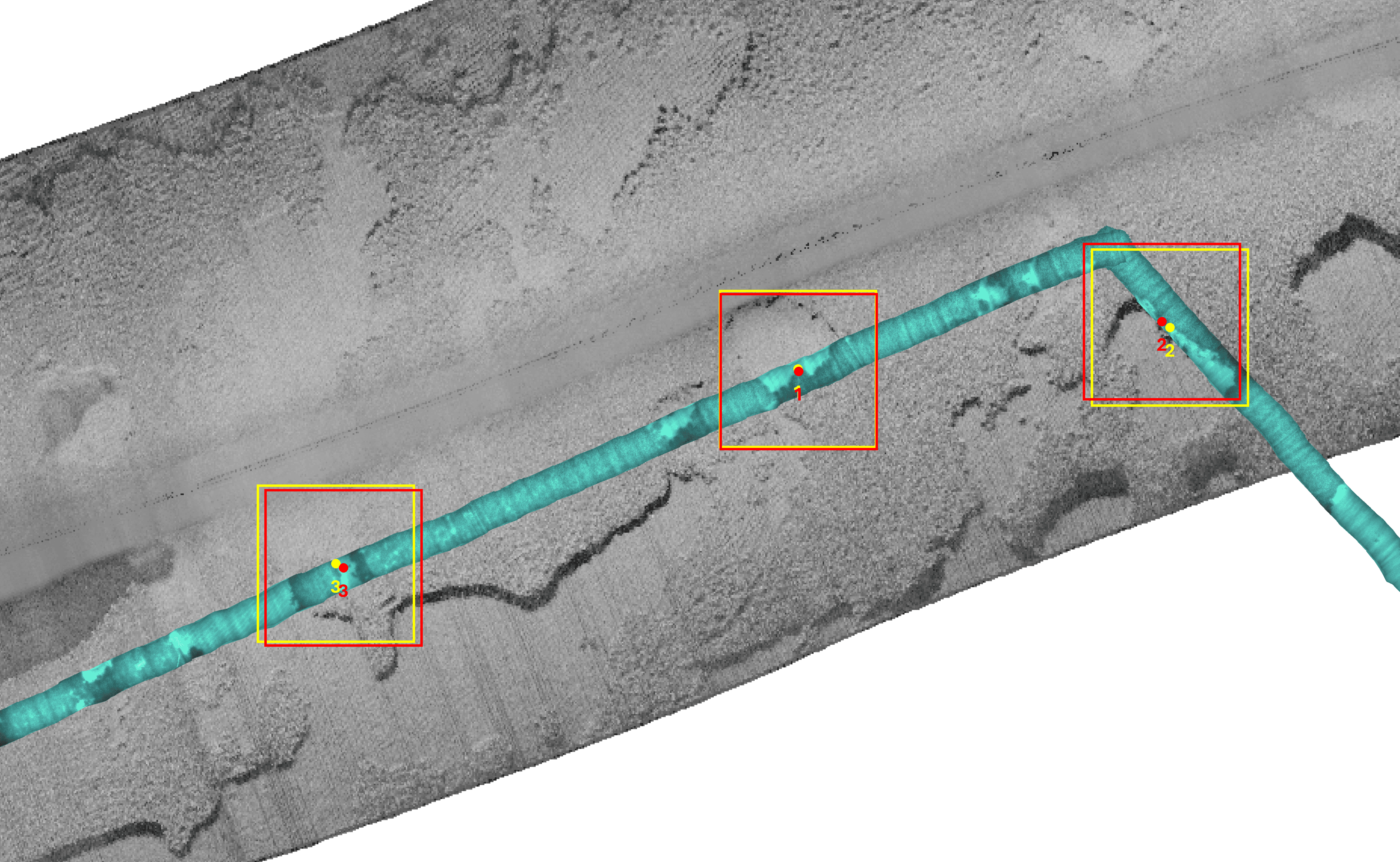}
}

\caption{Mosaics of \textit{SON3} and \textit{CAM4} before (a,c) and after (b,d) optimization. Red points and boxes indicate keypoint features and regions of interest in \textit{SON3}; yellow points and boxes indicate the corresponding features in \textit{CAM4}. In (c,d), \textit{CAM4} is overlaid on \textit{SON3}.}
\label{fig:Qualson4cam4}

\end{figure*}

Figures \ref{fig:Qualson23} through \ref{fig:Qualson4cam4} present a comparison of these sonar and optical mosaic patches before and after the optimization. The initial, unoptimized overlays show significant misalignments between features across different sessions and modalities. After optimization, the overlays (highlighted with colored outlines) demonstrate a substantial improvement in the co-registration of corresponding features. This visual evidence confirms the framework's effectiveness at correcting the large initial offsets and achieving a more coherent, multimodal map.

It is important to analyze the nature of the applied corrections, which are inherently non-rigid. The framework performs a two-level alignment:

\begin{itemize}
    \item A global rigid transformation via the session anchor, $\mathbf{\Delta}^j$, which corrects the overall offset and orientation of each session relative to the world frame.
    \item A local non-rigid deformation via the optimization of individual state nodes, \(\mathbf{X}_k^j\), which allows the trajectory within each session to be warped, stretched, and bent.
\end{itemize}

This non-rigid capability is designed to address complex, time-varying errors that stem from sources such as accumulated navigation drift, intermittent localization inaccuracies, or imperfect time synchronization of the various sensors.

For instance, Figures \ref{fig:Qualson23} and \ref{fig:Qualson23cam1} show a sonar survey with significant trajectory distortions where a purely rigid transformation would not be sufficient. The proposed framework, by allowing local pose adjustments, substantially reduces these nonlinear misalignments. While the optimization does not eliminate all non-rigid effects, the results demonstrate the promise of this approach.

\subsubsection{Comparison with Rigid Baseline}
\label{sec:rigid_baseline}
We compare our proposed approach with a baseline that performs a rigid $\mathrm{SE}(3)$ multi-session optimization. Rather than jointly optimizing all variables simultaneously, this baseline decouples the problem into two sequential stages. 
In the first stage, each optical session is optimized independently as a subgraph of the full factor graph, using only intra-session measurements — camera observations of landmarks seen only in that session, depth, attitude, and relative heading factors, constant velocity motion model, and, if available, DVL measurements. The objective for session $j$ is:

{\small
\begin{align}
  &\boldsymbol{\Theta}^{j*} = \underset{\boldsymbol{\Theta}^{j}}{\operatorname{argmin}} \Bigg(
    \left\|\mathbf{r}^j_{\text{X}_0}\right\|^2_{\boldsymbol{\Sigma}_{\text{X}_0}}
  + \sum_m \left\|\mathbf{r}^j_{\text{C},m}\right\|^2_{\boldsymbol{\Sigma}_\text{C}}
  \nonumber \\
  &+ \sum_k \left( \left\|r^j_{\text{d},k}\right\|^2_{\Sigma_\text{d}}
  + \left\|\mathbf{r}^j_{\text{a},k}\right\|^2_{\boldsymbol{\Sigma}_\text{a}} \right)
  + \sum_{k \geq 1} \left\|r^j_{\psi,k}\right\|^2_{\Sigma_\psi}
  \nonumber \\
  &+ \sum_k \left( \left\|\mathbf{r}^j_{\text{e},k}\right\|^2_{\boldsymbol{\Sigma}_\text{e}}
  + \left\|\mathbf{r}^j_{\text{u},k}\right\|^2_{\boldsymbol{\Sigma}_\text{u}} \right)
  \nonumber \\
  &+ \sum_{(j,\,m,\,k,\,i)\,\in\,\mathcal{O}^j_{\text{CAM}}}
    \left\|\mathbf{r}^j_{\text{CAM},k,i,m}\right\|^2_{\boldsymbol{\Sigma}_{\text{CAM}}} \Bigg)
\label{eq:least_squares_rigid_first_stage}
\end{align}
}

\noindent where the residuals are those of Eq.~\eqref{eq:least_squares_final} evaluated with $\boldsymbol{\Delta}^j = \bar{\boldsymbol{\Delta}}^j$, $\mathcal{O}^j_{\text{CAM}} \subset \mathcal{O}_{\text{CAM}}$ denotes the camera observations of landmarks observed only by session $j$, $\mathcal{P}^j$ is the set of these landmarks, and $\boldsymbol{\Theta}^j = \{\{\mathbf{X}_k^j\}, \{\mathbf{C}_m^j\}, \{\mathbf{q}_i\}_{i \in \mathcal{P}^j}\}$. The anchor $\boldsymbol{\Delta}^j$ is held fixed at its initial estimate during this stage and is not a variable of this optimization. Since sonar sessions contain only absolute pose measurements and no intra-session landmark observations, their subgraphs provide no additional constraints beyond the GNSS/USBL priors already incorporated and are therefore not optimized at this stage.

In the second stage, we fix all trajectory and extrinsic variables at their first-stage estimates, denoted $\hat{\mathbf{X}}_k^j$ and $\hat{\mathbf{C}}_m^j$, and jointly optimize the anchors and landmark positions using the full set of unimodal sonar and multimodal landmark measurements, and the navigation factors, with which fixed poses constrain only the anchors:

{\small
\begin{align}
  &\boldsymbol{\beta}^* = \underset{\boldsymbol{\beta}}{\operatorname{argmin}} \Bigg(
    \sum_{j \in \mathcal{J}} \left\|\mathbf{r}^j_{\Delta}\right\|^2_{\boldsymbol{\Sigma}_{\Delta}} + \sum_{j \in \mathcal{J}} \sum_k \left( \left\|r^j_{\text{d},k}\right\|^2_{\Sigma_\text{d}}
  + \left\|\mathbf{r}^j_{\text{a},k}\right\|^2_{\boldsymbol{\Sigma}_\text{a}} \right)
  \nonumber \\
  &+ \sum_{j \in \mathcal{J}} \sum_{k \geq 1} \left\|r^j_{\psi,k}\right\|^2_{\Sigma_\psi}
  + \sum_{j \in \mathcal{J}_T} \sum_{k \geq 1} \left\|\mathbf{r}^j_{xy,k}\right\|^2_{\boldsymbol{\Sigma}_{xy}}
  \nonumber \\
  &+ \sum_{(j,\,k,\,i)\,\in\,\mathcal{O}_{\text{SSS}}}
    \left\|\mathbf{r}^j_{\text{SSS},k,i}\right\|^2_{\boldsymbol{\Sigma}_{\text{SSS}}}
  \nonumber \\
  &+ \sum_{(j,\,m,\,k,\,i)\,\in\,\mathcal{O}_{\text{CAM}} \setminus \bigcup_{l} \mathcal{O}^l_{\text{CAM}}}
    \left\|\mathbf{r}^j_{\text{CAM},k,i,m}\right\|^2_{\boldsymbol{\Sigma}_{\text{CAM}}} \Bigg)
\label{eq:least_squares_rigid_second_stage}
\end{align}
}

\noindent where $\boldsymbol{\beta} = \{\{\boldsymbol{\Delta}^j\}_{j \in \mathcal{J}}, \{\mathbf{q}_i\}_{i \notin \bigcup_{l} \mathcal{P}^l}\}$, and every residual is evaluated with $\mathbf{X}_k^j = \hat{\mathbf{X}}_k^j$ and $\mathbf{C}_m^j = \hat{\mathbf{C}}_m^j$, and $\mathbf{p}_i = \pi_i(\mathbf{H}_i, \mathbf{q}_i)$ (Eq.~\eqref{eq:hosted_world_point}), as in the proposed method. The landmarks observed by a single optical session are not re-optimized; they move rigidly with their session anchor, $\mathbf{p}_i^* = \boldsymbol{\Delta}^{j*} (\bar{\boldsymbol{\Delta}}^j)^{-1} \hat{\mathbf{p}}_i$ for $i \in \mathcal{P}^j$.

Both methods use identical noise models to ensure a fair comparison. Table \ref{tab:rigid_non_rigid_reproj_error} reports the mean and standard deviation of the camera and sonar reprojection errors after optimization. For the sonar observations, the proposed method reduces both the mean error and the standard deviation by a factor of about four relative to the rigid baseline. For the camera observations, the mean errors are nearly identical, but the proposed method has a lower standard deviation. This result is expected, since the rigid baseline is constrained to a single global $\mathrm{SE}(3)$ correction per session and cannot resolve the intra-session distortions that the proposed method addresses through local trajectory deformations.

\begin{table}[t]
\centering
\caption{Average reprojection error for our proposed and rigid optimization approaches}
\label{tab:rigid_non_rigid_reproj_error}
\resizebox{\columnwidth}{!}{%
\begin{tabular}{lcc}
\toprule
Reprojection error type & Proposed Approach & Rigid Optimization \\
\midrule
Camera (px) & $1.08 \pm 1.72$  & $1.09 \pm 2.19$ \\
Sonar (m) & $0.57 \pm 0.75$ & $2.10 \pm 3.02$ \\
\bottomrule
\end{tabular}
}
\end{table}

\begin{figure*}
    \centering

    \subfloat[Sonar patch]{%
        \includegraphics[width=0.45\linewidth]{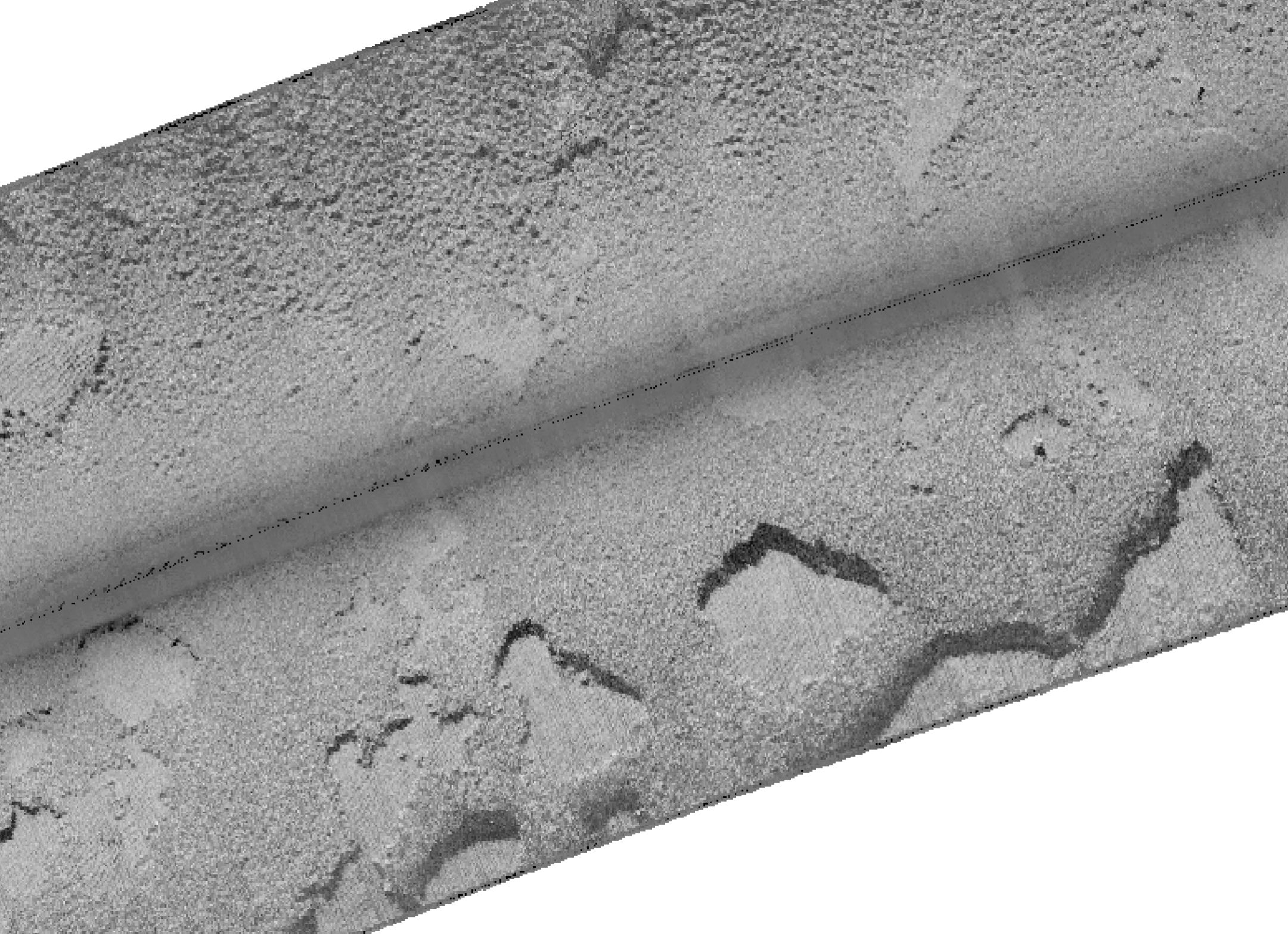}
    }
    \hfill
    \subfloat[Sonar annotation mask]{%
        \includegraphics[width=0.45\linewidth]{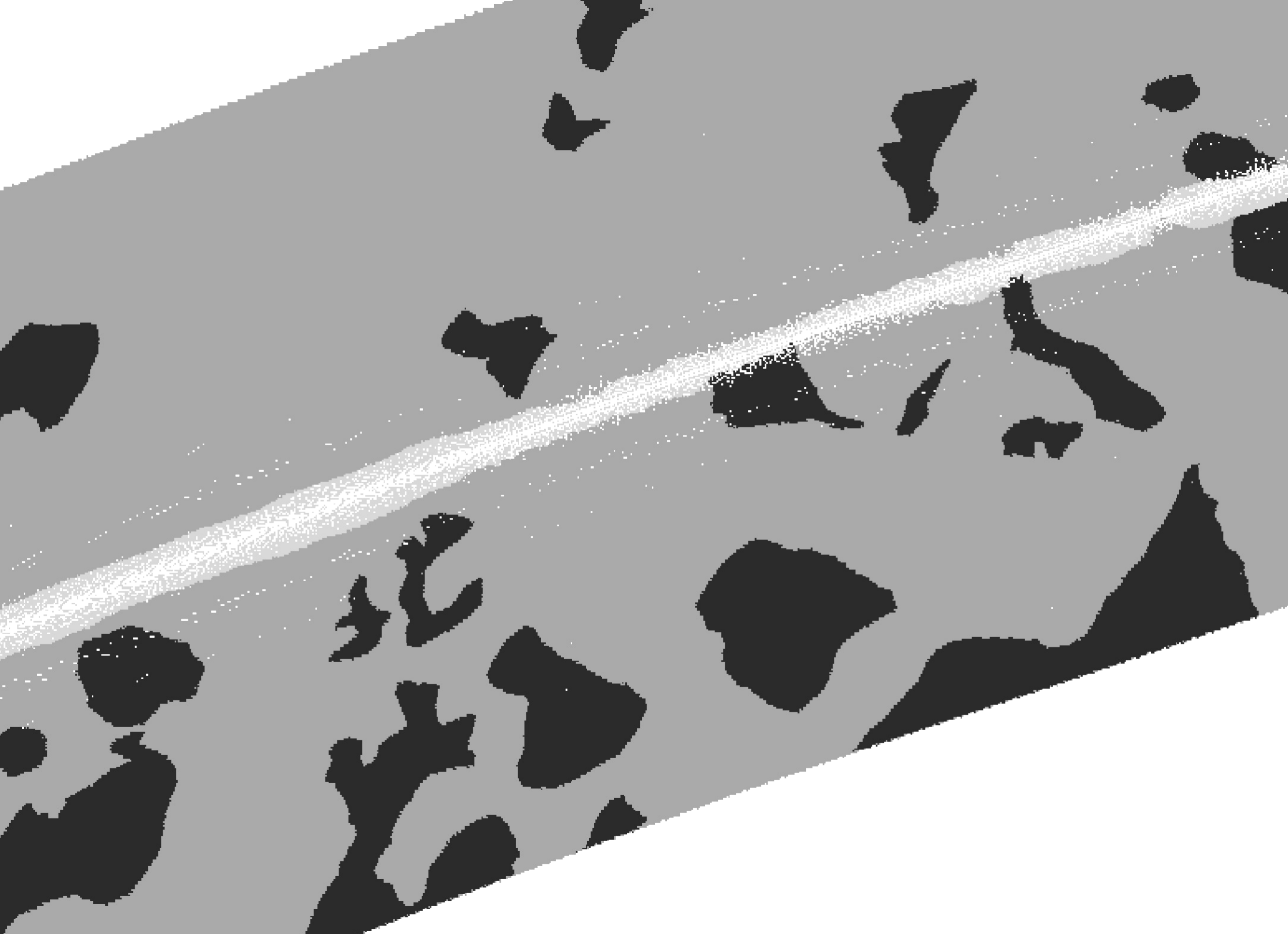}
    }
    \caption{Patch from a sonar mosaic and its corresponding segmentation mask}
    \label{fig:sonar_w_masks}
\end{figure*}

\begin{figure*}
    \centering

    \subfloat[Optical mosaic patch]{%
        \includegraphics[width=0.45\linewidth]{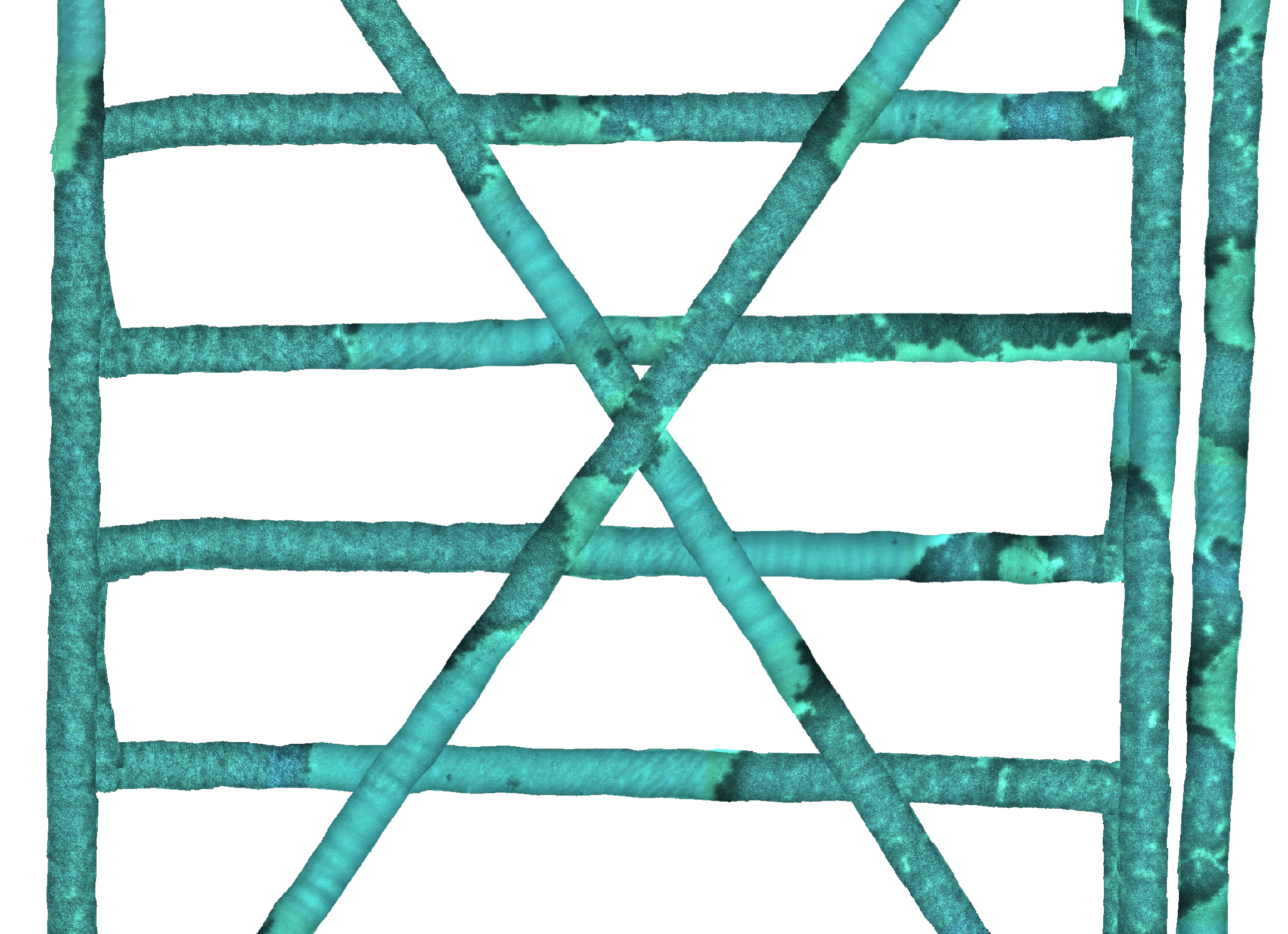}
    }
    \hfill
    \subfloat[Optical segmented mask]{%
        \includegraphics[width=0.45\linewidth]{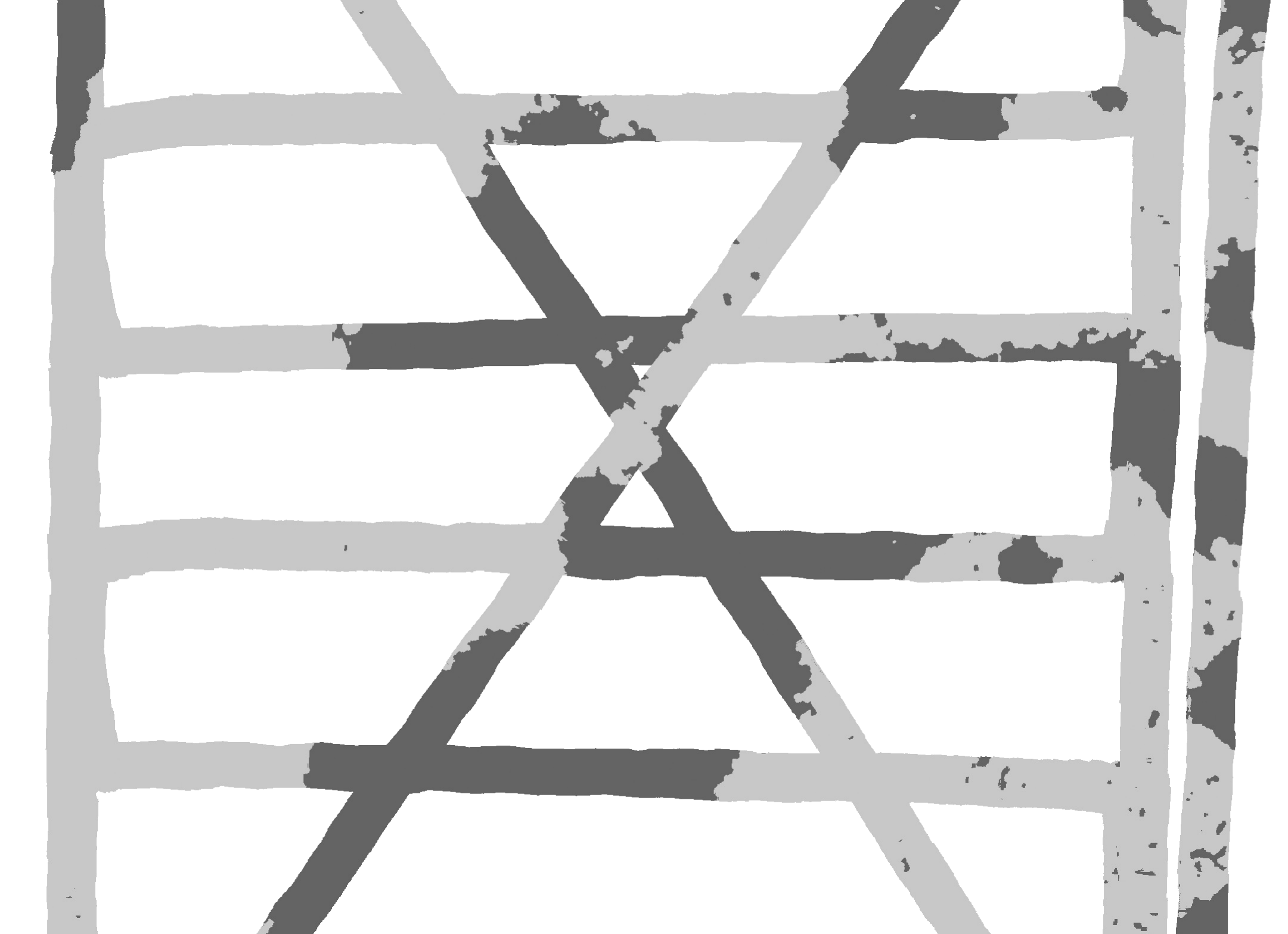}
    }
    \caption{Patch from an optical mosaic and its corresponding segmentation mask}
    \label{fig:optical_w_masks}
\end{figure*}

To evaluate the alignment accuracy of the multimodal maps generated by the two methods, annotated sonar and optical mosaics are generated using the trajectories estimated by each method. For sonar sessions, class-level annotations were done manually by an expert annotator in the waterfall image space. The waterfall images were segmented into three classes: seagrass, sand, and rocks. All ping intensity values in the sonar data are then replaced with intensities that correspond to their respective classes.
For optical sessions, we perform a binary semantic segmentation on all the images in our dataset using a SegFormer-type model trained to classify seagrass \citep{ruscio2023autonomous, xie2021segformer}.
The modified sonar data and predicted segmentation masks for optical images are then used to generate georeferenced maps as explained in Section \ref{sec: map gen}. Figures \ref{fig:sonar_w_masks} and \ref{fig:optical_w_masks} depict patches from sonar and optical mosaics with their corresponding masks.

We use the Pixel Accuracy (PA) and mean Intersection over Union (mIoU) metrics to evaluate the spatial alignment of the multimodal multi-session segmented map. Pixel accuracy (PA) measures the fraction of pixels on which all sessions agree on the class label, out of all pixels observed by at least two sessions:

$$\text{PA} = \frac{\sum_{p \in \mathcal{S}} \mathbf{1}[\text{all images agree at } p]}{|\mathcal{S}|}$$

\noindent where $\mathcal{S}$ is the set of pixels observed by at least two sessions with valid class labels. To ensure that the methods differ only in their registration, $\mathcal{S}$ is the common set of pixels that are valid under every method.

For each class, the IoU is computed for every pair of overlapping session maps in which the class occurs and averaged over these pairs, and the mIoU is the mean of the class IoUs. Since the rock class is missing in optical maps, the rock IoU of every sonar-optical pair is therefore zero regardless of the alignment, which lowers the mIoU without reflecting registration quality. We therefore also report mIoU$_2$, the mean IoU over the two classes shared by both modalities (sand and seagrass).

To quantify the uncertainty of these metrics, we compute 95\% confidence intervals (CIs) with a spatial block bootstrap \citep{hall1985resampling, lahiri2013resampling}. The map is tiled into non-overlapping $10 \times 10$ m blocks (4975 blocks containing pixels of $\mathcal{S}$), and the blocks are resampled with replacement 2000 times. PA, mIoU and mIoU$_2$ are recomputed from the pixels of each resample, and the CIs are taken as the 2.5th and 97.5th percentiles of the resulting distributions (percentile method \citep{diciccio1996bootstrap}). The bootstrap is paired, i.e., the same block resample is applied to every method, so the CIs of the differences between methods account for the fact that all methods are evaluated on the same seabed. A difference is considered significant when its CI excludes zero.

Table \ref{tab:consistency_map_result} shows the alignment metrics and their CIs for maps obtained from the unoptimized trajectories (no optimization), the rigid optimization, and our proposed method, along with the paired differences between our approach and the other two methods. Both the rigid optimization and our proposed approach improved the pixel accuracy compared to the unoptimized map. Our proposed approach achieved the best score on all metrics. Although the CIs of the individual methods overlap, the paired differences show that our approach is better than the rigid baseline and the unoptimized maps.

To contextualize these results, the session maps are compared on a common grid with a resolution of 0.3 meters/pixel. The common overlap region ($\mathcal{S}$) contains approximately 4.3 million pixels. Consequently, the 2.55 percentage point improvement in pixel accuracy achieved by our proposed method over the unoptimized map corresponds to correctly aligning the semantic labels over approximately 10000 $\text{m}^2$ of mapped area.

\begin{table}[t]
\centering
\caption{Multimodal map consistency results. 95\% CIs in brackets, from a paired spatial block bootstrap (10 m blocks, 2000 resamples) on the common pixel set $\mathcal{S}$; differences of PA are in percentage points.}
\label{tab:consistency_map_result}

\small
\setlength{\tabcolsep}{3pt}
\begin{tabular}{lccc}
\toprule
Method & PA (\%) & mIoU & mIoU$_2$ \\
\midrule
No optimization & 89.81 & 0.509 & 0.661 \\
 & {\scriptsize [89.17, 90.41]} & {\scriptsize [0.489, 0.530]} & {\scriptsize [0.635, 0.686]} \\
Rigid optimization & 91.97 & 0.523 & 0.681 \\
 & {\scriptsize [91.46, 92.49]} & {\scriptsize [0.503, 0.542]} & {\scriptsize [0.654, 0.707]} \\
Proposed Approach & \textbf{92.35} & \textbf{0.545} & \textbf{0.709} \\
 & {\scriptsize [91.86, 92.86]} & {\scriptsize [0.527, 0.570]} & {\scriptsize [0.685, 0.746]} \\
\midrule
\multicolumn{4}{l}{\textit{Paired difference}} \\
Proposed $-$ rigid & $+0.38$ & $+0.022$ & $+0.028$ \\
 & {\scriptsize [0.13, 0.63]} & {\scriptsize [0.008, 0.044]} & {\scriptsize [0.008, 0.061]} \\
Proposed $-$ no opt. & $+2.55$ & $+0.035$ & $+0.048$ \\
 & {\scriptsize [2.10, 3.02]} & {\scriptsize [0.016, 0.062]} & {\scriptsize [0.025, 0.088]} \\
\bottomrule
\end{tabular}

\end{table}

\subsubsection{Leave-one-out Evaluation on Tie Points}
\label{sec:loo}
To evaluate the registration on tie points that do not constrain the solution, we perform a leave-one-out (LOO) evaluation. Each of the 114 cross-session tie-point tracks (50 sonar--camera and 64 sonar--sonar) is withheld in turn, and the graph is re-optimized without it. Since a withheld track has no landmark in the re-optimized graph, and most tracks contain only two observations (61 of the 64 sonar--sonar and 35 of the 50 sonar--camera tracks), the withheld correspondences cannot be scored by reprojection. Instead, each keypoint is projected into the georeferenced mosaic of its session, and the error is the Euclidean distance between the two projected keypoints of a correspondence in the mosaic plane (easting and northing). The fit error is obtained by applying the same metric to the full optimization.

Projecting a keypoint into the mosaic requires the seabed height at the target. No bathymetry is available for the survey, so the mosaics assume a flat seabed: each sonar ping is projected onto a plane at the altitude measured below the towfish, and each image onto a plane at the altitude measured below the camera. For a sonar target at ground range $g$ from a sonar at altitude $h$, an error $\Delta z$ in the seabed height displaces the projected point across-track by approximately $h\Delta z/g$. The seabed below the towfish (pressure depth plus altitude) ranges from approximately 11 to 33~m across the survey. Camera landmarks provide local seabed estimates in areas where the optical surveys overlap the sonar swaths. 98.1\% of the sonar keypoints of the sonar--camera tracks have at least three camera landmarks within 5~m. At these keypoints, the local seabed differs from the flat-seabed assumption by a median of 2.6~m, which corresponds to a median ground-range error of 0.8~m (maximum 1.9~m) at the observed geometry (median $h = 11.3$~m, $g = 41.8$~m).

Table~\ref{tab:loo} reports the fit and LOO errors. The withheld tie points have larger errors than in the fit, so part of the fit reflects fitting to the tie points, but their error is still less than half of that without optimization. The improvement is therefore not limited to the tie points used in the optimization. The flat-seabed assumption also contributes to the errors.

We apply the same evaluation to the rigid baseline (Section~\ref{sec:rigid_baseline}), where each fold repeats only the second stage, as the first stage does not use cross-session tracks. The rigid baseline has larger errors on the withheld tie points than the proposed method, particularly for the sonar--camera correspondences. Its errors increase less from fit to LOO, as a single $\mathrm{SE}(3)$ correction per session cannot fit the tie points closely. Its fit error is also larger than the LOO error of the proposed method. Therefore, the additional degrees of freedom of the proposed method reduce the error on tie points that were not used in the optimization.

\begin{table}[t]
\centering
\caption{Tie-point errors (m) of the proposed method and the rigid baseline with all tracks (fit) and with each track withheld (LOO), measured as the Euclidean distance between the two keypoints of each correspondence in the mosaic plane. The unoptimized trajectories do not use the tie points, so their errors are the same for both methods and with or without a track withheld.}
\label{tab:loo}
\footnotesize
\setlength{\tabcolsep}{3pt}
\begin{tabular}{lcccccc}
\toprule
 & \multicolumn{3}{c}{Proposed} & \multicolumn{3}{c}{Rigid baseline} \\
\cmidrule(lr){2-4} \cmidrule(lr){5-7}
 & Median & Mean & RMSE & Median & Mean & RMSE \\
\midrule
\multicolumn{7}{l}{\textit{All correspondences ($n=134$)}} \\
\quad Unoptimized & 6.24 & 6.72 & 7.59 & 6.24 & 6.72 & 7.59 \\
\quad Fit & 1.84 & 2.29 & 2.81 & 3.38 & 4.30 & 5.25 \\
\quad LOO & 2.22 & 2.74 & 3.29 & 3.48 & 4.57 & 5.59 \\
\midrule
\multicolumn{7}{l}{\textit{Sonar--camera ($n=66$)}} \\
\quad Unoptimized & 5.97 & 6.50 & 7.24 & 5.97 & 6.50 & 7.24 \\
\quad Fit & 1.80 & 2.11 & 2.36 & 3.94 & 5.29 & 6.33 \\
\quad LOO & 1.91 & 2.35 & 2.80 & 4.38 & 5.74 & 6.83 \\
\midrule
\multicolumn{7}{l}{\textit{Sonar--sonar ($n=68$)}} \\
\quad Unoptimized & 7.25 & 6.93 & 7.92 & 7.25 & 6.93 & 7.92 \\
\quad Fit & 2.05 & 2.46 & 3.19 & 2.92 & 3.33 & 3.93 \\
\quad LOO & 2.56 & 3.11 & 3.70 & 3.02 & 3.43 & 4.05 \\
\bottomrule
\end{tabular}
\end{table}

\subsubsection{Camera Extrinsics}
\label{sec:extrinsics_results}

\begin{table*}[t]
\centering
\caption{Estimated camera extrinsics relative to the manual calibration. Rotation differences are rotation vectors about the calibrated camera axes ($x$, $y$, $z$), and translation differences are expressed along the same axes. $\sigma$ is the posterior standard deviation; the prior standard deviation is $50$\,mrad and $50$\,mm per axis.}
\label{tab:extrinsics}
\footnotesize
\setlength{\tabcolsep}{3pt}
\begin{tabular}{lcccc}
\toprule
Camera & $\Delta$ rotation $x/y/z$ (mrad) & $\sigma$ rotation (mrad) & $\Delta$ translation $x/y/z$ (mm) & $\sigma$ translation (mm) \\
\midrule
\textit{CAM1} left  & $-25.4$ / $-120.4$ / $-22.3$ & $3.1$ / $1.1$ / $10.9$ & $-109$ / $251$ / $-255$ & $23$ / $22$ / $30$ \\
\textit{CAM1} right & $1.4$ / $54.5$ / $-20.3$ & $4.1$ / $1.2$ / $10.5$ & $16$ / $227$ / $-265$ & $24$ / $22$ / $29$ \\
\textit{CAM2} & $-15.7$ / $-14.3$ / $-19.1$ & $0.5$ / $0.9$ / $2.4$ & $8$ / $-16$ / $-58$ & $22$ / $21$ / $39$ \\
\textit{CAM3} & $-13.8$ / $-15.8$ / $-7.3$ & $2.1$ / $13.3$ / $6.4$ & $-5$ / $-17$ / $-6$ & $47$ / $45$ / $49$ \\
\textit{CAM4} & $-17.6$ / $-23.9$ / $-9.9$ & $0.7$ / $8.2$ / $3.1$ & $79$ / $-3$ / $-15$ & $38$ / $35$ / $45$ \\
\bottomrule
\end{tabular}
\end{table*}

We compare the estimated extrinsics with the manual calibration used to initialize them (Table~\ref{tab:extrinsics}), and repeat the optimization with the extrinsics fixed at the calibration (prior standard deviation $10^{-6}$), all other settings unchanged.

For \textit{CAM2}--\textit{CAM4}, the rotation corrections are $1.3$--$1.8^\circ$, and the translation corrections are at most $0.08$\,m. Rotations are well constrained (posterior standard deviation 1--27\% of the prior), and translations are weakly constrained (42--97\%). For \textit{CAM1}, both cameras shift by nearly the same translation in the vehicle frame ($0.24$\,m backward, $0.27$\,m upward), about five prior standard deviations per axis. A shift common to the rig is equivalent to a constant offset between the rig and the navigation solution, so the extrinsics may absorb such offsets as well as calibration error.

Table~\ref{tab:extrinsics_ablation} compares the solutions with estimated and fixed extrinsics. Fixing the extrinsics more than doubles the final cost, increases the camera and sonar reprojection errors, and reduces PA by $0.42$ points and mIoU$_2$ by $0.025$, with the 95\% CIs of both differences excluding zero. Estimating the extrinsics therefore compensates for errors in the manual calibration and improves the alignment. However, no independent calibration is available to verify the estimated values.

\begin{table}[t]
\centering
\caption{Solutions with estimated and fixed (manual calibration) camera extrinsics. Camera and sonar: reprojection errors, mean $\pm$ standard deviation over all observations, as in Table~\ref{tab:rigid_non_rigid_reproj_error}. PA and mIoU$_2$ are computed on the common pixel set of the estimated, fixed, and unoptimized maps, with 95\% CIs in brackets computed as in Table~\ref{tab:consistency_map_result}. Difference: estimated $-$ fixed.}
\label{tab:extrinsics_ablation}
\small
\setlength{\tabcolsep}{3pt}
\begin{tabular}{lccc}
\toprule
 & Estimated & Fixed & Difference \\
\midrule
Final cost & $21{,}886$ & $51{,}690$ & \\
Camera (px) & $1.08 \pm 1.72$ & $1.37 \pm 1.93$ & \\
Sonar (m) & $0.57 \pm 0.75$ & $0.81 \pm 1.11$ & \\
\midrule
PA (\%) & $92.48$ & $92.06$ & $+0.42$ \\
 & {\scriptsize [91.96, 92.99]} & {\scriptsize [91.51, 92.57]} & {\scriptsize [0.24, 0.61]} \\
mIoU$_2$ & $0.711$ & $0.686$ & $+0.025$ \\
 & {\scriptsize [0.685, 0.749]} & {\scriptsize [0.662, 0.716]} & {\scriptsize [0.001, 0.054]} \\
\bottomrule
\end{tabular}
\end{table}

\subsection{Convergence Test}
We evaluate how far the initial estimate can be from the solution for the optimizer to still recover it with two tests: perturbations of the anchor positions and a dead-reckoning drift within the optical sessions. The anchor headings are not perturbed. The heading of both vehicles is measured by onboard sensors, and its error is small. The horizontal position, in contrast, can be off by several meters (Section~\ref{sec:system_overview}). A heading error of the AUV sessions is also included in the drift test as a heading bias. In every test, the noise level is increased incrementally and 10 independent runs, each with a fresh random seed, are conducted for each level. The errors are measured between the solution recovered from the perturbed initial estimate and the solution obtained from the unperturbed one.

\subsubsection{Anchor Position}
To evaluate how anchor initialization affects the proposed framework, we perturb the initial position estimates ($x$ and $y$ components) with zero-mean Gaussian noise. All session anchors are perturbed except that of the first sonar session (SON1), which is not perturbed and is excluded from the reported errors. The standard deviation of the noise is increased incrementally, and 10 independent runs are conducted for each noise level. The anchor prior of each perturbed session (Eq.~\eqref{eq:prior_anchor}) is centered on the perturbed estimate. The covariance of the injected noise is propagated into it to first order: its covariance becomes $\boldsymbol{\Sigma}_{\Delta} + \mathbf{J}\boldsymbol{\Sigma}_n\mathbf{J}^\top$, where $\boldsymbol{\Sigma}_n = \sigma_n^2\mathbf{I}_2$ is the covariance of the noise, which is applied in the anchor frame, and $\mathbf{J}$ is the Jacobian of the prior residual with respect to the noise, evaluated at the unperturbed anchor. The test therefore measures recovery from a wrong initial estimate combined with a prior that is equally wrong but correspondingly weaker.

\begin{figure}
    \centering
    \includegraphics[width=\linewidth]{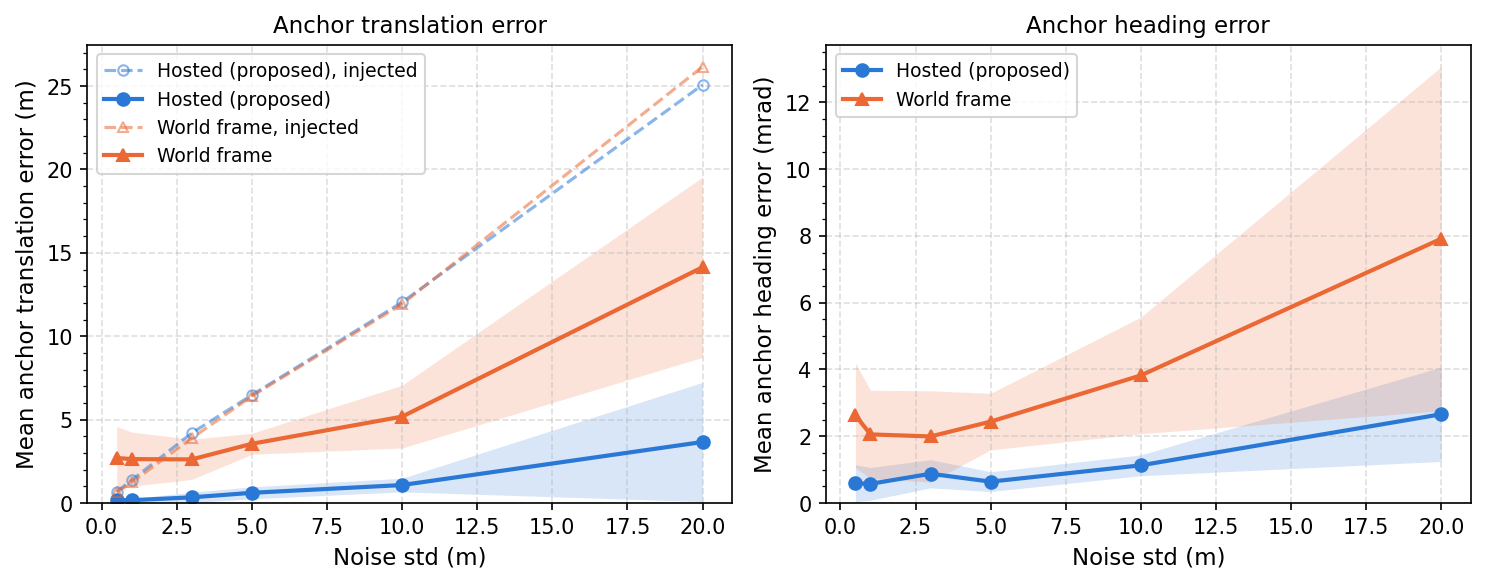}
    \caption{Anchor position perturbation test with hosted (proposed) and world-frame landmark parameterizations. Mean translation error (left) and absolute heading error (right) of the anchors after optimization, averaged over all anchors except that of \textit{SON1}, against the standard deviation of the noise added to the anchors' $x$ and $y$. The dashed lines are the mean offsets injected into the initial estimates of each test, which draw independent noise samples. The shaded regions are $\pm 1$ standard deviation over 10 runs. Translation error is defined as an L2 norm.}
    \label{fig:sensitivity_analysis_anchors}
\end{figure}

Figure~\ref{fig:sensitivity_analysis_anchors} (blue) shows the mean 3D translation error and the absolute heading error of the anchors, with their standard deviation over the runs. The error increases with the noise level. The anchors are recovered to within about 1~m up to 10~m of noise, and the error rises to 3.7~m at 20~m. The heading of the anchors is unaffected by the position perturbation, as its mean error remains below 3~mrad at all noise levels. Furthermore, the widening shaded region (representing $\pm 1$ standard deviation) indicates that the nonlinear least-squares solver becomes increasingly sensitive to the initial guess, frequently converging to different local minima at higher noise levels.

To isolate the effect of the landmark parameterization (Section~\ref{sec:landmark_param}), we repeat the test with every landmark parameterized by its world position $\mathbf{p}_i$. All other settings, the noise levels, and the number of runs are unchanged, and the errors are measured against the unperturbed solution of the same parameterization. No valid-side check is applied to the projection factors, as for the non-host observations of the proposed method. Figure~\ref{fig:sensitivity_analysis_anchors} (orange) shows the results. With world-frame landmarks, the anchors are not recovered even from small perturbations: at 0.5 and 1~m of noise, the error after optimization (about 2.7~m) is larger than the injected error. At every noise level, both errors are larger than with hosted landmarks. This is consistent with the coupling described in Section~\ref{sec:landmark_param}. With world-frame landmarks, moving an anchor also requires moving all landmarks in its session. The Levenberg--Marquardt damping therefore penalizes this movement in proportion to the number of landmarks, causing the optimizer to stop before the anchors have moved back. We also reran the unperturbed world-frame optimization with a tighter solver tolerance ($10^{-10}$ instead of $10^{-5}$). Its cost fell by only 0.5\,\%, yet its anchors moved by 2.4--6.1~m, mostly together relative to \textit{SON1}. The world-frame cost is therefore nearly flat in this direction. Measured against this solution, the error at 0.5 and 1~m of noise is still about 2~m. With hosted landmarks, the solutions from 0.5 and 1~m perturbations agree with the unperturbed one to within 0.2~m at the same tolerance.

\subsubsection{Drift Within the Optical Sessions}
The anchor tests perturb each session as a rigid body. However, the AUV navigation integrates the DVL velocity and the heading while the vehicle is submerged (Section~\ref{sec:nav_factors}), so its error also grows along the trajectory. In the second test, we keep the anchors unperturbed and apply a dead-reckoning drift to the initial estimates of the optical keyframes. In each run, a heading bias $b$, a speed scale error $s$, and a heading random walk $w_k$ are drawn for each optical session, and the keyframe positions are re-integrated along the path:
\begin{equation}
    \tilde{\mathbf{p}}_k = \tilde{\mathbf{p}}_{k-1} + s\,\mathbf{R}_z(b + w_k)\left(\mathbf{p}_k - \mathbf{p}_{k-1}\right), \qquad \tilde{\mathbf{p}}_0 = \mathbf{p}_0,
    \label{eq:drift_sim}
\end{equation}
\noindent where $\mathbf{p}_k$ is the unperturbed horizontal position of keyframe $k$, $b,\, s-1 \sim \mathcal{N}(0, \sigma_d^2)$, $w_0 = 0$ and $w_k = w_{k-1} + \mathcal{N}(0, \sigma_w^2\,\delta t_k)$, with $\delta t_k$ the time between keyframes $k-1$ and $k$. Without the random walk, Eq.~\eqref{eq:drift_sim} reduces to $\tilde{\mathbf{p}}_k = \mathbf{p}_0 + s\,\mathbf{R}_z(b)(\mathbf{p}_k - \mathbf{p}_0)$, which rotates and stretches the trajectory about its start. The drift level $d$ sets $\sigma_d = d/\sqrt{2}$, so this error is $d$ times the distance from the session start. The random walk, with $\sigma_w = \sigma_d/\sqrt{T}$ and $T$ the session duration, reaches the bias spread at the end of the session and bends the trajectory, adding a path-dependent error. The keyframe velocities receive the same rotation and scale; orientations, depths, and measurements are unchanged, and landmarks hosted in a keyframe move with it. The DVL factors are removed from the graph in both the perturbed and the reference runs, so the drift can only be corrected through landmarks observed by other sessions. We test $d$ = 0.1, 0.3, 1, 2, and 5\%. The error of a session is the mean horizontal distance between its keyframes and those of the unperturbed reference run, computed for the initial estimates (injected error) and after optimization.

\begin{figure*}
    \centering
    \includegraphics[width=\linewidth]{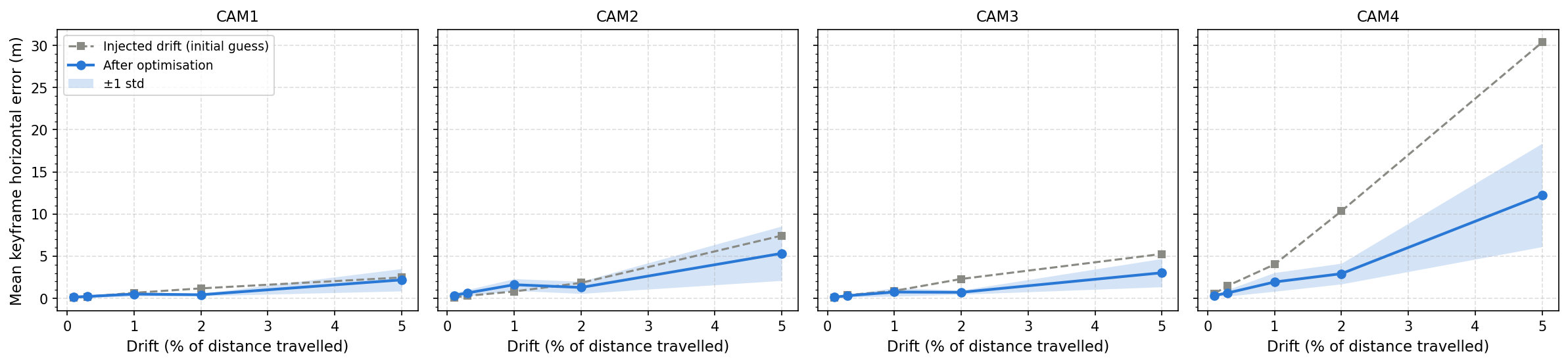}
    \caption{Drift within the optical sessions. Mean horizontal keyframe error of each optical session injected into the initial estimate (dashed) and remaining after optimization (solid), against the drift level. The shaded region is $\pm 1$ standard deviation over 10 runs.}
    \label{fig:sensitivity_analysis_drift}
\end{figure*}

Table~\ref{tab:drift_convergence} and Figure~\ref{fig:sensitivity_analysis_drift} show the results. For drift levels of 2\% and 5\%, the optimization reduces the error of every optical session. The reduction is largest for \textit{CAM4}, the session with the largest injected error. For drift levels up to 1\%, the remaining errors of the other sessions are 0.1--1.6~m, of the same order as the injected error and, for \textit{CAM2}, larger. We attribute this to the spread of the solutions themselves: the optimizer terminates once the relative decrease of the cost falls below its tolerance, which occurs in a flat region of the cost, and runs started from slightly different initial estimates stop at slightly different points. Drift smaller than this spread cannot be resolved by the test. Overall, the landmark observations shared with other sessions correct most of a dead-reckoning drift without DVL measurements, but not all of it, and the residual error is largest at 5\%.

\begin{table}[t]
\centering
\caption{Drift within the optical sessions: mean horizontal keyframe error (m) injected into the initial estimate and remaining after optimization (mean over 10 runs), for each drift level (\% of distance from the session start).}
\label{tab:drift_convergence}
\resizebox{\columnwidth}{!}{%
\begin{tabular}{lcccccccc}
\toprule
& \multicolumn{2}{c}{CAM1} & \multicolumn{2}{c}{CAM2} & \multicolumn{2}{c}{CAM3} & \multicolumn{2}{c}{CAM4} \\
\cmidrule(lr){2-3} \cmidrule(lr){4-5} \cmidrule(lr){6-7} \cmidrule(lr){8-9}
Drift (\%) & Inj. & Opt. & Inj. & Opt. & Inj. & Opt. & Inj. & Opt. \\
\midrule
0.1 & 0.06 & 0.15 & 0.09 & 0.35 & 0.10 & 0.13 & 0.55 & 0.34 \\
0.3 & 0.19 & 0.22 & 0.29 & 0.64 & 0.38 & 0.31 & 1.49 & 0.64 \\
1   & 0.65 & 0.50 & 0.82 & 1.62 & 0.88 & 0.75 & 4.02 & 1.94 \\
2   & 1.18 & 0.42 & 1.84 & 1.29 & 2.30 & 0.71 & 10.35 & 2.92 \\
5   & 2.48 & 2.19 & 7.44 & 5.34 & 5.26 & 3.03 & 30.37 & 12.25 \\
\bottomrule
\end{tabular}
}
\end{table}

\subsubsection{Discussion}
This sensitivity arises from the non-convex cost landscape of the problem. Both the sonar and camera projection factors are non-convex, and combining them does not resolve non-convexity. Instead, their coupling may introduce additional local minima. While certifiably correct solvers exist for specific structures~\citep{papalia2024certifiably}, extending these guarantees to heterogeneous, multimodal factor graphs remains an open problem~\citep{carlone2025slam}. In practice, RTK-GNSS and USBL measurements provide an initialization good enough for the optimizer to converge to a consistent solution in all our experiments, although without a certificate we cannot guarantee that this solution is globally optimal.

\section{Limitations}
\label{sec:limitations}

\paragraph{No ground truth} No independent ground truth, such as surveyed markers or bathymetry, is available for the survey area, so the absolute accuracy of the trajectories and maps cannot be measured. The evaluation uses map consistency (PA, mIoU) and tie-point errors, which measure agreement between sessions, not accuracy. The tie-point errors and map consistency results are also affected by the flat-seabed projection of the mosaics.

\paragraph{Few manual tie points} The cross-session constraints come from 68 sonar--sonar and 66 sonar--optical correspondences, all annotated manually. Annotation is slow and subjective, and the small number of correspondences gives little redundancy: a wrong match can bias the local solution, and large parts of the sonar swaths have no tie points. Automated cross-modal matching, for example with learned modality-invariant descriptors, would remove this bottleneck.

\paragraph{Risk of over-fitting} The non-rigid formulation lets each trajectory deform locally to satisfy the tie points. The tie-point error increases from the fit to the LOO evaluation (Sec.~\ref{sec:loo}), so part of the alignment reflects fitting to the tie points. On a dataset like the one used for evaluation in this paper, a simpler registration of the session mosaics using the same tie points is a relevant alternative. Because this alternative was not evaluated, the present results do not establish whether the additional model complexity yields better map consistency than 2D registration.

\paragraph{Unverified camera extrinsics} The camera extrinsics were calibrated offline and are refined in the graph. No independent calibration is available to verify the estimates, and the translations are only weakly constrained by the data. The estimated extrinsics are therefore free parameters that may absorb part of the trajectory error, not validated calibrations.

\paragraph{Temporal gap} The sonar data (October 2021) and the optical data (April 2024 to July 2025) were acquired 2.5--4 years apart. The method assumes a static seabed. Seagrass growth or loss and sediment transport can change the appearance or position of features, which can lead to incorrect correspondences. They can also change the class labels, which biases the map-consistency evaluation. Change detection that separates stable landmarks from transient ones is a direction for future work.

\paragraph{Lack of diverse evaluation dataset} All eight sessions come from a single site with seagrass, sand, and hard-bottom areas, and a seabed depth of approximately 11 to 33~m. The performance of the proposed method has not been tested in turbid water, which shortens the optical range and reduces camera matches, on low-texture seabeds such as mud, which give fewer tie points, or under strong relief, where the flat-seabed initialization and mosaics are less accurate.

\paragraph{Initialization and navigation inputs} The optimization is local and starts from the navigation solutions. In the convergence experiment, it reduces initial anchor errors of up to 5~m to about 0.5~m on average, and of 20~m to about 3.7~m. Since the residual error and its variability grow with the initial error, coarse pre-alignment is advisable when the georeferencing error is large. The framework also uses filtered navigation solutions instead of raw sensor measurements. A tightly coupled formulation with raw navigation measurements in the factor graph is left for future work.

\section{Conclusion}
\label{sec:conclusion}
This paper addressed the challenge of achieving globally consistent, multi-session, multimodal underwater mapping by aligning data from side-scan sonar and optical cameras. To solve this, a novel framework based on factor graph optimization was developed. The proposed method performs a comprehensive, non-rigid alignment by jointly optimizing vehicle trajectories, 3D landmark positions, session-to-world transformations, and sensor extrinsics. By combining a global rigid transformation for each session with local non-rigid deformations of the trajectory, the framework corrects a wide range of errors, from inter-session offsets to intra-session distortions caused by navigation drift. A host-relative landmark parameterization extends inverse depth to side-scan sonar, which confines each landmark to the valid side of the sensor that first observed it and allows a session to move rigidly through its anchor alone.

The experimental results, validated both qualitatively and quantitatively, demonstrated the framework's effectiveness. The optimization reduced projection errors across all sensor modalities and improved the pixel accuracy of the multi-session multimodal map by 2.55 percentage points. Compared with a rigid-alignment baseline, the mean sonar reprojection error was about four times lower. The median error on tie points withheld from optimization fell to 2.22 m, against 3.48 m for the baseline and 6.24 m before optimization. This shows that the improvement extends beyond the correspondences used in the optimization. The host-relative parameterization proved essential to convergence as it recovered session anchors to within about 1 m from 10 m offsets, whereas with world-frame landmarks even 0.5 m offsets grew rather than shrank. Furthermore, the qualitative results confirmed coherent, consistently co-registered sonar and optical maps.

Our proposed approach is sensitive to the initial guess provided to the optimizer due to the non-convex cost landscape. Future work will focus on developing methods that achieve global optimality under arbitrary initialization. Furthermore, automated feature matching will be investigated to replace manual correspondences, both across side-scan sonar images and between side-scan sonar and optical modalities.

\printcredits

\section*{Declaration of competing interest}
The authors have no conflict of interest to declare.

\section*{Data availability}
Data will be made available on request.

\section*{Acknowledgements}
This work was partially funded by the Spanish Agencia Estatal de Investigación through projects \textit{IURBI - Intelligent Underwater Robot for Blue Carbon Inventorying} (Grant CNS2023-144688) (V. Franchi and N. Gracias) and \textit{ASSiST - Automated Seabed Analysis through Self- Supervised Deep Learning Sonar Technology} (Grant PID2023-149413OB-I00), and by the European Commission through the \textit{Erasmus Mundus Joint Master in Intelligent Field Robotic Systems IFROS} (P. Philip-Ifabiyi).

This research was funded by the European Union under the Horizon Europe Teaming for Excellence project MARBLE (G.A. No. 101136349). Views and opinions expressed are those of the authors only and do not necessarily reflect those of the European Union or the European Research Executive Agency (REA). Neither the European Union nor the granting authority can be held responsible for them.

The authors would like to thank Can Lei, Hayat Rajani, and Rahul Thakkar for their support in image segmentation and map alignment evaluation.

\bibliographystyle{cas-model2-names}

\bibliography{literatura}




\end{document}